\documentclass[letterpaper,12pt,peerreviewca,draftcls]{IEEEtran}
\usepackage{csm16}
\usepackage[margin=1in]{geometry}

\usepackage{url}
\usepackage{graphicx,xcolor}
\usepackage{verbatim}
\makeatletter
\newcommand{\verbatimfont}[1]{\def\verbatim@font{#1}}%
\makeatother
\verbatimfont{\ttfamily\small}

\newcommand{\bi}{\begin{itemize}}\newcommand{\ei}{\end{itemize}}
\newcommand{\be}{\begin{equation}}\newcommand{\ee}{\end{equation}}
\newcommand{\bee}{\begin{enumerate}}\newcommand{\eee}{\end{enumerate}}
\newcommand{\bea}{\begin{eqnarray}}\newcommand{\eea}{\end{eqnarray}}
\newcommand{\beas}{\begin{eqnarray*}}\newcommand{\eeas}{\end{eqnarray*}}
\newcommand{\bc}{\begin{center}}\newcommand{\ec}{\end{center}}
\usepackage[left,pagewise]{lineno} 
\usepackage[english]{babel} 
\usepackage{blindtext}

\colorlet{Edit}{red!50!blue!50!}

\usepackage{amsmath,amssymb,amsthm}
\usepackage{graphicx}
\usepackage{comment}
\usepackage[margin=1in]{geometry}
\usepackage{fancyhdr}
\usepackage{caption}
\usepackage{subcaption}
\usepackage{authblk}
\usepackage{fancyhdr}
\usepackage{afterpage}
\usepackage{tikz}
\usetikzlibrary{positioning}
\usetikzlibrary{scopes}
\usetikzlibrary{decorations.pathreplacing}
\usetikzlibrary{arrows,shapes,automata,backgrounds,petri}
\usepackage{textcomp}

\usepackage[ruled,vlined]{algorithm2e}

\newcommand{\rT}{{\mathrm{T}}}

\title{AutoRally\\
\Large An open platform for aggressive autonomous driving}
\author{Brian Goldfain, Paul Drews, Changxi You,\\ Matthew Barulic, Orlin Velev, \\ Panagiotis Tsiotras, and James M.\ Rehg\\
	POC: B.\ Goldfain (bgoldfain3@gatech.edu)\\ \today }

\newif\ifPDF \ifx\pdfoutput\undefined\PDFfalse \else\ifnum\pdfoutput > 0\PDFtrue \else\PDFfalse \fi \fi
\ifPDF 
\usepackage[pdftex, plainpages = false, colorlinks=true, linkcolor=black, citecolor = green!50!blue, urlcolor = blue, filecolor=black, pagebackref=false, hypertexnames=false,  pdfpagelabels ]{hyperref}
\fi

\begin{document}
\bstctlcite{IEEEexample:BSTcontrol}
\maketitle
\CSMsetup

\begin{figure}[ht]
  \centering
    \includegraphics[width=\textwidth]{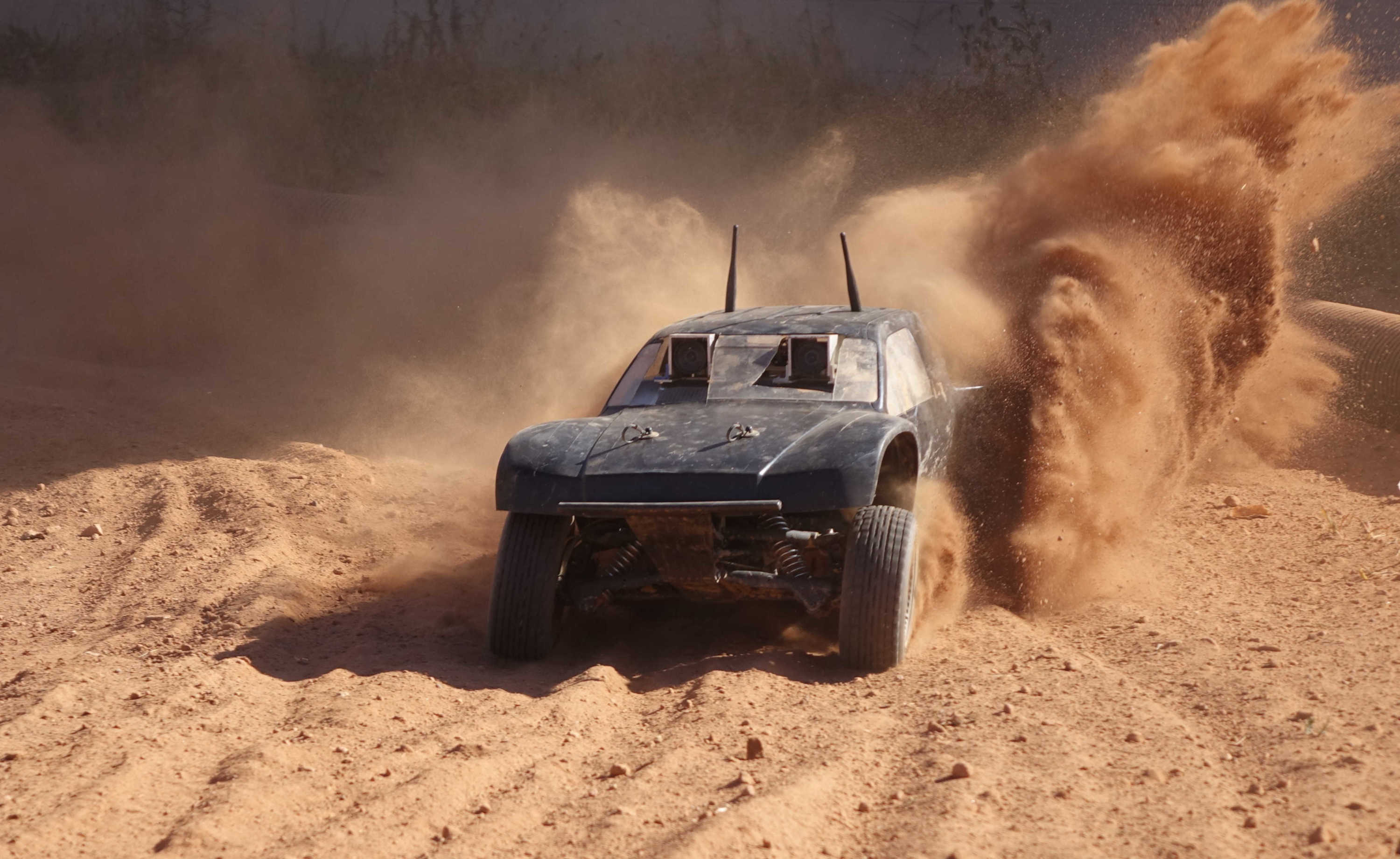}
  \caption{Autonomous driving with AutoRally at the Georgia Tech Autonomous Racing Facility.}
  \label{fig:autorally}
\end{figure}

The technical challenge of creating a self-driving vehicle remains an open problem despite significant advancements from universities, car manufacturers, and technology companies. 
Full autonomy, known as level 5 in the ``\nameref{sidebar:driveAutomation}", is defined as full-time performance by an automated driving system of all aspects of the dynamic driving task under all roadway and environmental conditions that can be managed by a human driver. 
It is estimated that level 5 autonomous vehicles on public roads will help eliminate more than 90 percent~\cite{humanErrors2016} of the 35,000 annual traffic fatalities caused by human error in the United States~\cite{nhtsa2016}, free up commute time for other activities, reduce road congestion and pollution, and increase driving resource utilization~\cite{bertoncello2015ten}.

A largely unexplored regime that level 5 vehicles must master before their introduction to public roads is abnormal and extreme driving conditions. 
These ``corner'' cases include pre-collision regimes, maneuvering at high speed, and driving on surfaces other than asphalt and concrete such as ice, gravel, and dirt, where severe understeer, oversteer, skidding, and contact loss  with the road surface is common. Additionally, dynamic, unpredictable environments such as close proximity to moving vehicles, pedestrians, and other obstacles require short time scales for perception, planning, and control that challenge traditional perception, planning and control methodologies.

It is vital to conduct extensive experimental testing to develop and validate solutions to these ``corner'' cases of abnormal and extreme driving, but unfortunately cost and safety considerations are major barriers that prohibit routine testing and experimentation using full-size vehicles. Large investments, often up to US \$1~million per vehicle, are required for vehicle development and testing, the maintenance of infrastructure and personnel, and the implementation of safety precautions, before any data can be collected. 
No company is prepared to do tests at the limits of performance where damage to the vehicle or injuries to the operator are very likely to occur.
Moreover, even for a trial run, a closed track is necessary unless special precautions are taken, or registration and in insurance is obtained in one of the states that have enacted autonomous vehicle legislation~\cite{selfDrivingLeg2017}.
Collecting data during normal driving conditions on public roads does not help much in terms of understanding these rare, but very important, driving events.

Computer simulations offer an alternative to testing with full-scale vehicles. 
Indeed, the fidelity of computer simulations has improved significantly in recent years.
They can be used to generate an almost unlimited amount of simulated driving conditions cheaply, but they cannot completely replicate the complex interactions of an autonomous vehicle with the real world. For autonomous vehicles to be safe, the failures and unforeseen circumstances encountered during real world testing, many of which are impossible to duplicate in simulation, must be overcome. 
Development and evaluation of new autonomous driving technologies may not be based solely on computer simulations, which may fail to capture critical aspects of the real world.

We believe that scaled vehicles can provide a useful complement to existing methods for testing autonomous vehicle technologies in aggressive driving regimes. Scaled vehicles ranging in size from 1:16 to 1:5 the size of an actual vehicle, often based on radio controlled (RC) vehicles, are easier and less expensive to operate than a full-sized platform. 
Despite recent progress and many publications detailing scaled autonomous testbeds~\cite{ donkeyCar,gonzales2016autonomous,f110,racecar,keivan2013realtime,jakobsen2011autonomous,song2015towards,cutler2014reinforcement, katzourakis2010open, travis2004using}, much of the available results lack reproducibility because of the one-off nature of these testbeds, restrictions imposed by the use of private datasets, and inconsistent testing methods.
Inconsistency is an especially critical problem, as many researchers should be able to test and compare potential algorithms under the same conditions and platforms in order to be able to obtain meaningful comparisons and advance the science of high-speed autonomy.

The remainder of this article describes AutoRally, shown in Figure~\ref{fig:autorally}, a 1:5 scale robotics testbed for autonomous vehicle research. We outline the offline and online estimation methods which were tested on AutoRally, and present experimental results collected with the fleet of six platforms at the Georgia Tech Autonomous Racing Facility (GT-ARF).

AutoRally is designed for robustness, ease of use, and reproducibility, so that a team of two people with limited knowledge of mechanical engineering, electrical engineering, and computer science can construct and then operate the testbed to collect real world autonomous driving data in whatever domain they wish to study. Complete construction and configuration instructions for the AutoRally platform are publicly available, and include all required computer-aided design (CAD) files for custom part fabrication, a complete parts list, and wiring diagrams~\cite{autorallyInstructions}. 
In addition, operating procedures, a simulation environment, core software and reference controllers written in C++ and Python, along with a collection of human and autonomous driving data are publicly available~\cite{autorallySoftware}. See ``\nameref{sidebar:buildautorally}" for more information about the build process. 
\section{Scaled Autonomous Driving Platforms}

Experimental testbeds are an essential component of robotics research that enable real-world experimentation and transition to practice. Below, we summarize some prior efforts in creating scaled platforms for autonomous vehicle research. Full size platforms based on passenger vehicles are not discussed, as they are outside the scope of this article and include many commercial development activities which are not in the public domain.

Scaled platforms constructed from modified RC cars are popular in the academic and hobby communities.
These platforms are typically 0.2~m to 1~m long and weigh between 1~kg and 25~kg. 
Costs range from a few hundred to tens of thousands of dollars, largely determined by the size, sensors, and computing. Construction, maintenance, and programming is typically handled by a small team of students or researchers. Recently, several open source projects released complete documentation and interface software, which is in contrast to the one-off nature of older work that often lacked enough information to replicate.

Documentation for open source platforms normally includes parts lists, build instructions, and interface software for the sensors and actuators. Availability of tutorials, simulation environments, and public datasets vary by project. Common sensors include wheel speed, inertial measurement unit (IMU), cameras, depth sensors, ultrasonic, and light detection and ranging (Lidar) units. The target environment for these platforms is typically indoors on a smooth surface. The Donkey Car~\cite{donkeyCar} is an easy to build 1:16 scale autonomous platform for the DIY Roborace events targeted at hobbyists. Onboard computing and sensing are a Raspberry Pi 3 with a matching wide angle camera. The Berkeley Autonomous Race Car (BARC) \cite{gonzales2016autonomous} is a 1:10 scale vehicle designed as a simple and affordable research platform for self-driving vehicle technologies that has been successfully used to demonstrate various control algorithms. The onboard ODROID-XU4 is similar in computational performance to the Raspberry Pi 3, and the sensor suite includes a hobby grade camera, IMU, four ultrasonic range finders, and Hall effect wheel speed sensors. The F1/10 project~\cite{f110} and accompanying Autonomous Racing Competition allows teams to race against one another using a common 1:10 scale platform developed at the University of Pennsylvania. Computing on the F1/10 platform is performed by an Nvidia Jetson. The sensor suite includes a hobby IMU, compact indoor Hokuyo 2D Lidar, and optional Structure and Zed depth and motion sensing cameras. The 1:10 scale Rapid Autonomous Complex-Environment Competing Ackermann-steering Robot (RACECAR)~\cite{racecar} from Massachusetts Institute of Technology is a platform for researchers creating applications for self driving cars. RACECAR also uses the Nvidia Jetson for computing, and includes the same Hokuyo Lidar and Zed stereo camera as the F1/10 platform. Table~\ref{table:platformsCompare} provides a comparison of these open source scaled platforms.

While all of these platforms are easy to build, moderately priced, and offer some onboard sensing and compute capabilities, their design limits their use to smooth surfaces, typically indoors. All of the platforms lack a global position system (GPS) device, which is a common sensor for outdoor vehicles. Instead of GPS, global position information can be provided by instrumenting the environment such as a VICON external motion capture system or beacons rigidly mounted around the environment. 
These systems restrict the possible operating space to a couple hundred square meters because of sensor field of view and resolution restrictions, and are priced in the tens of thousands of dollars for out-of-the-box solutions. 
The chassis, mounts, and enclosures of the platforms are typically not designed for repeated crashes and collisions that are inevitable when testing novel autonomous vehicle technologies, so the delicate sensors and electronics are easily damaged when something goes wrong. Onboard computing is inadequate for much of the state of the art research because of size and power limitations. This necessitates significant code optimization or offloading of computation to a remote computer. Offboard computation introduces its own set of problems including increased latency, dependence on a reliable, high bandwidth wireless connection, which dictates the size and configuration of testing environments. 
The limited payload capacity and power availability also severely limits the ability to test new sensors such as a Lidar and high frame rate machine vision cameras because the size, weight, and data rates quickly overwhelm the platforms.

\begin{table}
   \centering
   \caption{Comparison of open source scaled autonomous platforms. All platforms are based on 1:10 scale radio controlled cars and include C++ and Python software interfaces that use the Robot Operating System software libraries, except for the Donkey Car. The Donkey Car is 1:16 scale and includes a Python interface. The build time and cost of each platform does not include 3D printed parts, which will vary based on the printer used. In addition to the platforms themselves, the availability of a simulation world and public data sets are indicated.}
  \begin{tabular}{| l | c | c | c | c | c | c | c |}
    \hline
    Platform & Cost [\$] & Build Time [h] & Weight [kg] & Computing & Simulation & Data Sets \\
    \hline
    Donkey Car  & 200   & 2  & 2    & Raspberry Pi  & Y & Y \\
    BARC        & 500   & 3  & 3.2  & Odroid XU4    & Y & Y \\
    MIT RACECAR & 3,383 & 10 & 4.5  & Nvidia Jetson & Y & N \\
    F1/10       & 3,628 & 3  & 4.5  & Nvidia Jetson & N & Y \\
    \hline  
  \end{tabular}
  \label{table:platformsCompare}
\end{table}

Many one-off experimental platforms have been created for specific projects. In~\cite{keivan2013realtime}, a model predictive control (MPC) algorithm running on a stationary desktop computer with a motion capture system has been used to drive a custom 1:10 scale RC platform around an indoor track with banked turns, jumps, and a loop-the-loop. 
Platforms were developed to test autonomous drifting controllers in~\cite{jakobsen2011autonomous} and~\cite{gonzales2016autonomous}, and to push scaled autonomous driving to the friction limits of the system in~\cite{song2015towards}.
A framework with multi-fidelity simulation and accompanying hardware platform for use in reinforcement learning problems relating to autonomous driving was presented in~\cite{cutler2014reinforcement}. A 1:5 scale autonomous platform was developed to investigate stability control in~\cite{katzourakis2010open,travis2004using}. While these platforms were successfully used for the experiments in their respective publications, there is not enough public information available to be able to build, operate, and program one without essentially starting from scratch.

Traditionally, scaled autonomous driving platforms were purpose-built for one experiment, but a new wave of open source platforms are emerging. 
Still, none are robust enough to survive repeatedly pushing the vehicle to its mechanical and software limits, let alone operate in outdoor environments with the payload capacity to carry a variety of popular sensors and powerful onboard computing.
Scaled platforms therefore show great promise in the wide variety of experiments they enable, but previous attempts fall short in terms of design, fidelity, and repeatable performance.
\section{The AutoRally Robot}

\begin{figure}[ht]
  \centering
    \includegraphics[width=\textwidth]{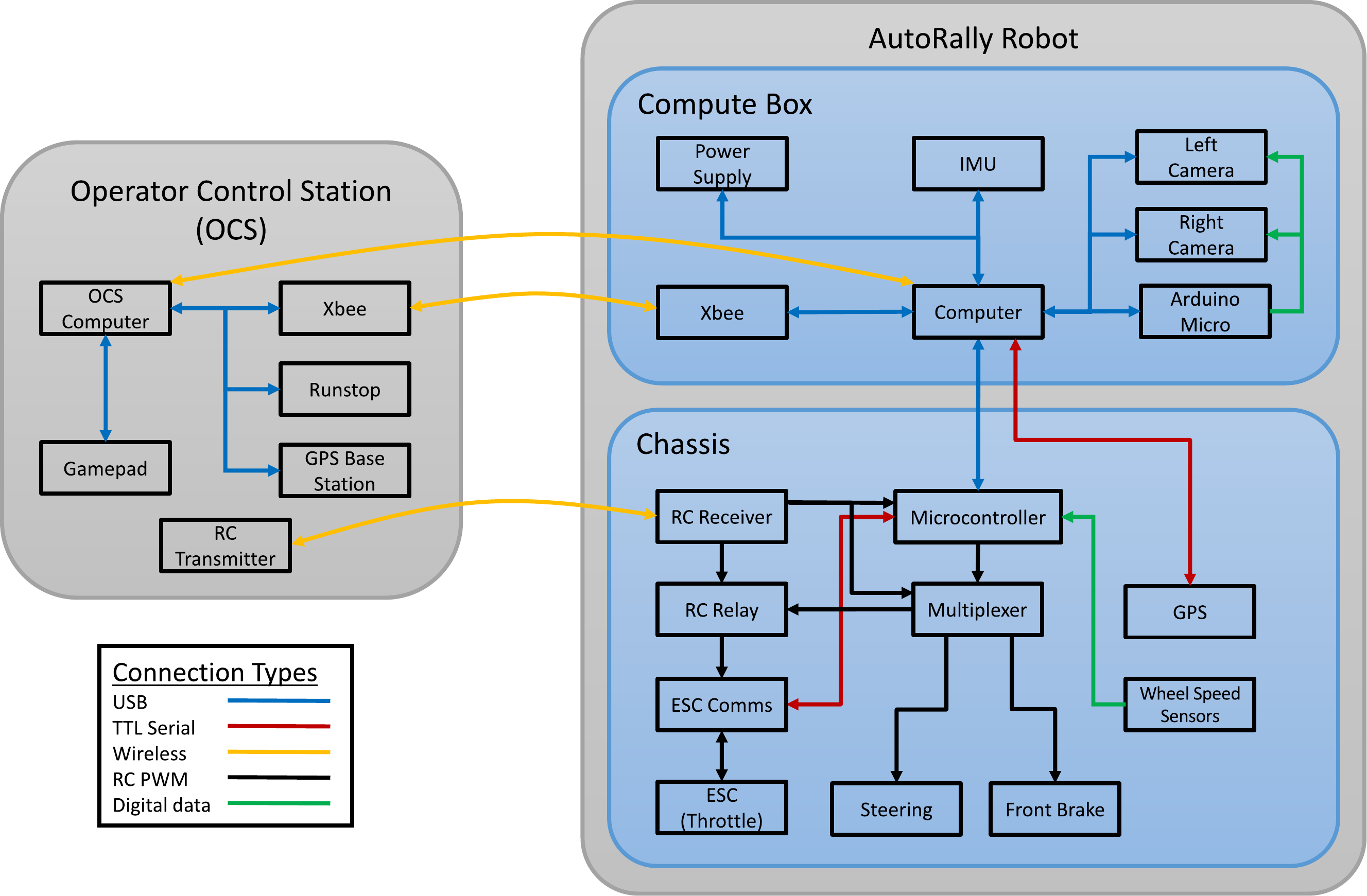}
  \caption{AutoRally system diagram. All major electronic system components and their connections to the rest of the system are shown. The setup includes the AutoRally robot composed of a chassis and compute box along with a remote operator control station (OCS). Communication between the OCS computer and robot are via hobby radio controlled (RC) signals, WiFi, and 900~MHz XBee radios. Sensors, including the inertial measurement unit (IMU) and global positioning system (GPS) receiver are connected to the computer with universal serial bus (USB) cables. The actuators in the chassis include the electronic speed controller (ESC), steering, and front brake, and are controlled by standard 50~Hz hobby pulse width modulation (PWM) signals.}
  \label{fig:systemDiagram}
\end{figure}

The AutoRally autonomous vehicle platform is based on a 1:5 scale RC trophy truck and is approximately 1~m long, 0.6~m wide, 0.4~m high, weighs almost 22~kg, and has a top speed of 90~kph. 
The platform is capable of autonomous driving using only onboard sensing, computing, and power. 
While larger than many other scaled autonomous ground vehicles, the platform offers a cost-effective, robust, high performance, and safe alternative to operating full-sized autonomous vehicles and retains a large payload capacity compared to other scaled platforms built from smaller RC cars. 
AutoRally's capabilities offer a large performance improvement over traditional scaled autonomous vehicles without the need for the large infrastructure investments and safety considerations required for full sized autonomous vehicles. The complete system diagram for the AutoRally robot with a remote operator control station (OCS) is shown in Figure~\ref{fig:systemDiagram}. 
This remainder of this section describes the configuration of the AutoRally chassis and compute box including mechanical components, sensors, and computing configuration.

\subsection{Chassis}

The chassis is designed as a self-contained system that can easily interface to a wide variety of computing and sensor packages.
Similarly to a standard RC car, the chassis can be driven manually using the included transmitter.
Computer control and chassis state feedback are provided by a single universal serial bus (USB) cable connected to an onboard computer. Feedback from the chassis to an attached computer includes wheel speed data, electronic speed controller (ESC) diagnostic information, and the manually-provided actuator commands read from the RC receiver.

\subsubsection{1:5 Scale Radio Controlled Truck}

\begin{figure}[ht]
  \centering
    \includegraphics[width=\textwidth]{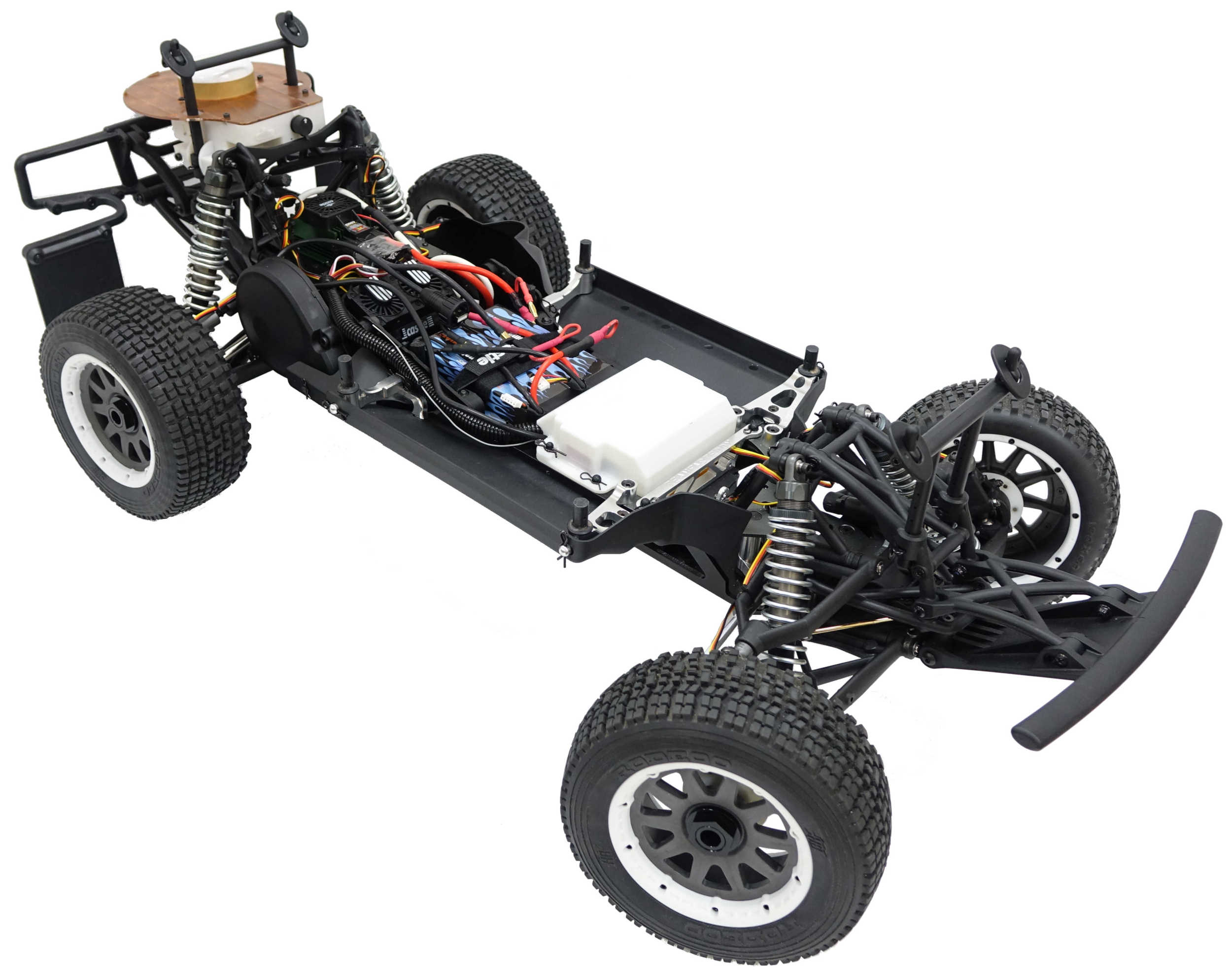}
  \caption{Assembled AutoRally chassis.}
  \label{fig:chassisSide}
\end{figure}

The chassis is based on a HPI Baja 5SC RC trophy truck. Figure~\ref{fig:chassisSide} shows the assembled chassis with all modifications installed and the plastic protective body removed. 
The total weight of the assembled chassis is 13 kg. 
The major upgrades from the stock chassis are an electric powertrain conversion, front brake installation, and electronics box replacement. 
The electric conversion replaces the stock 3~hp, 26~cc 2-stroke gasoline engine with a 10~hp peak output electric motor and ESC from Castle Creations. 
Compared to the stock engine, the electric motor is much more powerful, more responsive, and more reliable.
It also provides an integrated electronic rear brake, generates less heat and no exhaust residue, and requires minimal maintenance. 
The motor and chassis electronics are powered by two, 4-cell 14.8~V 6500~mAh Lithium-Polymer (LiPo) batteries connected in series. 
A full charge lasts between 20 and 90 minutes, depending on usage. 
Front hydraulic brakes are actuated by a separate brake servo.

Parts of the chassis structure were upgraded to handle the increased weight of the sensor and computing package. The stock plastic steering linkage was replaced with billet aluminum parts to withstand the increased steering torque of the upgraded steering servo and weight on the linkage. 
The plastic side rail guards used as mount points for the compute box were replaced with billet aluminum parts to carry the weight of the compute box without deflecting. 
Axle extenders were installed to increase the track of the vehicle by 3.8~cm in order to improve lateral stability and make room to mount the front brakes and the wheel rotation sensors.

The stock suspension springs were replaced with stiffer springs of similar overall dimensions to reduce body roll and improve driving dynamics. A full AutoRally platform weighs 58\% more than the stock chassis, so the spring constants were increased by roughly the same percentage. 
Custom springs are prohibitively expensive, so off-the-shelf springs were sourced as close to the desired dimensions and stiffnesses as possible. The front spring constant increased from 8.48~lb/in to 
15~lb/in and the rear spring constant from 11.17~lb/in to 19.09~lb/in. The shock oil viscosity was also increased roughly 58\% from 500~cSt to 850~cSt to properly damp the upgraded springs.

The stock 2-channel transmitter was replaced with a programmable 4-channel transmitter as part of the electronics box upgrade. The first two channels control the steering and throttle, respectively, and the remaining channels are used in the vehicle safety system discussed in the~\nameref{sec:safety} section.

\subsubsection{Sensors}

To sense wheel speeds, a Hall-effect sensor and magnets arranged in a circular pattern to trigger the sensor were installed on each wheel hub. 
The chosen sensor is a Hallogic OH090U unipolar switch and the magnets are N52 grade 0.3175~cm diameter, 0.1588~cm thick magnets. 
The magnet can trigger the sensor from up to 0.58~cm away. 
Larger magnets could be used to increase the maximum tripping distance but the chosen setup works reliably and fits easily in the wheel hub assemblies. 
Hardware timers in the Arduino Due in the electronics box are used to accurately measure the time between magnets. Inter-magnet timing information is translated to rotation rates and sent to the compute box at 70 Hz.

Inside the electronics box, the RC signals from the receiver are read by the Arduino Due at 50 Hz and sent to the compute box so that, even under manual control, the control signals sent to the actuators can be recorded. This is especially useful for collecting training data where human control signals are required. The Due also receives diagnostic information from the ESC that is forwarded to the compute box.
 
A Hemisphere P307 GPS receiver provides absolute position at 20~Hz, accurate to approximately 2~cm under ideal conditions with real time kinematic (RTK) corrections from a GPS base station. 
The GPS antenna is mounted on top of a ground plane at the back of the chassis along with the receiver. The antenna is located at the maximum distance from the compute box to reduce interference and maximize the view of the sky while still being protected during rollovers. 
The ground plane is an acrylic sheet coated with a copper conductive sheet and is designed to break before the GPS antenna or sensitive GPS board in the event of a severe crash.
    
\subsubsection{Actuators}
    
The chassis requires one servo to operate the steering linkage and one to actuate the master cylinder for the front brakes. Both servos use the 7.4~V digital hobby servo standard which offers more precise, higher torque output, faster response time, and a reduced dead band compared to traditional 6.0~V analog servos. 
All control signals, for both servos and the ESC, are standard 50 Hz hobby pulse width modulation (PWM) signals with a duty cycle from 1~ms to 2~ms, with a neutral value of 1.5~ms. The servos do not have position feedback.

\subsubsection{Custom Components}
    
The custom 3D printed ABS plastic parts installed in the AutoRally chassis are a new electronics box, a GPS box, mounts for the back wheel rotation sensors and magnets, and alignment guides for the front brake disks. ABS plastic is an easy and lightweight medium for quickly manufacturing complex geometries for components that do not experience significant loading. The electronics box replaces the stock one mounted in the front of the chassis superstructure, just behind the steering servo and linkage. Contained within the box are the radio receiver, Arduino Due, servo multiplexer, runstop relay, communication board for the ESC, and servo glitch capacitor. The GPS box contains the GPS board, a Cui 3.3 V, 10 W isolated power supply, a small fan, and the GPS antenna mounted to the ground plane, which is the lid. Front brake disk aligners and mounts for the rear wheel rotation sensors and magnets are installed on the chassis. The front brake disk aligners are needed to keep the disks rotating smoothly because the front wheel rotation sensor magnets unbalance the disks if left to freely rotate.
        
\subsection{Compute Box}

Most modern control and perception algorithms are CPU- and GPU-intensive. 
In order to maximize performance and reduce hardware development and software optimization time, the compute box employs standard components instead of specialized embedded hardware typical of scaled autonomous platforms. The compute box design provides a robust enclosure that mounts to the chassis and fits inside the stock protective body. The weight of the empty compute box is 3.3~kg, and 8.8~kg with all components installed.

\subsubsection{Enclosure Design}

The enclosure is designed to withstand a 10~g direct impact from any angle without damaging internal electronic components and fabricated out of 2.286~mm thick 3003 aluminum sheet. 
The 10~g impact is larger than impacts experienced when testing at GT-ARF according to IMU data that include collisions with fixed objects and rollovers. The box's impact tolerance was verified with finite element analysis (FEA) of the CAD model before fabrication. 
The 3003 aluminum alloy was chosen for its strength, ductility, and relatively light weight. 
The sides of the box are tungsten inert gas (TIG) welded to the bottom to accurately join the large panels of relatively thick aluminum sheet without leaving gaps. 
Aluminum dust filters coupled with a foam membrane allow continuous airflow through the box while keeping out environmental contaminants such as dust and rocks.
Combined with the all-aluminum exterior, the assembled compute box provides excellent electromagnetic interference (EMI) containment.

The cameras and lenses are mounted facing forward on the top of the box on an aluminum plate for rigidity and are protected by covers made from structural fiberglass that does not affect the signal quality for the antennas mounted on the top of the box. Each camera cover is secured to the compute box with four clevis and cotter pins for quick access to the lens and cameras as needed.

Four 3D printed components are inside the compute box: a battery holder, a SSD holder, a GPU holder, and a RAM holder. The battery holder tightly secures the compute box battery and power supply. 
The battery slot is slightly undersized and lined with foam so that the battery press-fits into the mount and can be removed for charging and maintenance without removing any internal screws. The SSD holder is used to securely mount a 2.5~inch SSD to the side wall of the compute box. The GPU holder fits over the GPU and secures it to the main internal strut while still allowing adequate airflow.

The compute box attaches to the chassis with four 3D printed mounts attached to the bottom of the compute box. The mounts fit over vertical posts on the chassis rail guards and are secured to the chassis with a cotter pin though the mount and post. Special consideration was taken to design the mounts as break-away points for the compute box in the event of a catastrophic crash. The mounts are easy and inexpensive to replace and break away before any of the aluminum compute box parts fail to protect the electronics within the compute box. 
By applying lateral forces with FEA and the CAD models, the failure point of the mounts is designed to be at 8~g of force on the compute box compared to the 10~g design load for the rest of the compute box. In practice, the compute box mounts break away during hard rollover crashes and leaves the internal components undamaged. 
All panel mount components such as the power button, LEDs, and ports are dust resistant or protected with a plug to keep out debris.

\begin{figure}[t]
  \centering
  \begin{subfigure}{0.49\textwidth}
    \includegraphics[width=\textwidth]{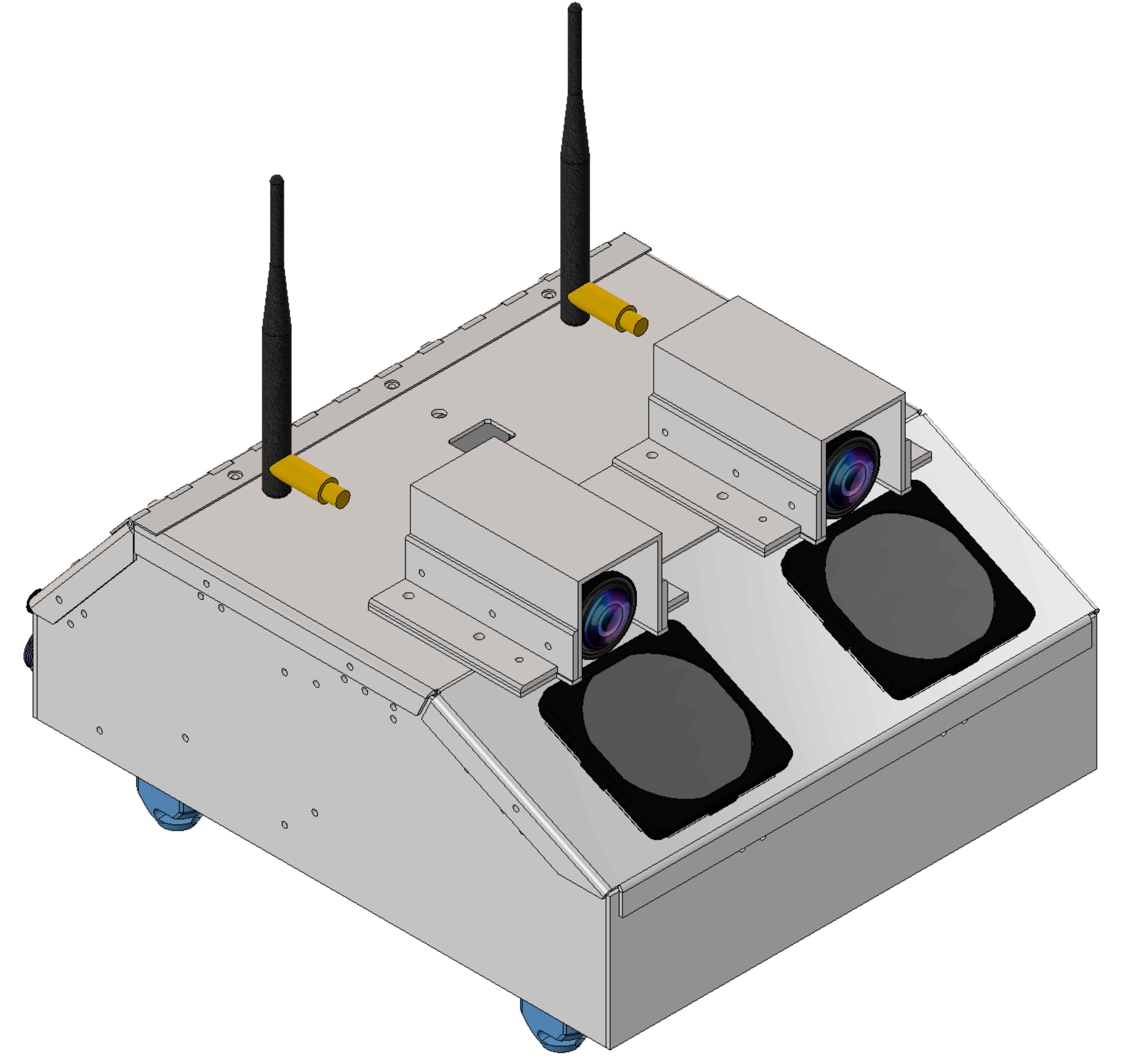}
    \caption{}
    \label{fig:computeBoxCADAssembled}
  \end{subfigure}
  \begin{subfigure}{0.49\textwidth}
    \includegraphics[width=\textwidth]{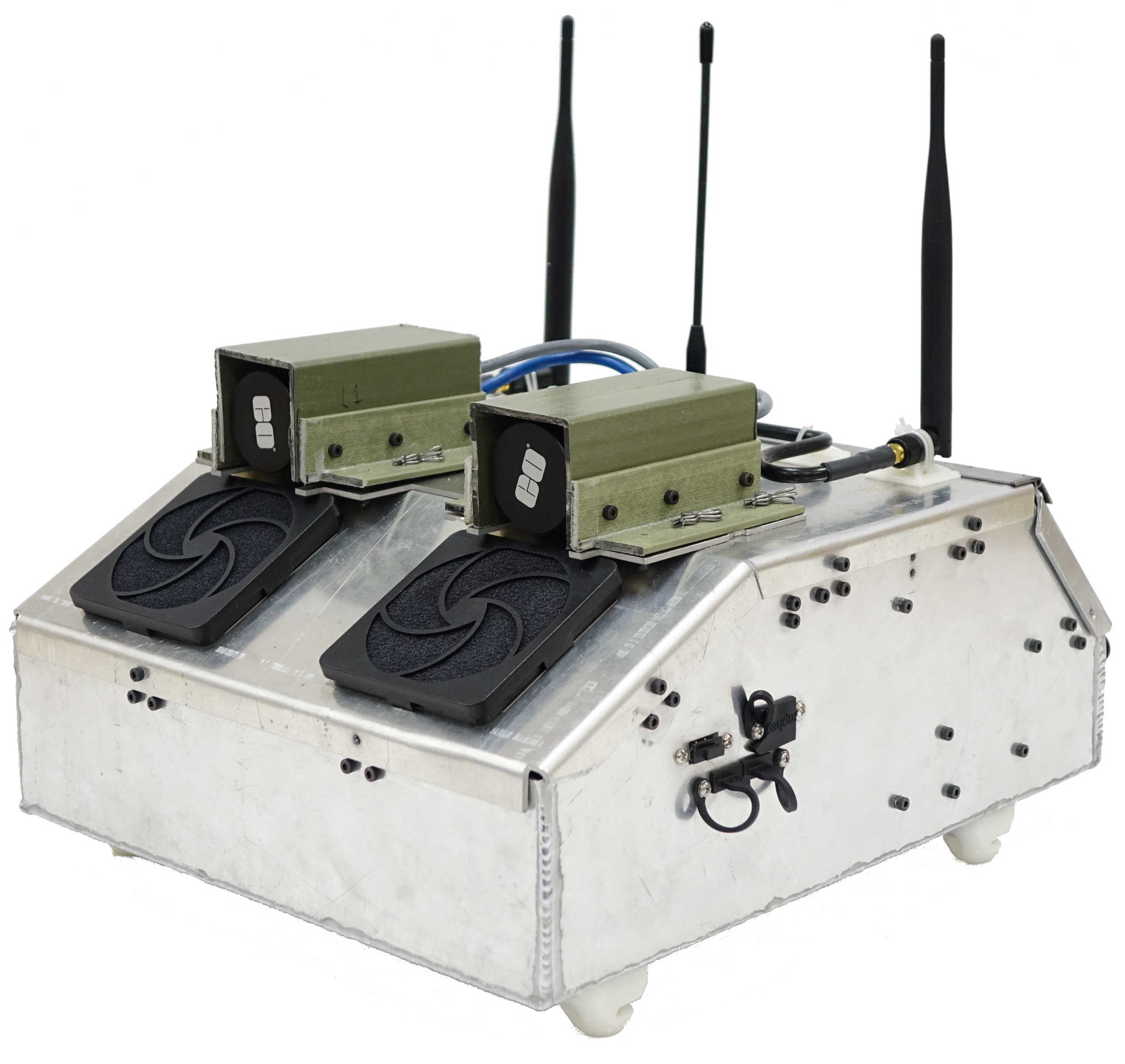}
    \caption{}
    \label{fig:computeBoxAssembled}
  \end{subfigure}
  \begin{subfigure}{0.49\textwidth}
    \includegraphics[width=\textwidth]{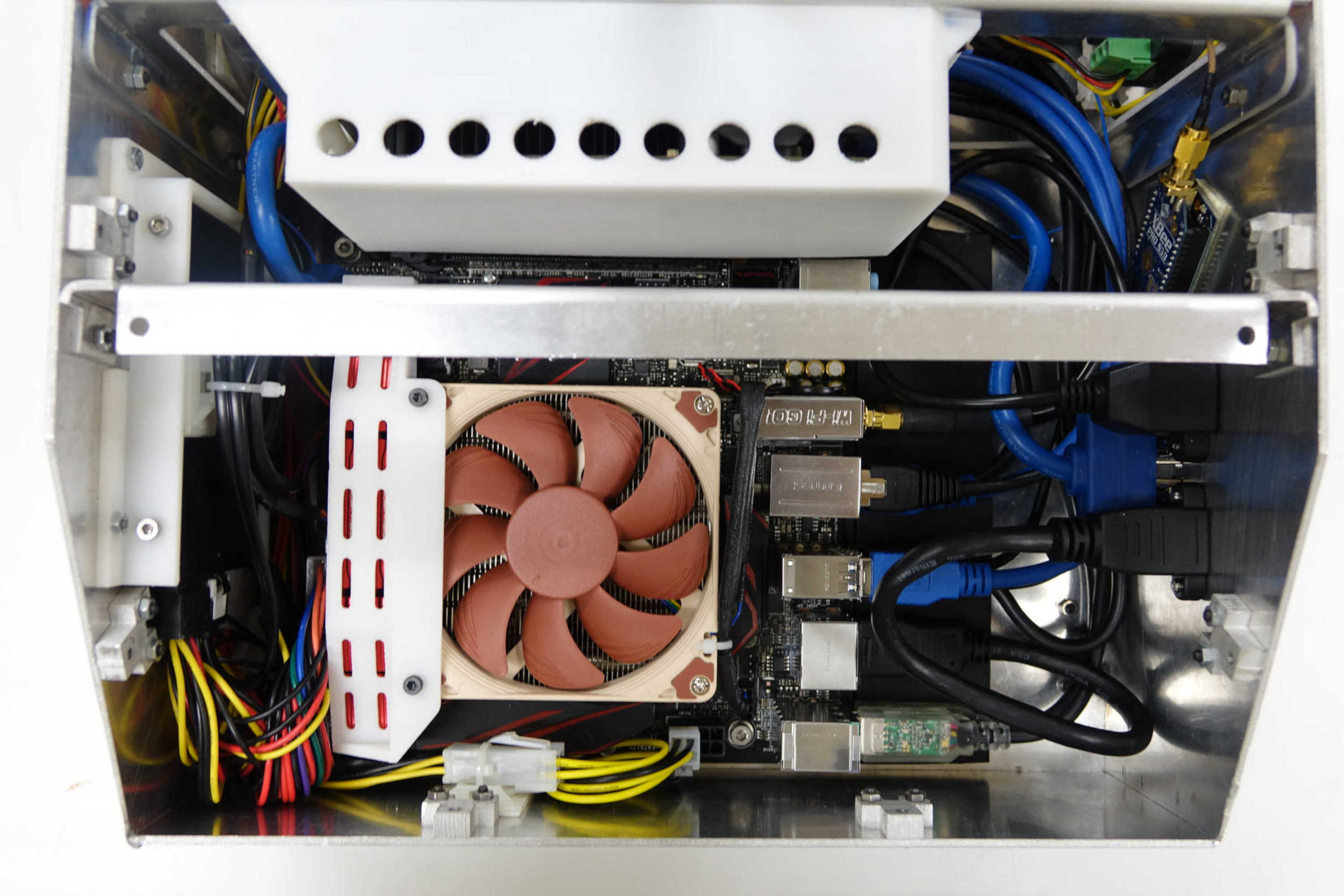}
    \caption{}
    \label{fig:computeBoxFront}
  \end{subfigure}
  \begin{subfigure}{0.49\textwidth}
    \includegraphics[width=\textwidth]{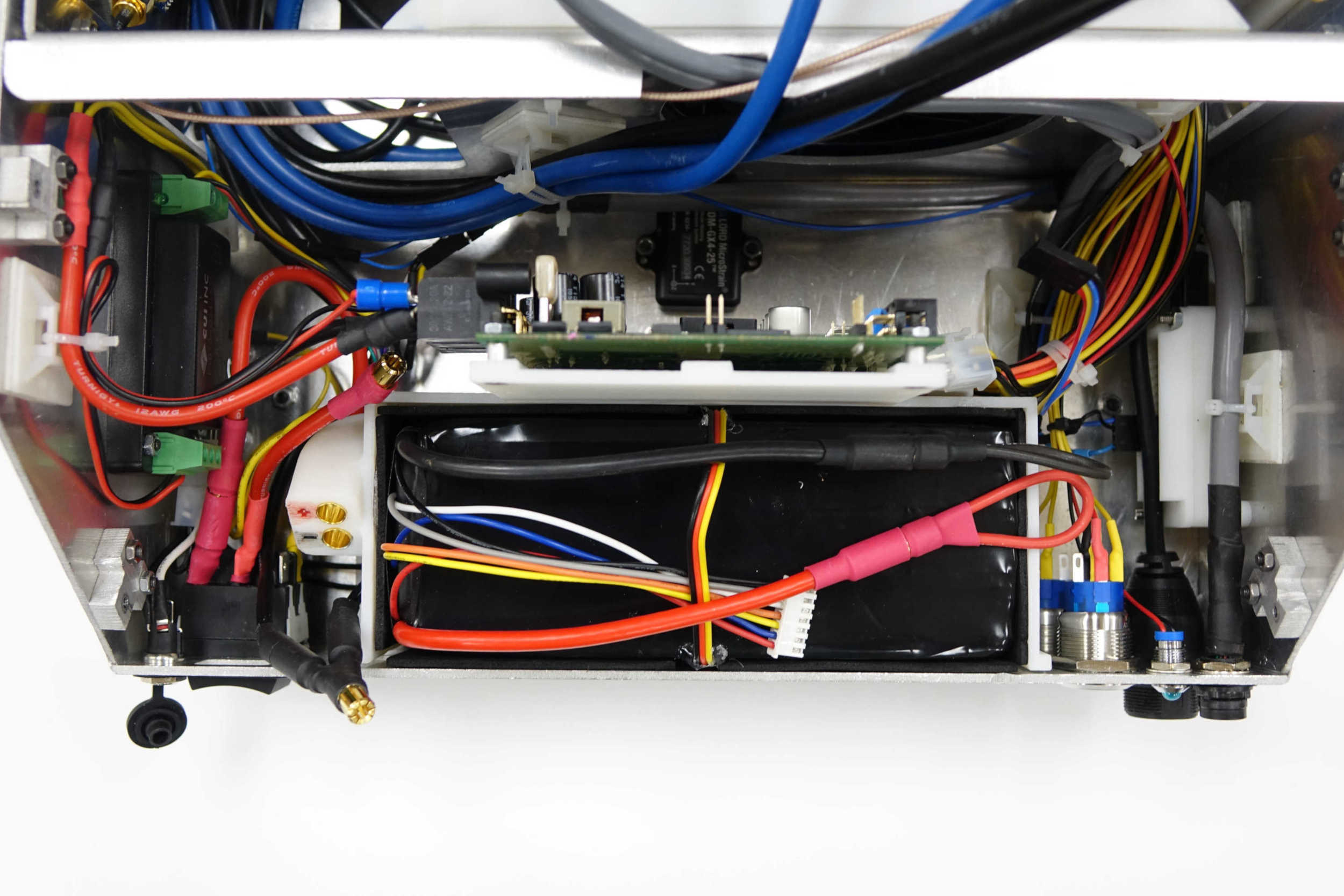}
    \caption{}
    \label{fig:computeBoxBack}
  \end{subfigure}
  \caption{AutoRally mini-ITX compute box. (\subref{fig:computeBoxCADAssembled}) Assembled computer-aided design (CAD) model. (\subref{fig:computeBoxAssembled}) Fully assembled compute box ready to be mounted on a chassis. (\subref{fig:computeBoxFront}) Front of compute box viewed from above with motherboard and compute components visible. (\subref{fig:computeBoxBack}) Rear of compute box viewed from above with power system components visible.}
  \label{fig:computeBoxInsides}
\end{figure}

\subsubsection{Sensors}
    
A Lord Microstrain 3DM-GX4-25 IMU provides raw acceleration and angular rate data at 200~Hz (max 1~kHz) and fused orientation estimates at 200~Hz (max 500~Hz). The two machine vision cameras are mounted on top of the compute box are Point Grey Flea3 FL3-U3-13E4C-C color cameras with a global shutter that run up to 60~Hz. Lenses are 70~degree field of view (FOV), 4.5~mm fixed focal length. Each camera connects to the motherboard with a USB3.0 cable and is externally triggered by an Arduino Micro microcontroller with the general purpose input/output (GPIO) connector. Both cameras are connected to the same trigger signal which runs at a configurable rate. Internal battery voltage and computer temperature sensors are used to monitor system health.

\subsubsection{Computing}

A modular, reconfigurable onboard computing solution was designed that uses standard consumer computer components based on the Mini-ITX form factor. 
Computing hardware development outpaces advancements in almost all other components so the standard form factor, mounting method, and data connections enables the reconfiguration of sensing and computing payloads without mechanical modifications as requirements evolve. Table~\ref{table:computeBoxDet} lists the details of the compute box components.
\begin{table}
   \centering
   \caption{AutoRally compute box components. Major computing and power parts are listed along with their specifications.}
  \begin{tabular}{| l | c |}
    \hline			
    \textbf{Component} & \textbf{Detail}\\
    \hline
    Motherboard & Asus Z170I Pro Gaming, Mini-ITX \\
    CPU & Intel i7-6700, 3.4~GHz quad-core 65~W \\
    RAM & 32~GB DDR4, 2133~MHz\\
    GPU & Nvidia GTX-750ti SC, 640 cores, 2~GB, 1176~MHz\\
    SSD storage & 512~GB M.2 and 1~TB SATA3\\
    Wireless & 802.11ac WiFi, 900~MHz XBee, and 2.4~GHz RC \\
    Power supply & Mini-Box M4-ATX, 250~W \\
  	Battery & 22.2~V, 11~Ah LiPo, 244~Wh \\
    \hline  
  \end{tabular}
  \label{table:computeBoxDet}
\end{table}
WiFi is used to remotely monitor high bandwidth, non-time critical data from the platform such as images and diagnostic information. 
A 900~MHz XBee Pro provides a low-latency, low-bandwidth wireless communication channel. 
The GPS on the robot receives RTK corrections from the GPS base station, transmitted over the XBee radio, at about 2~Hz to improve GPS performance. The XBee radio onboard the robot also receives a global software runstop signal at 5~Hz, and the position and velocity of other AutoRally robots within communication range at up to 10~Hz.
 
The base station XBee, connected to the same computer as the base station GPS, transmits the software runstop message and RTK correction messages to all AutoRally robots in communication range. The runstop message allows all robots within radio range, each running its own self-contained software system, to be stopped simultaneously with one button.

\section{Software Interface} \label{sec:software}

The AutoRally software was designed to leverage existing tools wherever possible. 
All computers in the system run the latest long term support (LTS) version of Ubuntu Desktop to take advantage of the wide availability of compiled packages and minimal configuration requirements. All AutoRally software is developed using the Robot Operating System (ROS)~\cite{quigley2009ros}. 
ROS is a flexible framework for writing robot software. It is a collection of tools, libraries, and conventions that aim to simplify the task of creating complex and robust robot behavior across a wide variety of robotic platforms. Custom ROS interface programs were developed for each AutoRally component that lacked a publicly available interface. The time synchronization and safety systems presented in this section are critical components often overlooked in other scaled platforms, that enable a safe and robust autonomous system. They are a combination of electronics and software. The software interface, OCS graphical user interface (GUI), and simulation environment for the robot are also presented.

\subsection{Time Synchronization}

\begin{figure}
  \centering
  \includegraphics[width=\textwidth]{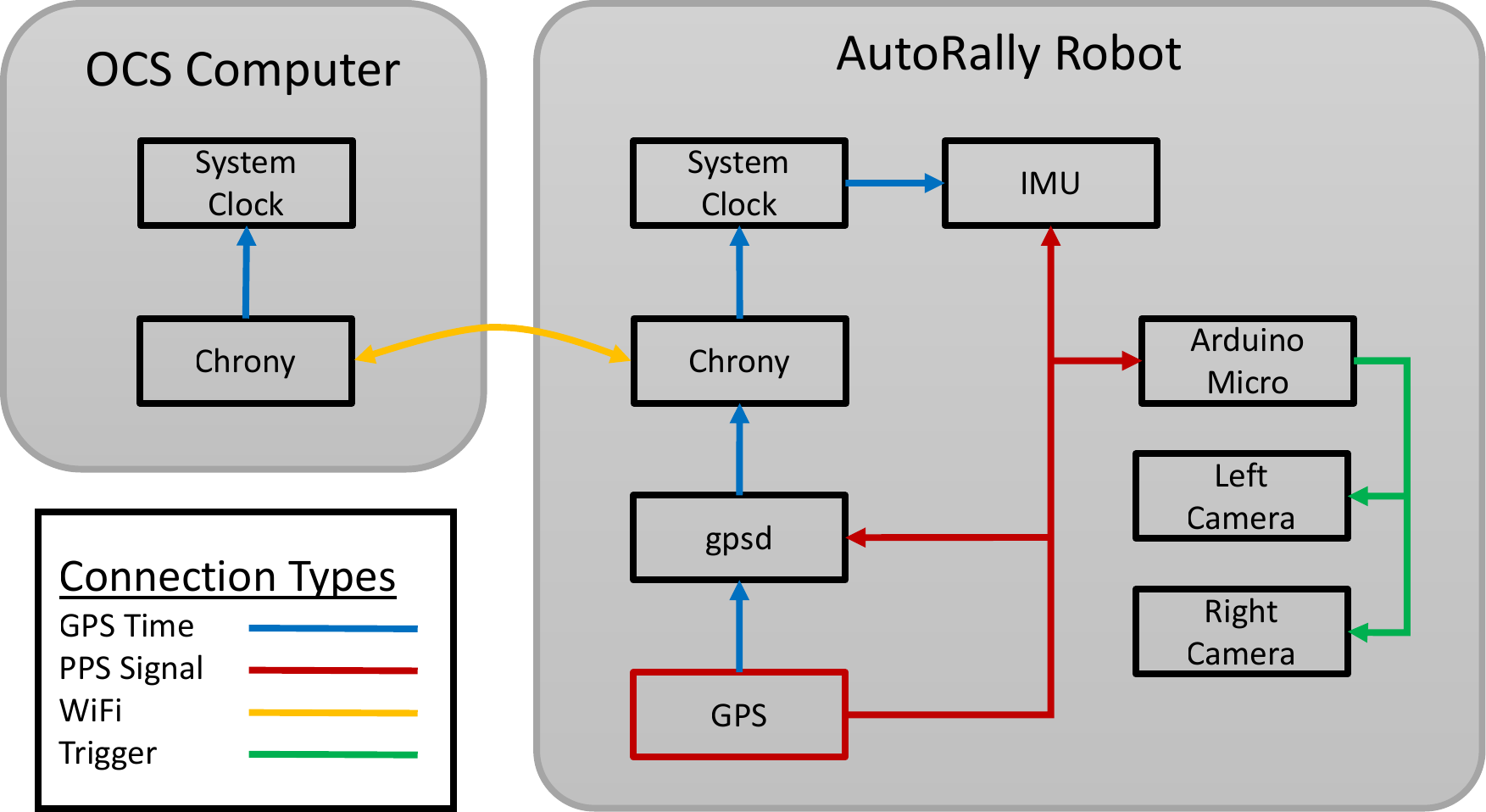}
  \caption{AutoRally time synchronization diagram. Clocks on computers and sensors that support clock control are synchronized to global positioning system (GPS) time with a combination of the pulse-per-second (PPS) signal that marks the beginning of a second, and timing messages that identify which second the PPS signal represents. On the AutoRally robot and operator control station (OCS) computers, the system utilities gpsd and Chrony are used for clock synchronization. The two cameras rely on an external tigger signal to capture frames at the same time. The Arduino Micro microcontroller generates the camera trigger signal.}
  \label{fig:timesync}
\end{figure}

Distributed system design requires robust time synchronization across all components in the system. Accurate timing is especially important as asynchronous data and control rates increase. Time synchronization is performed within the AutoRally system on all computing and sensing components with a combination of Ubuntu tools. 
Figure~\ref{fig:timesync} shows how timing information is propagated for time synchronization. The time source for the entire system is the GPS board on the chassis which emits National Marine Electronics Association (NMEA) 0183 messages and a pulse-per-second (PPS) signal. 
The PPS signal provides a marker that is accurate to within a few nanoseconds of the start of every second according to GPS time. NMEA 0183 time messages corresponding to each PPS pulse provide timing information about that pulse. NMEA 0183 and PPS signals are widely supported by devices that require time synchronization. The PPS signal is routed into GPSD running on the motherboard, the IMU, and the Arduino Micro.

GPSD is a daemon used to bridge GPS time sources with traditional time servers. GPSD runs on the compute box and receives the GPS PPS signal and NMEA messages. 
Processed timing information is communicated through a low latency shared memory channel to Chrony, the time server running on the computer. Compared to traditional Network Time Protocol (NTP) servers, Chrony is designed to perform well in a wide range of conditions including intermittent network connections, heavily congested networks, changing temperatures, and systems that do not run continuously. 
Chrony's control of system time makes time synchronization transparent to programs running on the computer. The system time of the OCS computer is synchronized to the AutoRally robot by a second Chrony instance on the OCS Computer that communicates over WiFi with Chrony on the robot.

The IMU provides a dedicated pin for a PPS input. 
In addition to the PPS signal, it requires the current GPS second (GPS time is given in seconds since Jan 6, 1980) to resolve the time of the PPS pulse. This value can be derived from the computer's system clock. The IMU uses these two pieces of information to synchronize its own clock and time stamp each measurement with an accuracy of significantly less than one millisecond to system time.

The cameras provide an external trigger interface to control when each image is captured. 
The Arduino Micro is used to provide the cameras with the triggering pulse at a specified frame rate. Each time a PPS pulse comes from the GPS, a train of evenly spaced pulses at the rate specified in the ROS system is sent to the cameras. The cameras images are time stamped with the system time when they are received by the computer.

\subsection{Safety System} \label{sec:safety}

\begin{figure}
\centering
\includegraphics[width=\textwidth]{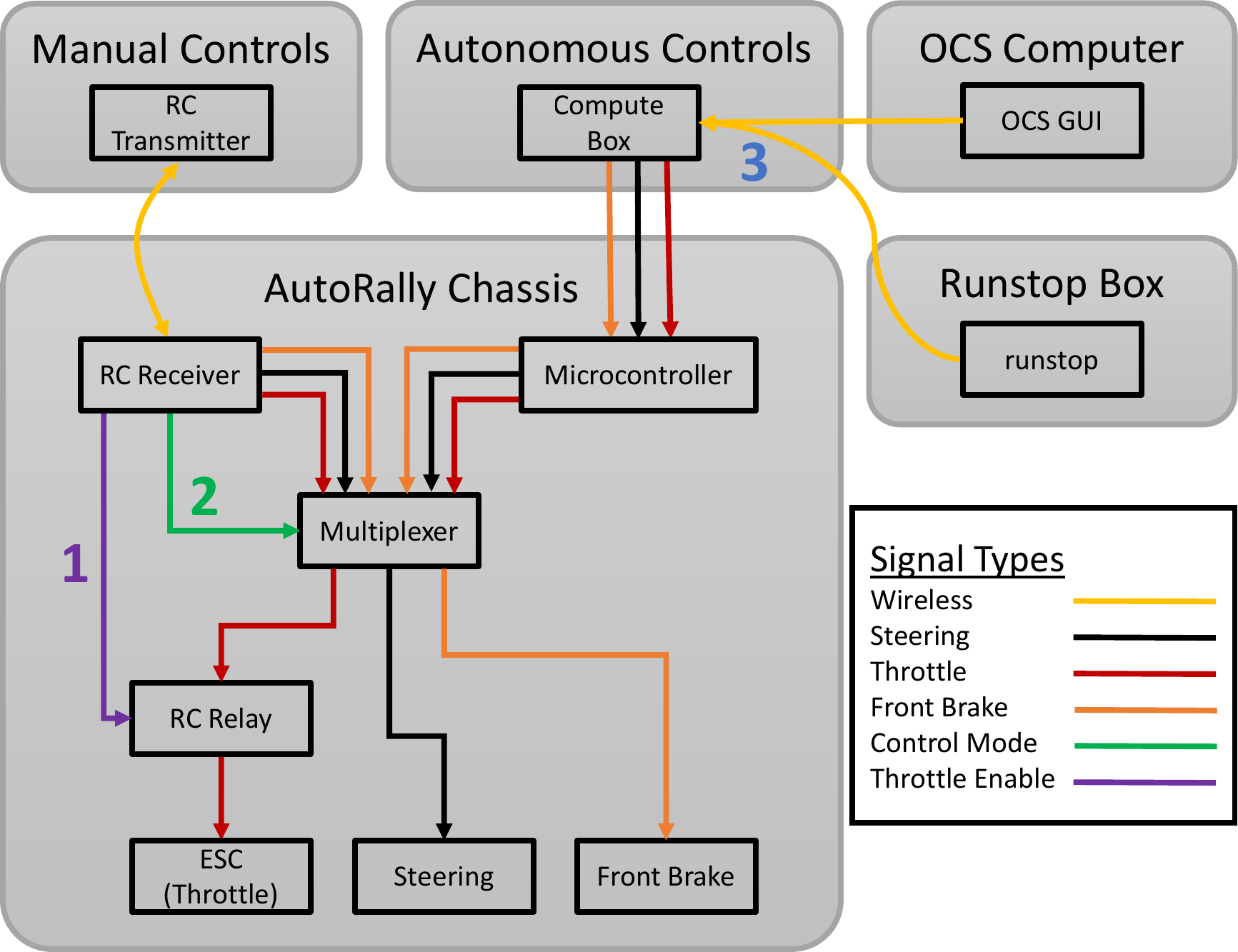}
\caption{AutoRally safety system. The human operated radio control (RC) transmitter sends signals to the RC receiver in the AutoRally chassis. The RC receiver provides actuator signals from the human driver, controls a safety relay to enable and disable the throttle signal into the electronic speed controller (ESC), and switch between human and computer control. Information relating to the state of the safety system is presented to the human operator in the operator control station (OCS) graphical user interface (GUI). Layer 1 of the safety system, shown in purple, is the throttle relay that acts as a wireless throttle live man switch. Layer 2, shown in green, allows seamless, remote switching between autonomous and manual control modes. Layer 3, shown in blue and yellow, is the software runstop used to disable autonomous motion.}
\label{fig:safety}
\end{figure}

The three layer AutoRally safety system is designed to remotely disable robot motion in the event of any software or electronics failure.
The three layers, shown in Figure~\ref{fig:safety}, are a wireless deadman relay located in the electronics box to disconnect the throttle signal, remote switching between autonomous and manual control with a PWM signal multiplexer, and a software based runstop message. 
The relay and autonomous/manual modes are controlled by the state of buttons on the transmitter which circumvent the Wifi, XBee, and software control on the compute box by using the additional RF link between the RC transmitter and receiver located in the electronics box of the chassis. The deadman relay monitors the quality of the incoming PWM control signal so that the throttle signal is automatically disabled in the event of a signal failure between the transmitter and receiver. Additionally, the throttle signal is connected through the normally open contact of the deadman relay so that the throttle signal disengages in the case of a power failure on the robot.

\begin{figure}
\centering
\includegraphics[width=\textwidth]{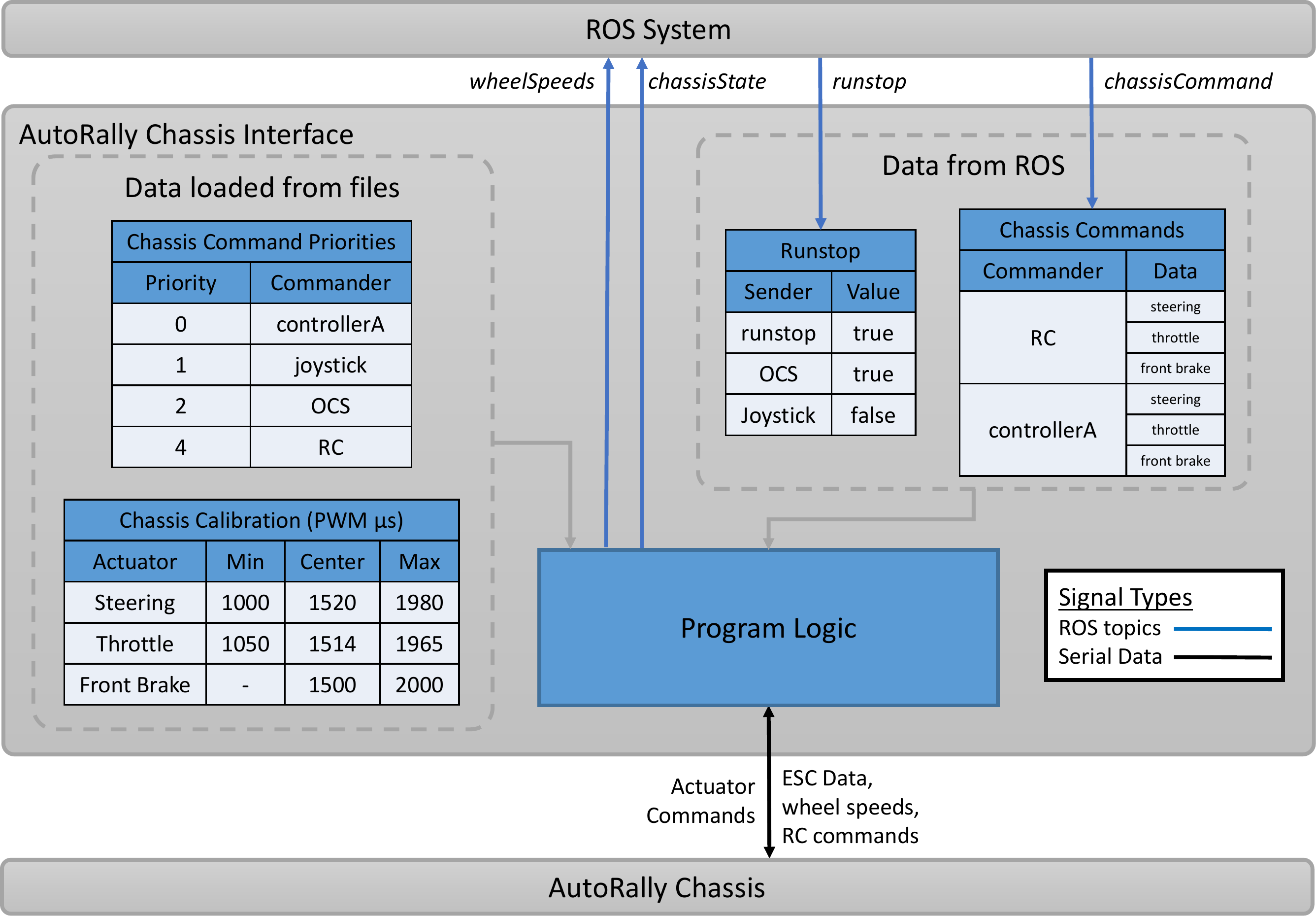}
\caption{AutoRally chassis interface program information flow. The program, which runs on the compute box onboard the robot, uses a combination of configuration files loaded at runtime and messages arriving from the Robot Operating System (ROS) interface to send the highest priority actuator commands over a universal serial bus (USB) connection to the microcontroller in the AutoRally chassis. Simultaneously, the AutoRally chassis sends state information back to the chassis interface program that includes electronic speed controller (ESC) data, wheel speeds, and human provided actuator commands from the radio controlled (RC) receiver. The information received from the chassis is published into the ROS system, and can be viewed in the Operator Control Station (OCS) graphical user interface.}
\label{fig:chassisDiagram}
\end{figure}

Runstop is implemented in software by the AutoRally chassis interface program, shown in Figure~\ref{fig:chassisDiagram}, using incoming runstop ROS messages to enable and disable software control of the robot. 
Any program in the AutoRally system can publish a runstop ROS message. The chassis interface determines whether autonomous control is enabled with a bitwise OR operation of the most recently received runstop message from each message source. By default, the OCS GUI and runstop box send runstop messages. The OCS runstop message is controlled by a button in the GUI and is transmitted over WiFi from the OCS computer to the robot. 
The runstop box sends a runstop message, controlled by the button state of the runstop box, over XBee to the robot. 
Data transmitted over the base station XBee is delivered to every robot within communication range. 
This means that, even though there could be multiple AutoRally robots running at the same time, each with its own self-contained ROS system, the runstop box signal controls autonomous motion for all of the robots simultaneously. The AutoRally robot does not have a true emergency stop that disconnects actuator power because the size, cost, and power requirements for such a system do not fit within the current package. In practice, the three layer AutoRally safety system allows an operator to disable motion and assume manual control of the platform without delay.

\subsection{AutoRally Chassis Interface}

The AutoRally chassis interface software is implemented as a ROS nodelet and communicates with the micro-controller in the chassis electronics box through a USB cable. The interface sends actuator commands to the chassis and receives chassis state information including wheel speeds, the human provided control commands read from the RC receiver, ESC diagnostic information, and safety system state information.

The throttle, steering, and front brake of the robot are controlled by 50~Hz PWM signals standard in the hobby RC community. 
The AutoRally chassis software interface provides a calibration layer above the PWM signal for standardization across platforms and to prevent physical damage so commands do not exceed the mechanical limits in the steering and brake linkages. 
The chassis calibration is stored in a file loaded at runtime by the chassis interface software. Specified in the file is the minimum, center, and maximum pulse width for each actuator in $\mu$s. 
When properly calibrated, a \textit{/chassisCommand} ROS message on any AutoRally platform will elicit the same behavior. 
For example, commanding a steering value of zero will make any calibrated AutoRally platform drive in a straight line. 
Valid actuator command values in the \textit{/chassisCommand} message are between [-1,1]. A steering value of -1 will turn the steering all the way left and a value of 1 will steer all the way right. A throttle value of -1 is full (rear) brake and 1 is full throttle. The front brake value ranges from 0 for no brake to 1 for full front brake while negative values are undefined.

On startup, the chassis interface loads a priority list of controllers from a configuration file. 
The priority list is used while operating to determine which actuator commands arriving from various controllers are sent to the actuators. 
The priorities encode a hierarchy of controllers and define a mechanism to dynamically switch between controllers and use multiple controllers simultaneously. This system allows high priority controllers to subsume control from lower priority controllers, as desired. 
Additionally, each actuator can be controlled by a separate controller such as a waypoint following controller for the steering and a separate velocity controller for the throttle and front brake.

\subsection{Operator Control Station}

The OCS GUI is a tab-based program built using QT that presents real-time diagnostic information, debugging capabilities, and a software runstop to a remote human operator for the AutoRally robot. 
Wheel speed data, real time images from the onboard cameras, and all diagnostic messages from the ROS \textit{/diagnostics} topic which contain detailed information about the health of running nodes are displayed. Diagnostic messages are color-coded by status and grouped by source for fast status recognition by the human operator. 
All of the data between the OCS GUI running on a laptop and the robot travels over a local WiFi network.

The OCS GUI also provides an interface for direct control of the robot's actuators via sliders. While this interface is not appropriate for driving the car, it is used to debug software and hardware issues related to the actuators.

\subsection{Simulation}

\begin{figure}
  \centering
  \begin{subfigure}{0.49\textwidth}
    \includegraphics[width=\textwidth]{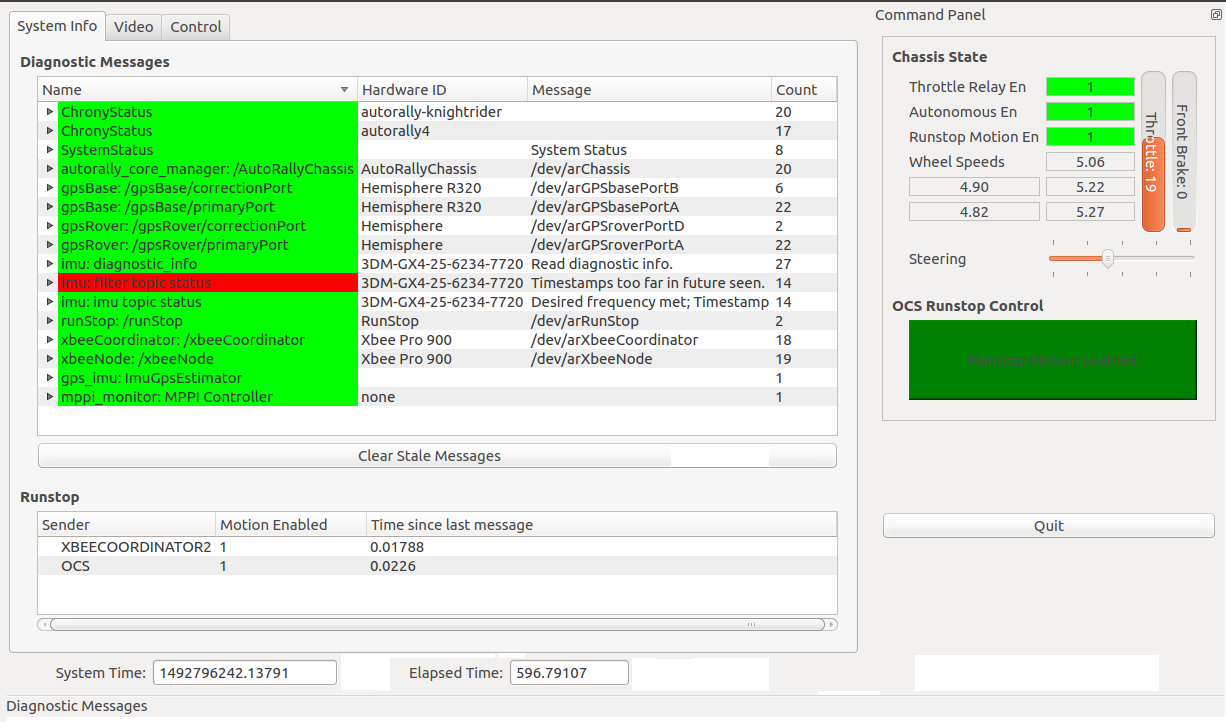}
    \caption{}
    \label{fig:ocs}
  \end{subfigure}
  \begin{subfigure}{0.47\textwidth}
    \includegraphics[width=\textwidth]{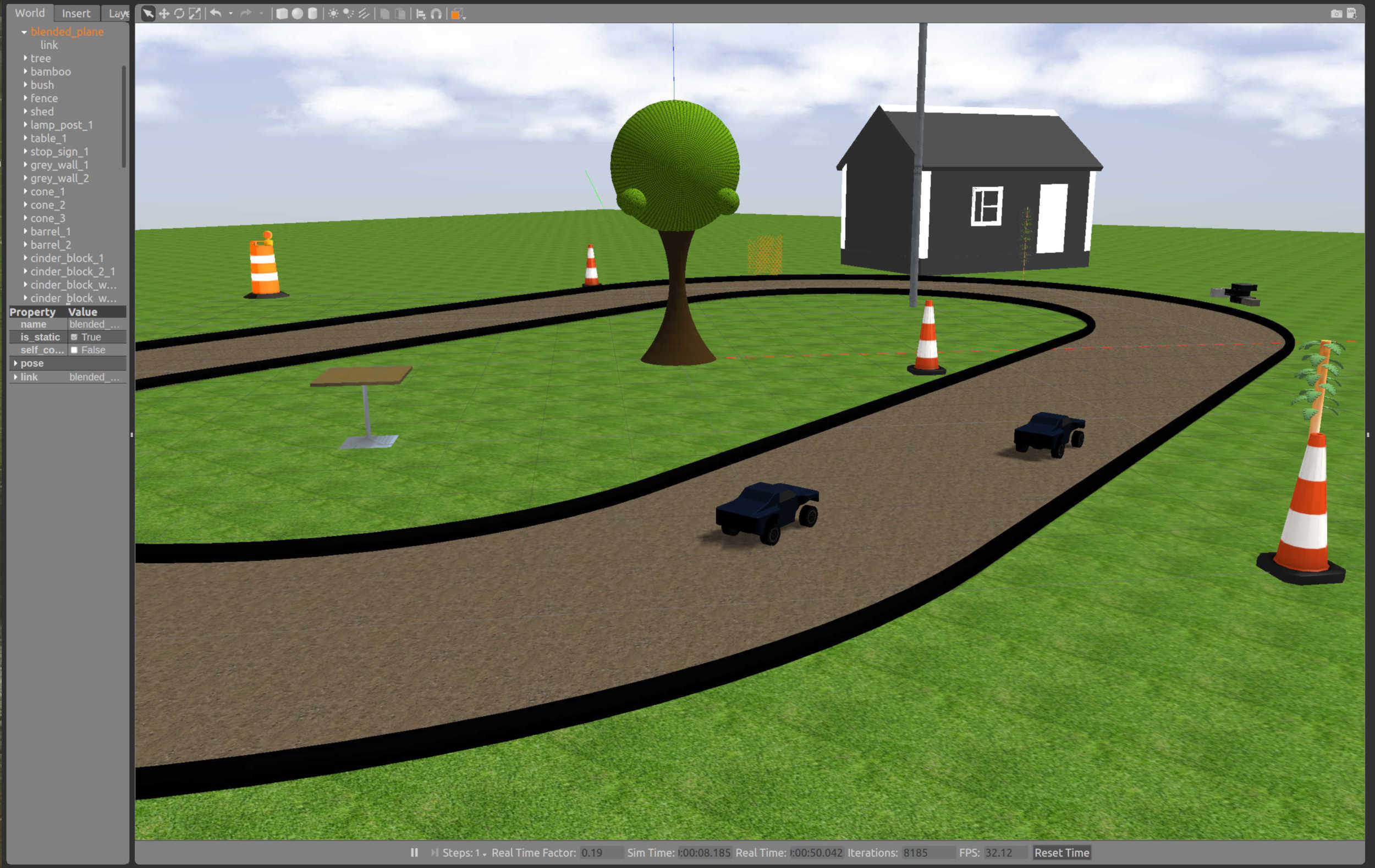}
    \caption{}
    \label{fig:simulation}
  \end{subfigure}
  \caption{Operator Control Station (OCS) graphical user interface (GUI) and simulation environment. (\subref{fig:ocs}) OCS GUI showing diagnostic, sensor, and actuator information captured during autonomous testing. (\subref{fig:simulation}) The simulation environment built in Gazebo is modeled after the Georgia Tech Autonomous Racing Facility. Multiple AutoRally robots can be simulated together and each simulated platform has the same Robot Operating System (ROS) messaging interface and simulated sensors as the physical AutoRally platform. All simulation vehicle parameters such as mass, moments of inertia, and sensor placement and characteristics are set according to their experimentally determined values from a physical robot.}
  \label{fig:ocsGazebo}
\end{figure}


Despite the robust AutoRally hardware platform, there are still high-risk maneuvers and software testing that are best run in a simulation environment before executing them on the physical platform. 
A simulation also allows the careful control of environmental parameters for gathering statistical data which requires performing repetitive or time-consuming experiments that would take weeks or more of testing on the physical platform.

The simulation environment, shown in Figure~\ref{fig:simulation} is based on Gazebo. Gazebo is a robot simulator with tight ROS integration that includes graphical interfaces and multiple physics engines to choose from. 
The AutoRally Gazebo simulation environment and robot model match their real-world counterparts and support the same software interface through ROS messages as the real hardware. 
The simulated track environment is the same size and configuration as GT-ARF. The steering servo and Ackermann linkage of the physical robot are approximated by ROS joint effort controllers that apply torque to turn each front wheel about the vertical axis. The no-load rotation speed, maximum torque, and joint limits used in simulation are measured from the steering servo specification provided by the manufacturer and by measuring the steering linkage angles relative to the chassis center line. The powertrain is approximated by another ROS effort controller that applies torque on the rear axle of the Gazebo model. 
The maximum applied torque and angular velocity are calculated from the motor manufacturers' specifications. 
The differential in the physical platform is neglected in the simulation. 
The suspension for each wheel is modeled with a proportional-integral-derivative (PID) controller on a linear actuator with a target set-point which determines the ride height of the vehicle. The I and D terms are calculated from the dimensions and coefficients of the robot's spring configuration.

The simulation and physical platform implement identical ROS messaging interfaces to enable seamless software migration between hardware and simulation. Simulated GPS, IMU, and cameras come from the hector\_gazebo\_plugins ROS package from TU Darmstadt~\cite{hectorGazebo2017} and are configured according to the specifications of their physical analogs. We developed our own wheel speed sensor node, as no similar functionality was publicly available. 

Overall, Gazebo is not considered a high fidelity simulator with respect to graphics rendering and physics realism for autonomous vehicles, especially as the vehicle approaches and surpasses the friction limits of the system.
The main reason for using the Gazebo as the simulator was not so much to produce the most accurate visuals and dynamics, but rather as part of the hardware and software infrastructure that allows for
a smooth testing of software with ROS and the AutoRally platform, and to easily debug the control, perception, and communication software.
\section{Estimation}
\label{sec:estimation}

Parameter estimation is an essential part of controller design, especially for model-based controllers such as MPC, which rely on accurate dynamics models for motion prediction. 
This section details the offline and online estimation performed with the AutoRally platform. 
Offline, parameters were estimated to determine the platform moments of inertia (MOI) using the bifilar pendulum method.
Three different vehicle models of increasing fidelity are presented. While the higher fidelity models can be used to more accurately predict vehicle motion, the model parameters can be significantly more difficult to estimate, and computationally expensive to compute. A joint-state unscented Kalman filter (JS-UKF) was used to find the parameters of single-track and double-track vehicle models with a realistic tire forces model. An 11 DOF full vehicle model was estimated using an adaptive limited memory unscented Kalman filter (ALM-UKF).
Online, vehicle state is estimated using a factor graph-based optimization framework with GPS and IMU data and a cost map of the terrain, similar to a traversability grid, is generated from monocular camera images for use in a stochastic MPC framework.

\subsection{Moment of Inertia Estimation with Bifilar Pendulum}

\begin{figure}
  \centering
  \begin{subfigure}{0.49\textwidth}
    \begin{tikzpicture}[
    string/.style={draw=gray, very thick},
    dash/.style={draw=gray, densely dashed, very thick},
]

\def\scale{3.0}
\def\wireLength{2.0}
\def\mountDist{1.0}
\tikzstyle{hang}=[circle,draw=black,fill=black,inner sep=0pt,minimum size=2pt,line width=2pt, very thick]

\begin{scope}[scale=\scale]



%



  \begin{scope}[xshift=0.25cm,yshift=-1.5cm]
    \input{robot};
  \end{scope}

  \node at (cg) [below = 0.3mm of cg] {CG};

  \node[hang, left = \mountDist of cg] (mountLeft) {};
  \node[hang, right = \mountDist of cg] (mountRight) {};

  \node[above = \wireLength of cg] (anchorMid) {};
  \node[hang, above = \wireLength of mountLeft] (anchorLeft) {};
  \node[hang, above = \wireLength of mountRight] (anchorRight) {};

  \draw (anchorLeft) ++(-1,0) -- (anchorRight) -- ++(1,0);
  \draw[very thick] (anchorLeft) -- (mountLeft);
  \draw[very thick] (anchorRight) -- ++(mountRight);
 
  \draw[dashed] (mountLeft) -- (cg) -- (mountRight);
  
  \draw [decorate,decoration={brace,amplitude=6pt},xshift=-4pt,yshift=0pt]
(anchorRight) -- ++(mountRight) node [black,midway,xshift=0.6cm] {$d$};

\draw [decorate,decoration={brace,amplitude=6pt},xshift=-4pt,yshift=4pt]
(mountLeft) -- (cg) node [black,midway,yshift=0.4cm] {$b$};

\draw [decorate,decoration={brace,amplitude=6pt},xshift=-4pt,yshift=4pt]
(cg) -- (mountRight) node [black,midway,yshift=0.4cm] {$b$};

\end{scope}
\end{tikzpicture}
    \caption{}
    \label{fig:bifilar}
  \end{subfigure}
  \begin{subfigure}{0.47\textwidth}
    \includegraphics[width=\textwidth, angle=180]{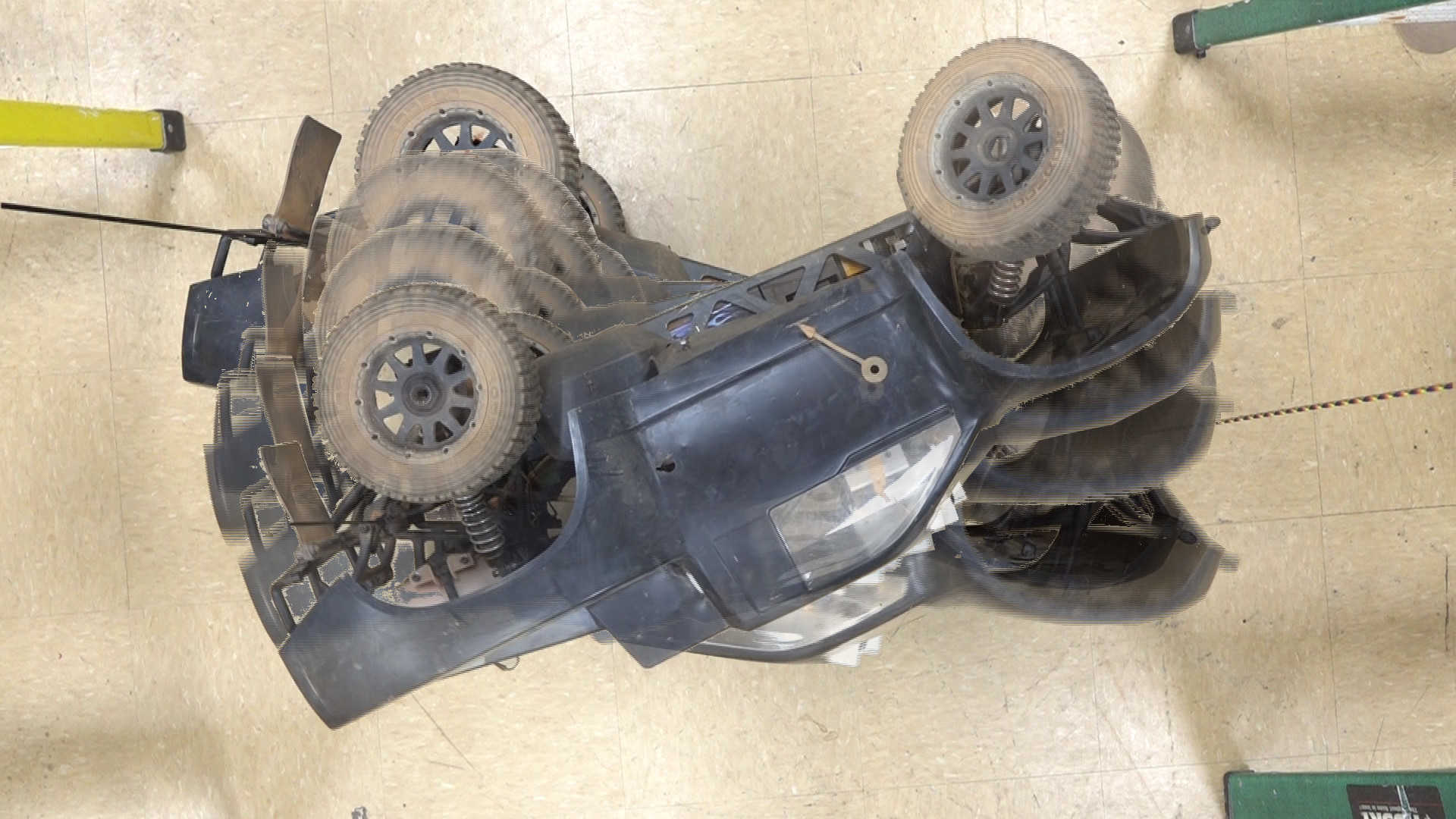}
    \caption{}
    \label{fig:bifilarComposite}
  \end{subfigure}
  \caption{Bifilar pendulum setup for experimental determination of moments of inertia (MOI) with AutoRally. (\subref{fig:bifilar}) Example setup with labeled support strings of length $d$, attached equidistant from the center of gravity, CG, for computing the yaw MOI, $I_{\psi}$. (\subref{fig:bifilarComposite}) Composite image of 1/2 oscillation period for computing the pitch MOI, $I_{\theta}$.}
  \label{fig:moi}
\end{figure}

The MOI of the platform are more difficult to determine than other parameters, such as the mass of the vehicle, but are important variables for physics-based controllers. Modern CAD software can automatically compute MOI if an accurate model exists. No complete CAD model of AutoRally is available, so the platform MOI cannot be determined with these tools. Methods also exist to compute MOI by precomputation~\cite{Doniselli2003, Gobbi2011}, or online estimation~\cite{Rozyn2010} in cases where the full model is unknown or changing. An extensive survey of popular methods for experimentally determining MOI is presented in \cite{Genta1994}. Many methods rely on custom calibration rigs that are time consuming, expensive, and difficult to build. In this work, we computed the necessary MOI experimentally using the bifilar pendulum method~\cite{Genta1994}. As shown in Figure~\ref{fig:bifilar}, two fixed, parallel cables are attached equidistant from the center of gravity of the body to isolate the desired calibration axis. The body is rotated by a small angle around the desired axis, then released and let freely oscillate. Given the dimensions of the test rig and known dimensions and weights of the robot, the period of a free oscillation after an excitation determines the MOI about that axis. 
The equation for the moment of inertia about a single axis using the bifilar pendulum method from \cite{Genta1994} is
\begin{equation}
\label{eq:moiAxis}
I=m\frac{gR_{\mathrm{1}}R_{\mathrm{2}}}{4{\pi}^2h}T^2,
\end{equation}
where $m$ is the platform mass, including wheels, $T$ is the oscillation period, $h$ is the distance of the calibration object from the support taking into account non-vertical support wires, and $R_{1}$ and $R_{2}$ are the distances from the center of gravity to the support wire attachment. The change in height from the mounting location as the vehicle is rotated is assumed zero when the angle of rotation is small. Our setup is simplified by using parallel support wires, $R_{1}$ = $R_{2}$ = $b$ so that
\begin{equation}
h = \sqrt{d^2-(R_{\mathrm{1}}-R_{\mathrm{2}})^2} = d,
\end{equation}
which makes (\ref{eq:moiAxis}) for our setup
\begin{equation}
I=m\frac{gb^2}{4{\pi}^2d}T^2.
\label{eq:bifilar}
\end{equation}
\subsection{Single-Track Vehicle Model}\label{subsec:VehicleModel}
\begin{figure}[!htbp]
	\centering
	\includegraphics[width=\textwidth]{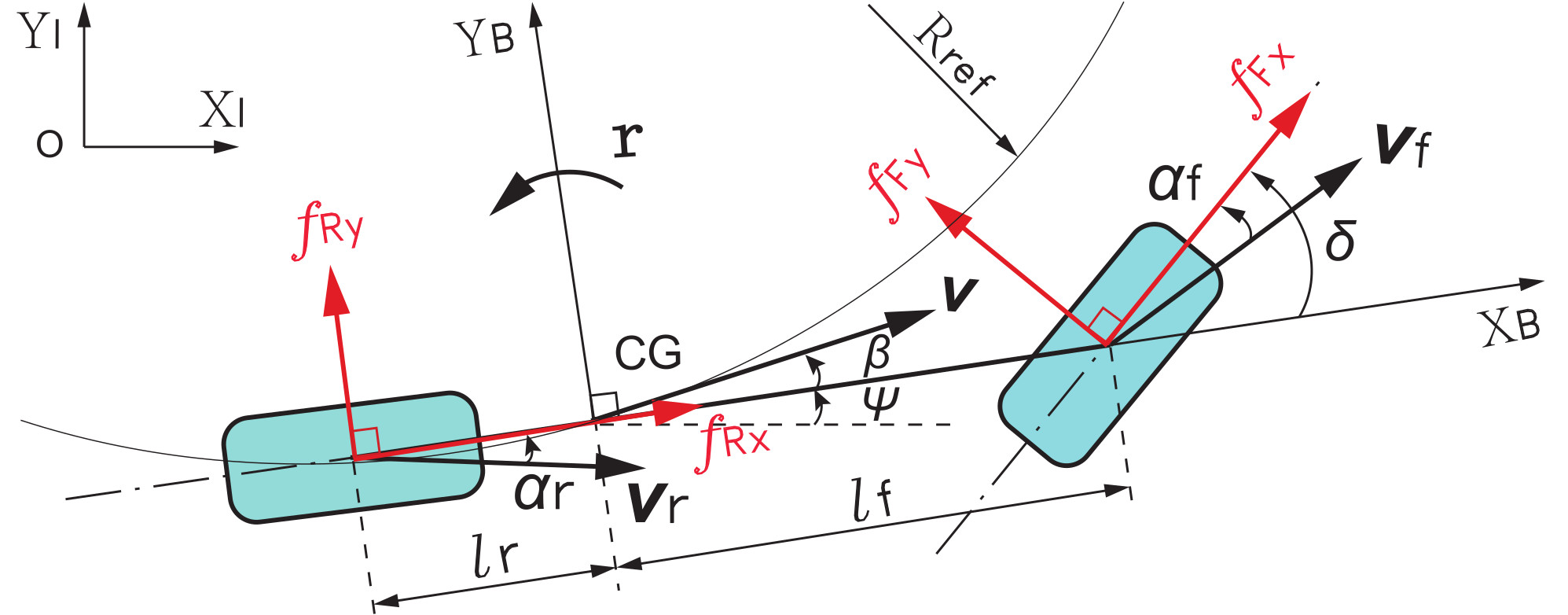}
	\caption{Single track vehicle model. This model includes the longitudinal and lateral displacement, as well as the yaw motion of the vehicle. We use $\rm{X}_{\rm I}-\rm{o}-\rm{Y}_{\rm I}$ and $X_{\rm B}-\rm{CG}-Y_{\rm B}$, where CG is the center of gravity, to denote the inertial frame fixed on the ground and the body frame fixed on the vehicle, respectively.}
	\label{fig:singleTrack}
\end{figure}

Three different vehicles models of increasing fidelity were used to test and compare the results. Below, we briefly summarize each model. The performance of each model against experimental data is given in the~\nameref{sec:expResults} section. First, the single-track vehicle model~\cite{Velenis2010,Lundahl2011,Burhaumudin2012}
used to model the AutoRally vehicle is described.
It takes into consideration the longitudinal and lateral displacement, as well as the yaw motion of the vehicle, as shown in Figure~\ref{fig:singleTrack}. \par
We use $X_{\rm I}-O-Y_{\rm I}$ and $X_{\rm B}-CG-Y_{\rm B}$, where $CG$ is the center of gravity, to denote the inertial frame fixed on the ground and the body frame fixed on the vehicle, respectively.
The equations of motion of the model can be expressed in a body-fixed frame with the origin at the CG from \cite{Velenis2010}
\begin{subequations}
	\begin{align}
	\dot{V}_{\rm x} &= ( f_{\rm Fx}\cos{\delta}-f_{\rm Fy}\sin{\delta}+f_{\rm Rx} )/m+V_{\rm y}\dot{\psi}, \label{eqn:SingleTrack1}\\
	\dot{V}_{\rm y} &= ( f_{\rm Fx}\sin{\delta}+f_{\rm Fy}\cos{\delta}+f_{\rm Ry} )/m-V_{\rm x}\dot{\psi}, \label{eqn:SingleTrack2} \\
	\dot{r} &= \big( ( f_{\rm Fy}\cos{\delta}+f_{\rm Fx}\sin{\delta})\ell_{\rm f}-f_{\rm Ry}\ell_{\rm r}  \big)/I_{\rm z}, \label{eqn:SingleTrack3}
	\end{align}
\end{subequations}
where $V_{\rm x}$ and $V_{\rm y}$ are the components of $V$ along the $X_{\rm B}$ and $Y_{\rm B}$ directions, respectively; $m$ is the total mass, and $I_{\rm z}$ is the moment of inertia of the vehicle about the vertical axis. $f_{ij}$ ($i=F, R$ and $j=x, y$) denote the longitudinal and lateral friction forces at the front and rear wheels, $\psi$ denotes the yaw angle, and $\delta$ is the steering angle of the front wheel.

\subsection{Double-Track Model}

The double-track model takes into consideration the longitudinal, lateral and yaw motion of the vehicle, but considers the load difference between the left and right wheels arising from the lateral load transfer.

We use $f_{i,j,k}$ ($i=L, R$, $j= L, R$ and $k=x, y$) to denote the longitudinal or lateral friction force for each wheel, respectively. The vehicle's equations of motion are then
\begin{subequations}
	\begin{align}
		\dot{V}_{\rm x} &= \big( (f_{\rm LFx}+f_{\rm RFx})\cos{\delta}-(f_{\rm LFy}+f_{\rm RFy})\sin{\delta}+f_{\rm LRx}+f_{\rm RRx} \big)/m+V_{\rm y}\dot{\psi}, \label{eqn:DoubleTrack1}  \\
		\dot{V}_{\rm y} &= \big( (f_{\rm LFx}+f_{\rm RFx})\sin{\delta}+(f_{\rm LFy}+f_{\rm RFy})\cos{\delta}+f_{\rm LRy}+f_{\rm RRy} \big)/m-V_{\rm x}\dot{\psi}, \label{eqn:DoubleTrack2} \\
		\dot{r} &= \Big( \big( (f_{\rm LFy}+f_{\rm RFy})\cos{\delta}+(f_{\rm LFx}+f_{\rm RFx})\sin{\delta}\big)\ell_{\rm f}-(f_{\rm LRy}+f_{\rm RRy})\ell_{\rm r}  \Big)/I_{\rm z}. \label{eqn:DoubleTrack3}
	\end{align}	
\end{subequations}

\subsection{Full Vehicle Model}

\begin{figure}[!htbp]
	\centering
	\includegraphics[width=\textwidth]{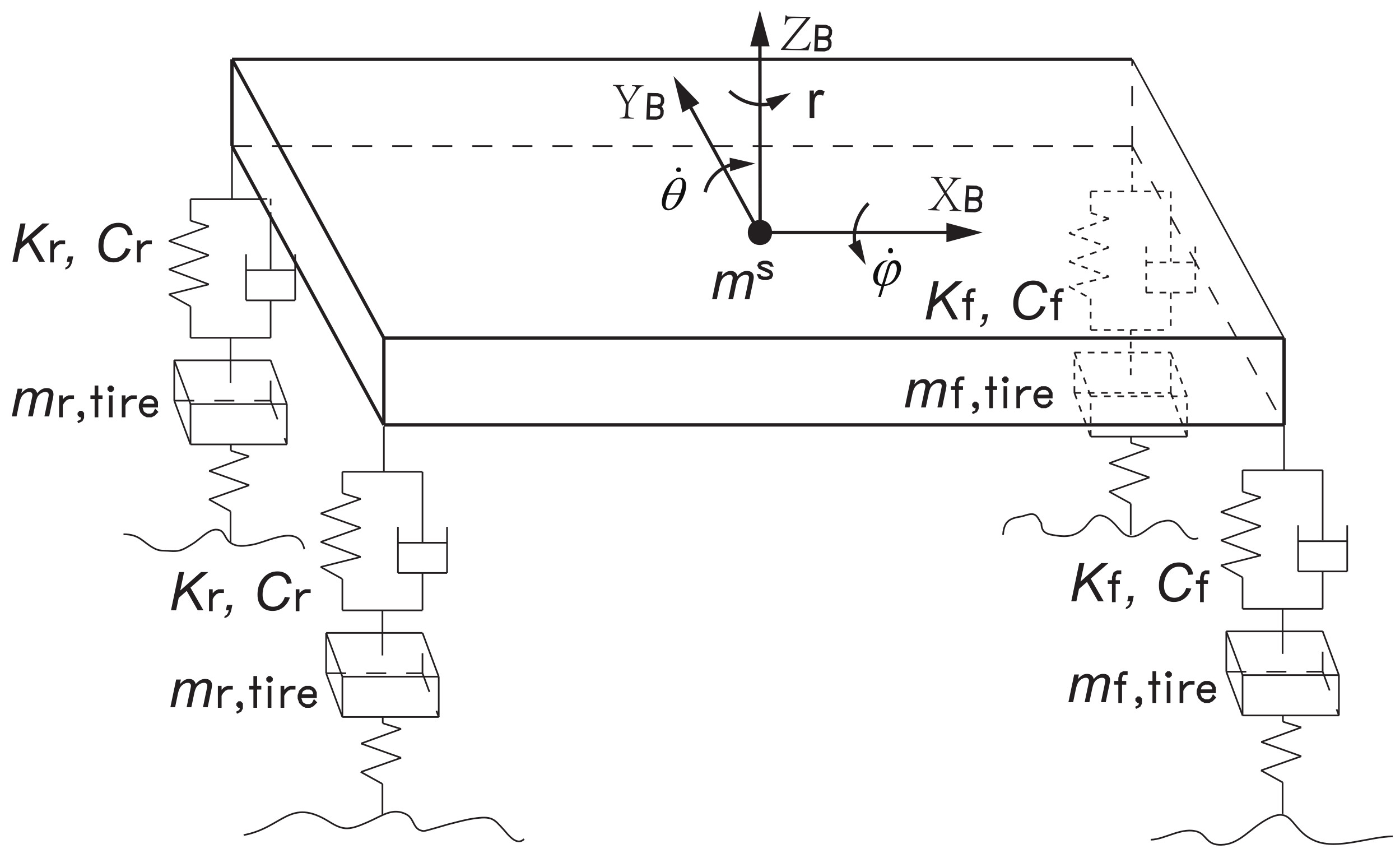}
	\caption{Riding model. $K_i$ and $C_i$ ($i={\rm f, r}$) denote the spring stiffness and the damping coefficient of the suspension system related to each wheel, $m_{i,\rm tire}$ ($i={\rm f, r}$) denotes the mass of the front and rear tire, respectively, $m^{\rm s}$ is the sprung mass, and $\dot{\varphi}$ and $\dot{\theta}$ are the rolling and pitching rate, respectively.}
	\label{11RidingModel}
\end{figure}

\begin{figure}[!htbp]
	\centering
	\includegraphics[width=\textwidth]{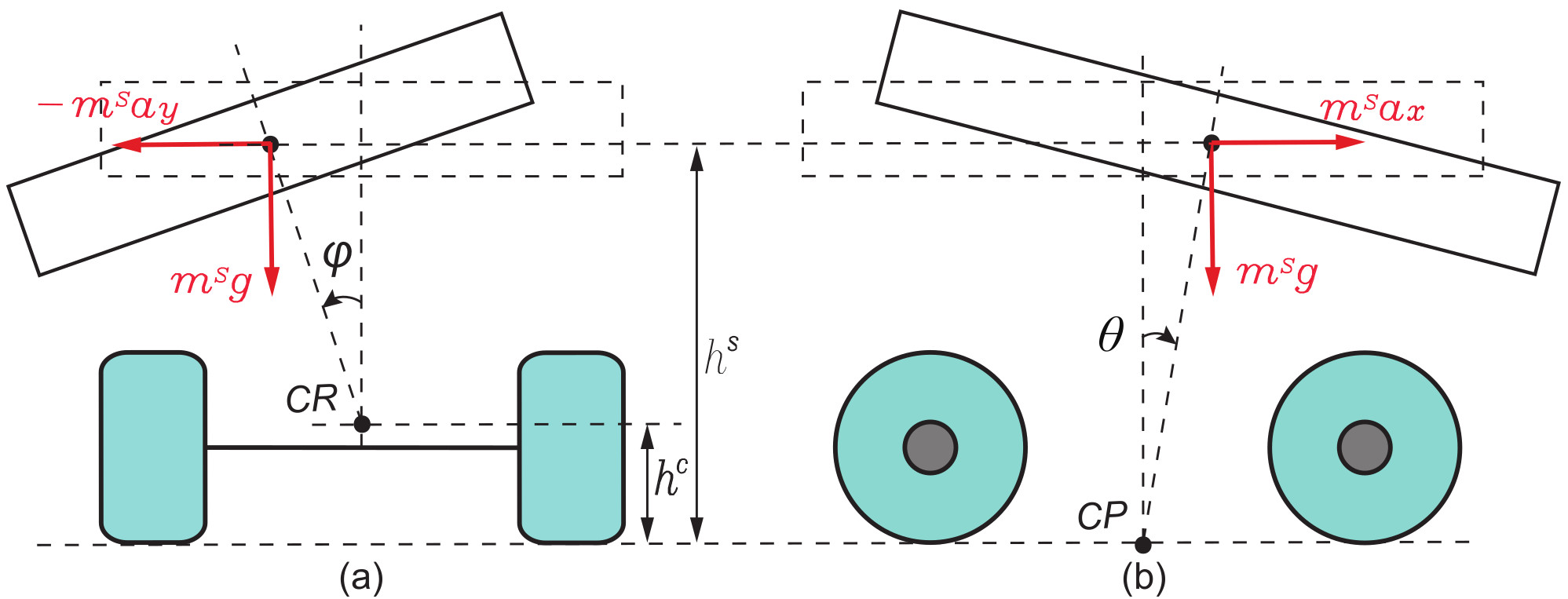}
	\caption{Rolling and pitching model. (a) Rolling motion arises from the lateral acceleration and the gravity center offset from the rolling center. The parameters $h^{\rm s}$ and $h^{\rm c}$ are the heights of the sprung mass center and CR is the rolling center, respectively, and $\varphi$ is the roll. (b) Pitching motion arises from the longitudinal acceleration and the gravity center offset from CP, the pitching center, that is assumed to be on the ground, $\theta$ is the pitch.}
	\label{12RollPitch}
\end{figure}

The full vehicle model considers the dynamics of the sprung and unsprung mass of the vehicle separately and are derived using Newton-Euler equations for the motion of rigid body systems.
The equations of motion for the total mass are the same as (\ref{eqn:DoubleTrack1})-(\ref{eqn:DoubleTrack3}) for the double-track model.
We also take the air resistance into account and modify (\ref{eqn:DoubleTrack1}) so that
\begin{align}
\dot{V}_{\rm x} &= \big( (f_{\rm LFx}+f_{\rm RFx})\cos{\delta}-(f_{\rm LFy}+f_{\rm RFy})\sin{\delta}+f_{\rm LRx}+f_{\rm RRx} \big)/m+V_{\rm y}\dot{\psi}-C_{\rm D}\rho_{\rm air}AV_{\rm x}^2/2, \label{eqn:FullVehicle1}
\end{align}	
where $C_{\rm D}$ is the air resistance coefficient, $\rho_{air}$ is the air density, and $A$ is the frontal area of the vehicle.
The vertical translation is accounted for by a riding model as shown in Figure~\ref{11RidingModel}.
The rolling and pitching model are given in Figure~\ref{12RollPitch}.

In Figure~\ref{11RidingModel}, $K_i$ and $C_i$ ($i={\rm f, r}$) denote the spring stiffness and the damping coefficient of the suspension system related to each wheel, $m_{i,\rm tire}$ ($i={\rm f, r}$) denotes the mass of the front and rear tire, respectively, $m^{\rm s}$ is the sprung mass, and $\dot{\varphi}$ and $\dot{\theta}$ are the rolling and pitching rate, respectively. 

Figure~\ref{12RollPitch}(a) shows the rolling motion arising from the lateral acceleration and the gravity center offset from the rolling center.
The parameters $h^{\rm s}$ and $h^{\rm c}$ are the heights of the sprung mass center and the rolling center (CR), respectively.
Figure~\ref{12RollPitch}(b) shows the pitching motion arising from the longitudinal acceleration and the gravity center offset from the pitching center (CP) that is assumed to be on the ground.
The dynamical equations of the vertical, rolling and pitching motion of the sprung mass are
\begin{subequations}
	\begin{align}
	\begin{split}
    	\dot{V}^{\rm s}_{ \rm z} =&\Big( -2(K_{\rm f}+K_{\rm r})\theta-2(C_{\rm f}+C_{\rm r})V^{\rm s}_{\rm z}+2(\ell_{\rm f}K_{\rm f}-\ell_{\rm r}K_{\rm r})\varphi+2(\ell_{\rm f}C_{\rm f}-\ell_{\rm r}C_{\rm r})\dot{\theta} \Big)/m^{\rm s}, \label{eqn:FullVehicle4}  
    \end{split}
	\\
	\begin{split}
    	\ddot{\theta}=&\Big( 2(\ell_{\rm f}K_{\rm f}-\ell_{\rm r}K_{\rm r})z^{\rm s}+2(\ell_{\rm f}C_{\rm f}-\ell_{\rm r}C_{\rm r})V^{\rm s}_{\rm z}-2(\ell_{\rm f}^2K_{\rm f}+\ell_{\rm r}^2K_{\rm r})\theta \\
		&-2(\ell_{\rm f}^2C_{\rm f}+\ell_{\rm r}^2C_{\rm r})\dot{\theta}+m^{\rm s}gh^{\rm s}\sin\theta+m^{\rm s}a^{\rm s}_{\rm x}h_{s}\cos\theta \Big)/I^{\rm P}_{\rm y}, \label{eqn:FullVehicle5}
	\end{split}
    \\
    \begin{split}
		\ddot{\varphi}=&\Big( -w_{\rm f}^2K_{\rm f}\varphi/2-w_{\rm f}^2C_{\rm f}\dot{\varphi}/2-w_{\rm r}^2K_{\rm r}\varphi/2-w_{\rm r}^2C_{\rm r}\dot{\varphi}/2 \\
	&+ m^{\rm s}g(h^{\rm s}-h^{\rm c})\sin\varphi+m^{\rm s}a^{\rm s}_{\rm y}(h^{\rm s}-h^{\rm c})\cos\varphi \Big)/I^{\rm R}_{\rm x}, \label{eqn:FullVehicle6}
	\end{split}
    \end{align}	
\end{subequations}
where  $w_{\rm i}$ ($i={\rm f, r}$) denote the front and rear track, respectively; $a^{\rm s}_{\rm x}$ and $a^{\rm s}_{\rm y}$ are the longitudinal and lateral acceleration of the sprung mass center in the body-fixed frame, and $I^{\rm R}_{\rm x}$ and $I^{\rm P}_{\rm y}$ are the MOI of the sprung mass about the rolling axis and the pitching axis, respectively. Note that in this derivation, we make the small angle assumption for $\phi$ and $\theta$, which is a standard assumption.

\subsection{Tire Force Model}

\begin{figure}
\centering
    \includegraphics[width=\textwidth]{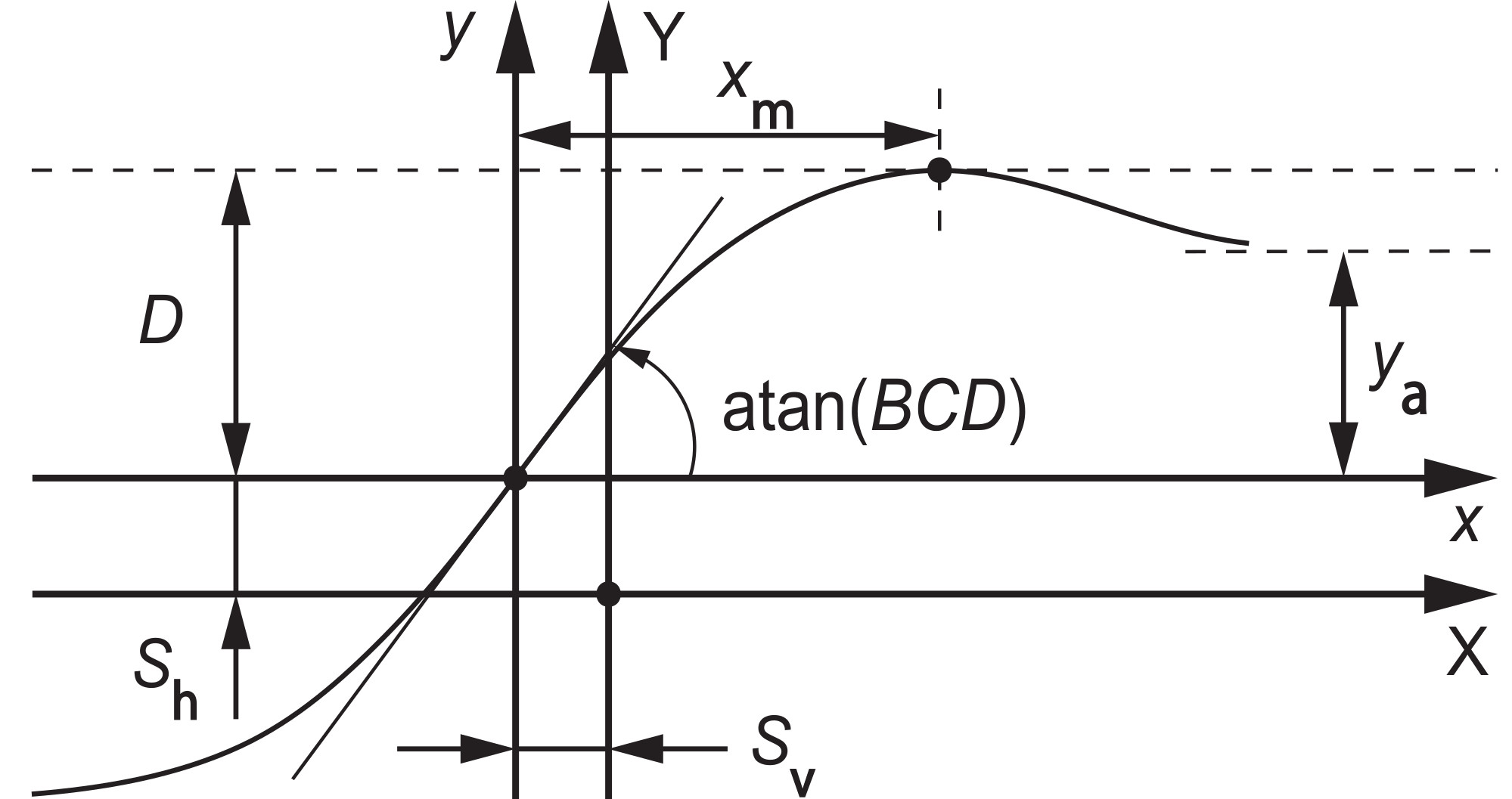}
    \caption{Pacejka's Magic Formula tire model. The graph shows slip angle (X-axis) vs. normalized lateral force (Y-axis). Parameters $B, C, D, E$ are the stiffness, shape, peak and curvature factors, respectively. $S_{\rm h}$ is the horizontal shift and $S_{\rm v}$ is the vertical shift.}
    \label{fig:magicTire}
\end{figure}

Tire models are needed to model the complex interactions of the vehicle tires with the ground, and are especially important in the driving regime of this work where the vehicle frequently slides on a dirt surface and is almost always accelerating.
One of the most common tire models, which we use here, is based on Pacejka's ``magic tire formula" (MF). The important aspect of this model is its ability to capture the tire saturation the coupling between the lateral and longitudinal axes. According to this model, the generated tire force depends on the tire slip. Tire slip is defined by the nondimensional relative velocity, not angle as is sometimes used, of each tire with respect to the road, along the longitudinal and lateral directions
\begin{align}
s_{ijx}=\frac{V_{ijx}-\omega_{ijx}R_{j}}{\omega_{ijx}R_{j}}, \qquad s_{ijy}=\frac{V_{ijy}}{\omega_{ijx}R_{j}},
\end{align}	
where $i=L, R$ correspond to left and right side, and $j=F, R$ correspond to front and rear, and
$V_{ijk}$ ($k=x ,y$) is the tire frame component of the vehicle velocity of each tire. The total slip of each tire is defined by $s_{ij}=\sqrt{s_{ijx}^2+s_{ijy}^2}$ and the total friction coefficient related to each tire is calculated using MF from~\cite{Velenis2010}
\begin{align} \label{eqn:MF}
\mu_{ij}=D\sin\Big(C{\rm atan}\Big(BS_{\rm E}-E\big(BS_{\rm E}-{\rm atan}S_{\rm E} \big)\Big)\Big)+S_{\rm v},
\end{align}	
where $B, C, D, E$ are the stiffness, shape, peak and curvature factors, respectively; $S_{\rm E}=s_{ij}-S_{\rm h}$, where $S_{\rm h}$ is the horizontal shift. $S_{\rm v}$ is the vertical shift.
%
The tire friction force components are
\begingroup\makeatletter\check@mathfonts
\def\maketag@@@#1{\hbox{\m@th\normalsize\normalfont#1}}
\begin{align} \label{eqn:TireForces}
f_{ijk}=-\frac{s_{ijk}}{s_{ij}}\mu_{ij}f_{ijz}, \quad i=L, R; ~ j=F, R; ~ k=x, y,
\end{align}
\endgroup
where $f_{ijz}$ is the normal load on the corresponding tire and can be calculated from \cite{Velenis2010}.

\subsection{Unscented Kalman Filter} \label{sec:UKF}


In order to find the unknown vehicle parameters, including the parameters for the tire force model, a JS-UKF was used.
A JS-UKF includes the unknown parameters in the original state vector and estimates the new augmented state.
In this framework, the state and the noise are assumed to be Gaussian random variables.
Recall that for a system
\begin{align} \label{eqn:system}
\ x_{k+1} = f(x_k,u_k)+w_k,  \quad \ y_k = h(x_k, u_k)+ v_k,
\end{align}
where $w_k\sim \mathcal{N}(q, Q)$ and $v_k\sim \mathcal{N}(r, R)$ are Gaussian process and measurement noise, respectively, the extended Kalman Filter (EKF) propagates the Gaussian random variable $x_k$ by linearizing the nonlinear state transition (observation) function $f:\mathbb{R}^n\times\mathcal{U}\mapsto \mathbb{R}^n$ ($h:\mathbb{R}^n\times\mathcal{U}\mapsto \mathbb{R}^m$) with the Jacobian matrix at each time step $k$ \cite{Hong2013}.
Instead of an EKF, in this work we used an UKF filter since: a) the UKF propagates the Gaussian random variable through a nonlinear function more accurately than the EKF;
and b) The UKF avoids calculating the Jacobians that may be too cumbersome for highly nonlinear systems.

\subsubsection{Adaptive Limited Memory Unscented Kalman Filter}

A UKF is based on the unscented transformation (UT), and avoids calculating the Jacobian matrices at each time step. Assuming an $L$-dimensional Gaussian random variable $x$ with mean $\hat{x}$ and covariance $P_x$,
to calculate the statistics of $y=g(x)$, one selects $2L+1$ discrete sample points $\{\mathcal{X}_i\}_{i=0}^{2L}$ which are propagated through the system dynamics.
The UKF redefines the state vector as $x^a_k=[x^\mathrm{T}_k, w^\mathrm{T}_k, v^\mathrm{T}_k ]^\mathrm{T}$, which concatenates the original state and noise variables, and then estimates $x^a_k$ recursively \cite{Wan2000}.

The ALM-UKF, introduced in~\cite{you2017vehicle}, is an estimation algorithm for nonlinear systems that builds on previous works on UKF to simultaneously estimate the system state, model parameters, and the Kalman filter hyperparameters related to the noise. 
First, recall that the adaptive Kalman filter algorithm~\cite{Myers1976} adjusts the mean and the covariance of the noise online, which is expected to compensate for time-varying modeling errors.
Define the set of unknown time-varying hyperparameters for the Kalman filter 
corresponding to the noise statistics at the $i^{\rm th}$ time step, as
$\mathcal{S}_i\triangleq\{ q_i, Q_i, r_i, R_i\}$.
$\mathcal{S}_i$ is estimated simultaneously with the system state and parameters.
Since an optimal estimator for $\mathcal{S}_i$ is not available, and many suboptimal schemes are either too restrictive for nonlinear applications or too computationally demanding \cite{Jazwinski2007,han2009adaptive}, this work adopts the adaptive limited memory algorithm in \cite{Myers1976}, with the following two extensions: a) the algorithm is developed for a nonlinear application (i.e., UKF); b) the unknown parameters of the system are estimated along with the state, instead of just the system state.
In the following, it is assumed that $\mathcal{S}_i$ is constant and is denoted by $\mathcal{S}=\{ q, Q, r, R\}$.

For the observation noise statistics $r$ and $R$, we consider the nonlinear observation at time $k$, which is $y_k=h(x_k,u_k)+v_k$. 
Since the true value of $x_k$ is unknown, $v_k$ is approximated by
\begin{align} \label{eqn:r_k1}
r_k= y_k-\hat{h}(x_k,u_k),
\end{align}
where $r_k$ represents a sample of the observation noise $v$ at time $k$, and
\begin{align} \label{eqn:hat_h}
\hat{h}(x_k,u_k)=\sum\limits_{i=0}^{2L}W_i^{(m)}h(\mathcal{X}^x_{i,k}, u_k)\triangleq \hat{h}_k.
\end{align}
We define a new random variable $\xi \sim \mathcal{N}(r, C_r)$, and assume that there are $N$ samples $r_k$ ($k=1,\ldots, N$), such that the $r_k$'s are $N$ empirical measurements for $\xi$. 
An unbiased estimator for $r$ can be the sample mean
\begin{align} \label{eqn:mean_r}
\ \hat{r}=\frac{1}{N}\sum\limits_{k=1}^{N}r_k,
\end{align}
where the term ``unbiased'' implies that $\mathbb{E}[\hat{r}]=\mathbb{E}[\xi]= r$. An unbiased estimator for the covariance of $\xi$ is
\begin{align} \label{eqn:Cov_r}
\ \hat{C}_r=\frac{1}{N-1}\sum\limits_{k=1}^{N}(r_k-\hat{r})(r_k-\hat{r})^\mathrm{T},
\end{align}
where the term ``unbiased'' implies $\mathbb{E}[\hat{C}_r]=\mathbb{E}[(\xi-r)(\xi-r)^\mathrm{T}]$.
Because $y_k=h(x_k,u_k)+v_k$, (\ref{eqn:r_k1}) becomes
\begin{align} \label{eqn:r_k2}
r_k=h(x_k,u_k)-\hat{h}_k+v_k.
\end{align}
We can therefore calculate the covariance of $\xi$
\begin{align}
\mathbb{E}[(\xi-r)(\xi-r)^\mathrm{T}]=\frac{1}{N}\sum\limits_{k=1}^{N}\mathbb{E}[(r_k-r)(r_k-r)^\mathrm{T}]  \nonumber
\end{align}
\begin{flalign}
\hspace{-5 em}
=\frac{1}{N}\sum\limits_{k=1}^{N}\mathbb{E}[(h(x_k,u_k)-\hat{h}_k+v_k-r)(h(x_k,u_k)-\hat{h}_k+v_k-r)^\mathrm{T}]  \nonumber
\end{flalign}
\begin{align} \label{eqn:2R}
\hspace{-3.8 em}
=\frac{1}{N}\sum\limits_{k=1}^{N}\big(\mathbb{E}\big[(h(x_k,u_k))(h(x_k,u_k))^\mathrm{T}\big]-(\hat{h}_k)(\hat{h}_k)^\mathrm{T}\big)+R,
\end{align}
where 
\begin{align}
	\hspace{-1 em}
\ \mathbb{E}[(h(x_k,u_k))(h(x_k,u_k))^\mathrm{T}]=\sum\limits_{i=0}^{2L}W_i^{(m)}(h(\mathcal{X}^x_{i,k}, u_k))(h(\mathcal{X}^x_{i,k}, u_k))^\mathrm{T}.
\end{align}
Note that we assume that $x_k$ and $v_k$ are independent in (\ref{eqn:2R}).
An unbiased estimate of $R$ from (\ref{eqn:Cov_r}) and (\ref{eqn:2R}) is
\begin{align} \label{eqn:hat_R}
&\hat{R}=\frac{1}{N-1}\sum\limits_{k=1}^{N}\Big( (r_k-\hat{r})(r_k-\hat{r})^\mathrm{T}-\frac{N-1}{N}\Big(\mathbb{E}[(h(x_k,u_k))(h(x_k,u_k))^\mathrm{T}]-(\hat{h}_k)(\hat{h}_k)^\mathrm{T}\Big) \Big).
\end{align}
For the process noise statistics $q$ and $Q$, we consider the nonlinear state propagation at time $k$, which is $x_k=f(x_{k-1},u_{k-1})+w_{k-1}$. Since the true values of $x_k$ and $x_{k-1}$ are unknown, $w_{k-1}$ is approximated by
\begin{align} \label{eqn:q_k1}
q_k = \hat{x}_k-\hat{f}(x_{k-1},u_{k-1}), 
\end{align}
where $q_k$ represents a sample of the process noise $w$ at time step $k-1$, and
\begin{align} \label{eqn:hat_f}
\ \hat{f}(x_{k-1},u_{k-1}) = \sum\limits_{i=0}^{2L}W_i^{(m)}f(\mathcal{X}^x_{i,k-1},u_{k-1})\triangleq \hat{f}_{k-1}.
\end{align}
We define a new random variable $\zeta \sim \mathcal{N}(q, C_q)$, and assume that there are $M$ samples $q_k$ ($k=1,\ldots, M$), where the $q_k$'s are $M$ empirical measurements for $\zeta$. An unbiased estimator for the mean value of $\zeta$ is the sample mean
\begin{align} \label{eqn:mean_q}
\ \hat{q}=\frac{1}{M}\sum\limits_{k=1}^{M}q_k.
\end{align}
An unbiased estimator for the covariance of $\zeta$ is
\begin{align} \label{eqn:Cov_q}
\ \hat{C}_q=\frac{1}{M-1}\sum\limits_{k=1}^{M}(q_k-\hat{q})(q_k-\hat{q})^\mathrm{T},
\end{align}
such that $\mathbb{E}[\hat{C}_q]=\mathbb{E}[(\zeta-q)(\zeta-q)^\mathrm{T}]$. We then calculate the covariance $Q$
\begin{flalign}
	\mathbb{E}[(w_{k-1}-q)(w_{k-1}-q)^\mathrm{T}]=\mathbb{E}[(w_{k-1}-q_k+q_k-q)(w_{k-1}-q_k+q_k-q)^\mathrm{T}]  \nonumber
\end{flalign}
\begin{align}
	\hspace{-0.1 em}
	=\frac{1}{M}\sum\limits_{k=1}^{M}\mathbb{E}\Big[\Big( (x_k-\hat{x}_k)-(f(x_{k-1},u_{k-1})-\hat{f}_{k-1}) +(q_k-q) \Big)\Big(q_k-q\Big)^\mathrm{T}\Big] \nonumber
\end{align}
\begin{flalign} \label{eqn:2Q}
	\hspace{-0.9 em}
	=\frac{1}{M}\sum\limits_{k=1}^{M}\Big(\mathbb{E}[(f(x_{k-1},u_{k-1}))(f(x_{k-1},u_{k-1}))^\mathrm{T}]-(\hat{f}_{k-1})(\hat{f}_{k-1})^\mathrm{T}-P_k\Big)+\mathbb{E}[\hat{C}_q],  \nonumber\\
\end{flalign}
where
\begin{flalign}
\hspace{-1.3 em}
\ \mathbb{E}[(f(x_{k-1},u_{k-1}))(f(x_{k-1},u_{k-1}))^\mathrm{T}] = \sum\limits_{i=0}^{2L}W_i^{(m)}(f(\mathcal{X}^x_{i,k-1},u_{k-1}))(f(\mathcal{X}^x_{i,k-1},u_{k-1}))^\mathrm{T}.
\end{flalign}
Then $Q$ can be estimated without bias following the equations (\ref{eqn:Cov_q})-(\ref{eqn:2Q}),
\begin{align} \label{eqn:hat_Q}
	\hat{Q}= &\frac{1}{M-1}\sum\limits_{k=1}^{M}\Big( (q_k-\hat{q})(q_k-\hat{q})^\mathrm{T}+\frac{M-1}{M}\Big(\mathbb{E}[(f(x_{k-1}, u_{k-1}))(f(x_{k-1},u_{k-1}))^\mathrm{T}]\nonumber \\ 
&-(\hat{f}_{k-1})(\hat{f}_{k-1})^\mathrm{T}-P_k\Big) \Big). 
\end{align}
Equations (\ref{eqn:mean_r}), (\ref{eqn:hat_R}), (\ref{eqn:mean_q}) and (\ref{eqn:hat_Q}) provide unbiased estimates for $r$, $R$, $q$ and $Q$, which are based on $N$ observation noise samples and $M$ process noise samples, respectively. 
All samples $r_k$ and $q_k$ are assumed to be statistically independent and identically distributed. 
The summarized algorithm of the ALM-UKF based on equations (\ref{eqn:r_k1})-(\ref{eqn:hat_Q}) is presented in~\cite{you2017vehicle}, where $\alpha, \beta, \kappa$ and $\lambda$ are the UT parameters \cite{Wan2000}.

\subsubsection{Parameter Estimation}

For the system given in (\ref{eqn:system}), we introduce the following dynamics for the parameter vector $p$,
\begin{align}  \label{eqn:P}
	\ p_{k+1} &= p_k+w^{\rm p}_k,
\end{align}
where $w^{\rm p}_k\sim \mathcal{N}(q^{\rm p}, Q^{\rm p})$ is Gaussian process noise. We define the augmented state as $x^{\rm a}=[x^\mathrm{T}, p^\mathrm{T}]^\mathrm{T}$.
It then follows from (\ref{eqn:system}) and (\ref{eqn:P}) that
\begin{align} \label{eqn:System}
\ x^{\rm a}_{k+1} = F(x^{\rm a}_k,u_k)+w^{\rm a}_k, \quad 	\ y_k = H(x^{\rm a}_k, u_k)+ v_k,
\end{align}
where $w^{\rm a}_k=[w^\mathrm{T}_k, (w^{\rm p}_k)^\mathrm{T}]^\mathrm{T}$.


It is noticed that the matrices $\hat{R}$ and $\hat{Q}$ in (\ref{eqn:hat_R}) and (\ref{eqn:hat_Q}) may become negative definite during the process of the implementation (this is also mentioned in \cite{Myers1976}). 
In this work we calculate the nearest positive definite matrices of $\hat{R}$ or $\hat{Q}$ when negative eigenvalues of $\hat{R}$ or $\hat{Q}$ are observed, such that a symmetric positive definite matrix nearest to $\hat{R}$ or $\hat{Q}$ in terms of the Frobenius norm can be obtained~\cite{Qi2015}.

It is worth mentioning that the artificial Gaussian process noise $w_k^{\rm p}$ in (\ref{eqn:P}) is used to change the parameter $p$ when the UKF is running.
However, if the value of $w_k^{\rm p}$ is large, the parameter $p$ will be changed by a large amount at each time step.
This condition may further cause the filter to diverge, since the parameterized vehicle model in Subsection~\nameref{subsec:VehicleModel} are sensitive to $p$ and may thus become unstable for unreasonable values of $p$.
We addressed this problem by rescaling the diagonal entries of $Q^{\rm p}$ to be some small positive values at each time step.
Other discussions on the numerical instability problems of the UKF can be found in \cite{Myers1976,Qi2015}.

\subsection{Online State Estimation}
\label{sec:online_state_estimation}

Accurate state estimation is required for the controllers to run reliably online. 
Many control methods assume accurate knowledge of the state of the system. 
In general, the more accurate and high-rate this information is, the better.  
Some states, such as the wheel speeds, can be directly measured. 
However,  position, orientation, and velocity estimate are needed for control and planning for the AutoRally platform.
This information, in general, cannot be measured with a single sensor and some form of sensor fusion estimation is needed.

GPS is inherently low rate and lacks orientation information. IMU measurements are relatively high rate but do not directly provide heading or linear velocity information. By combining the time-synchronized signals from these two sensors, a very accurate and high rate estimate of position, velocity, and orientation can be obtained.  This state information is sufficient for many advanced control systems.

Previously, many systems have had success with sensor fusion using methods such the EKF, or the UKF, as presented in the previous section.  
Filtering methods are limited by the fact that one throws away all previous state and measurement information. If, instead, the system is modeled as a set of hidden states (in our case, the position and velocity of the vehicle), with sensor measurements providing probabilistic information about these states, more accurate estimates, via smoothing, can be achieved. 

Factor graphs combined with advanced inference algorithms such as incremental smoothing and mapping 2 (iSAM2)~\cite{isam2} allow smoothing over many types of measurements, while retaining the ability to re-linearize previous information. 
This reduces many of the problems found in the Kalman filter with states or measurements that are not approximately linear in the measurement time frame.

\begin{figure}
  \begin{center}
  \includegraphics[width=\textwidth]{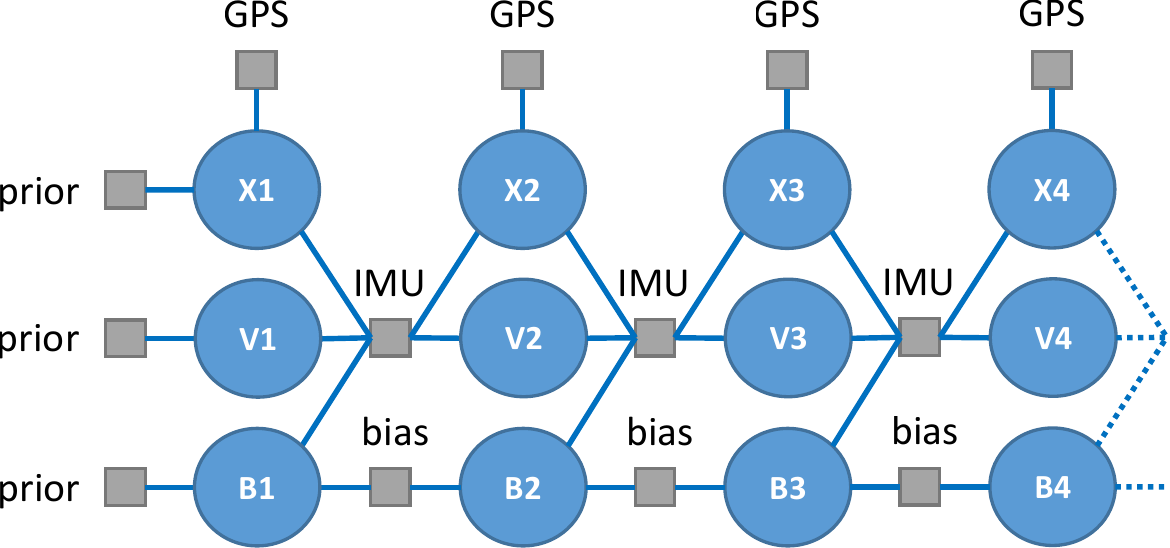}
  \end{center}
  \caption{Factor graph structure used for global positioning system (GPS) and inertial measurement unit (IMU) sensor fusion using the Georgia Tech smoothing and mapping optimization library. Circles represent states and squares represent factors.}
  \label{fig:FactorGraph}
\end{figure}

The factor graph representation of sensor fusion is a method of visualizing states and measurements as a bipartite graph (an example is shown in Figure~\ref{fig:FactorGraph}). The factor graph has two types of nodes, \emph{factor nodes} $f_i\in F$ and \emph{variable nodes} $\theta_j\in \Theta$. \emph{Edges} $e_{ij}$ always connect factor and variable nodes. Variable nodes correspond to the unmeasured quantities to be estimated.  
Factor nodes correspond to probabilistic information gained from a measurement $z_i$ about a set of variables (connected to the factor by edges). The factor graph as a whole represents the probability distribution generated by the probabilistic information encoded in the factors
\begin{equation}
p(\theta_1,\theta_2,\dots\theta_n|z_1,z_2,\dots,z_k),
\end{equation}
where $\theta_i$ is a variable that is not directly observed, and $z_j$ is a measured variable.  This function can then be factorized
\begin{equation}
\label{eq_factorization}
f(\Theta) = \prod_i{f_i(\Theta_i)},
\end{equation}
where $\Theta$ is the set of all variables in the graph and $\Theta_i$ is the set of variables connected to factor $f_i$ by an edge.  $f_i(\Theta_i)$ takes, for example, the following form for the first IMU factor in Figure~\ref{fig:FactorGraph} is
\begin{equation}
p(X_1,V_1,B_1,X_2,V_2|\omega_x,\omega_y,\omega_z,a_x,a_y,a_z),
\end{equation}
where $X_i,V_i,B_i$ are the position, velocity, and bias state variables and $\omega_n$ and $a_n$ are measured angular velocity and linear acceleration from an IMU.

Independence relationships in the measurements in $f(\Theta)$ are encoded in the edges $e_{ij}$, where each factor $f_i$ is a function of variables $\Theta_j$.  
The goal in sensor fusion is to find the variable assignment $\Theta^*$ that minimized the function $f(\Theta)$ in (\ref{eq_factorization})
\begin{equation}
\label{eq_factorgraphmax}
\Theta^* = \arg\max_\Theta f(\Theta).
\end{equation}

Each $f(\Theta)$ can be written in terms of a difference between the measured value $z_i$ and the predicted value from the measurement function $h_i(\Theta_i)$.  
By framing this in terms of log-likelihood, this maximization problem becomes
\begin{equation}
\label{eq:fglogliklihood}
\arg\max_\Theta f(\Theta) = \arg\min_{\Theta}(-\log f(\Theta))=\arg\min_{\Theta}\frac{1}{2}\sum_{i}||h_i(\Theta_i)-z_i||_{\sum_{i}}^2,
\end{equation}
where $h_i(\Theta_i)$ are the measurement functions relating a set of variables $\Theta_i$ to a sensor measurement $z_i$.  
This minimization problem can be solved with several different nonlinear minimization strategies including Gauss-Newton or Levenberg-Marquardt that iteratively linearize and solve this problem.  
Using these methods, one can create an entire graph of measurements as in Figure~\ref{fig:FactorGraph}, solve for $\Theta^*$, and this will be the maximum likelihood estimate of the variables $\Theta$ being estimated.

However, in the case of estimating the state for the AutoRally platform, we wish to solve this problem at each time step to produce a maximum likelihood estimate of the current state variables position $X_i$, velocity $V_i$ and accelerometer and gyroscope bias $B_i$.  
Re-optimizing the entire factor graph would be very inefficient, so we instead use the iSAM2 algorithm.  

\begin{figure}
    \centering
    \begin{subfigure}[b]{0.47\textwidth}
      \includegraphics[width=\textwidth]{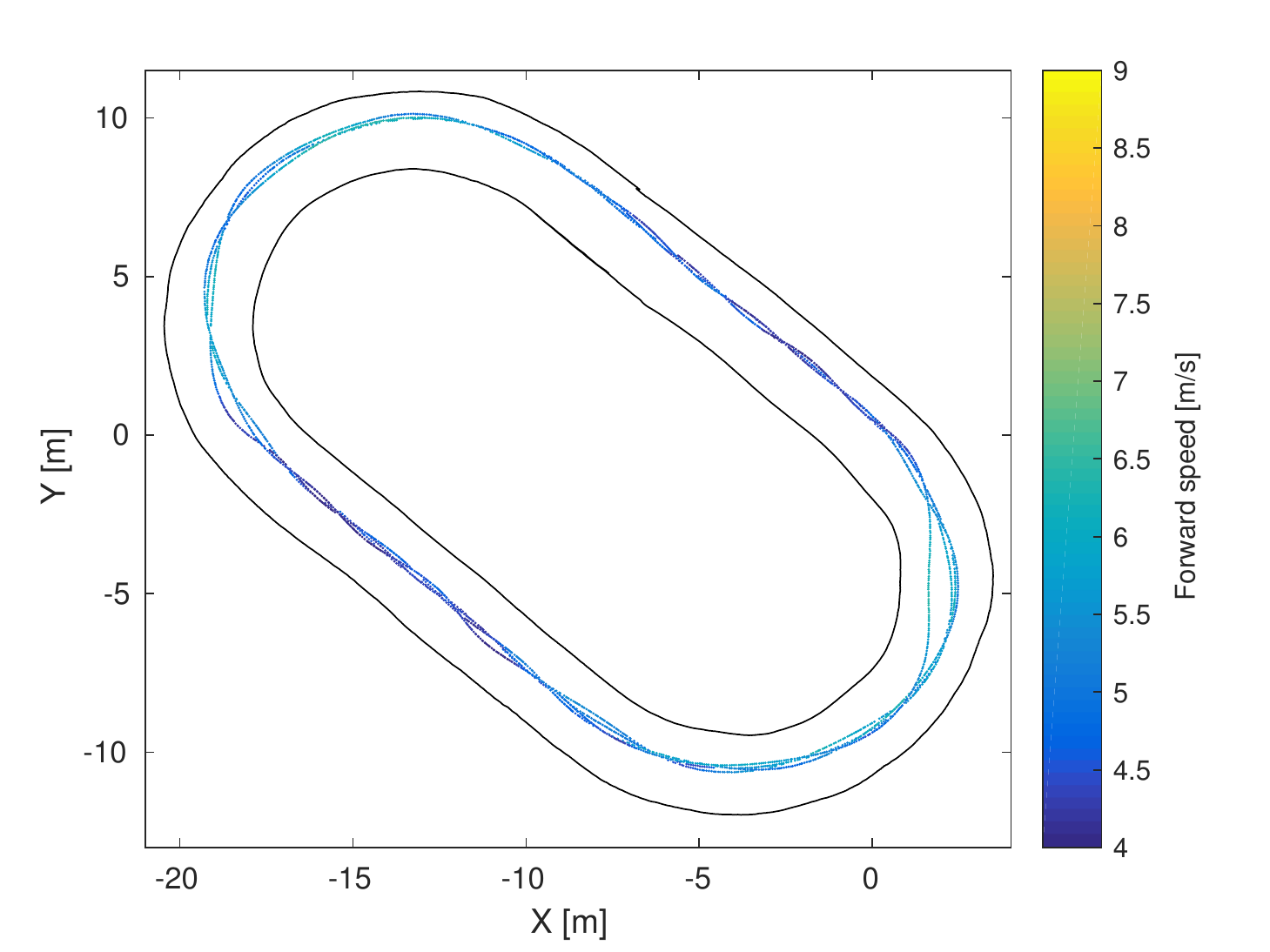}
      \caption{}
      \label{fig:pose}
    \end{subfigure}
    ~ 
    \begin{subfigure}[b]{0.47\textwidth}
      \includegraphics[width=\textwidth]{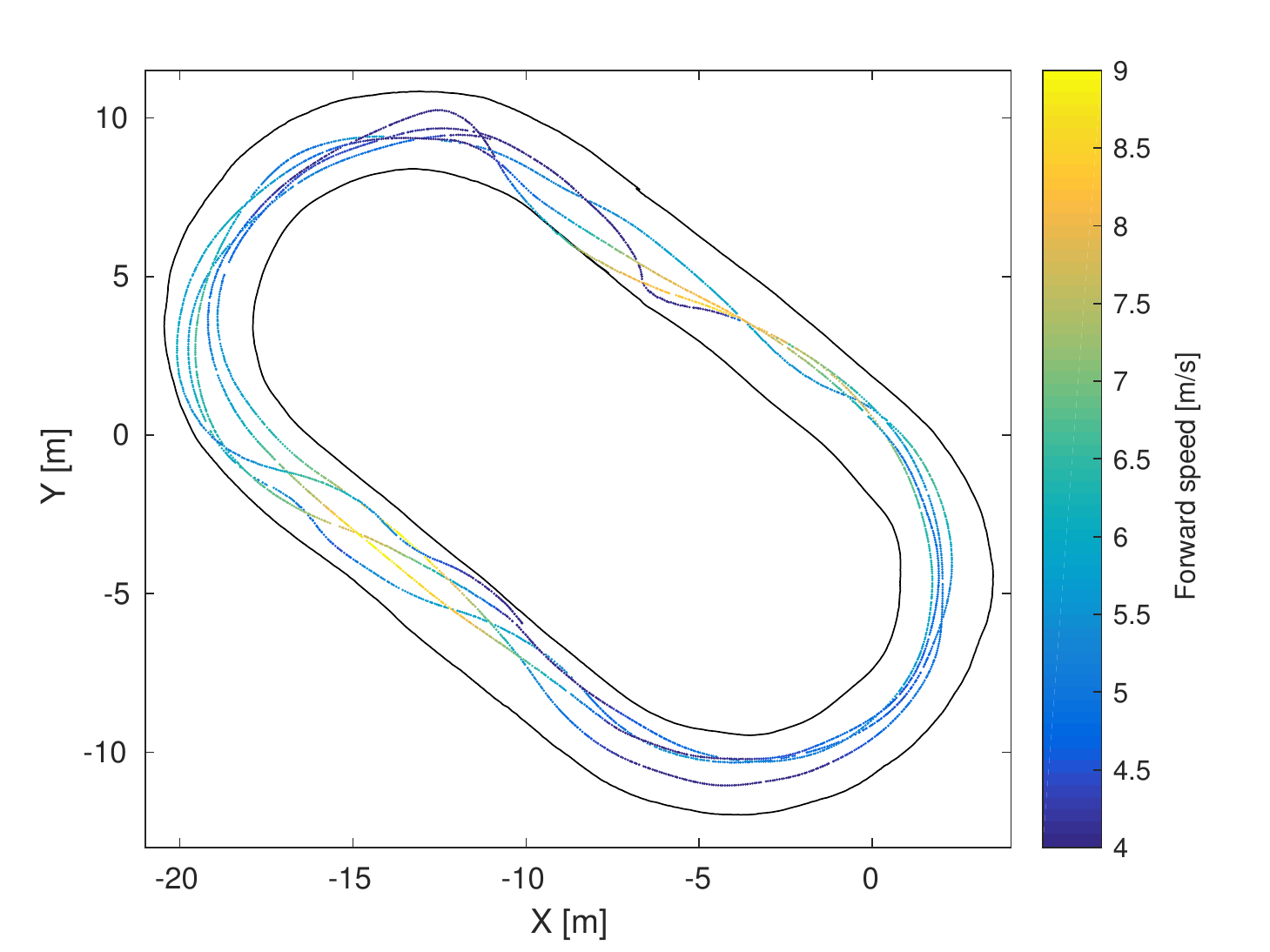}
      \caption{}
      \label{fig:pose_waypoint}
    \end{subfigure}
    \caption{Example states generated by the state estimator built with the Georgia Tech smoothing and mapping optimization library. Inputs to the state estimator are global positioning system and inertial measurement unit sensor data. Each experiment is composed of four laps of data collected at the Georgia Tech Autonomous Racing facility oval track. Track boundaries are colored black state estimates are color coded according to the speed at which the AutoRally robot was traveling. (\subref{fig:pose}) Model predictive path integral controller driving with a target speed of 6~m/s. (\subref{fig:pose_waypoint}) Manual driving for the system identification dataset.}
    \label{fig:poses}
\end{figure}

The iSAM2 algorithm is a part of the Georgia Tech smoothing and mapping (GTSAM)~\cite{dellaert2012factor,dellaert2006square} software package, which uses a factor graph representation to solve the smoothing and mapping problem iteratively. See ``\nameref{sidebar:gtsam}" for a brief discussion of GTSAM.
iSAM2 efficiently performs iterative updates to a factor graph and optimizes this new graph without re-linearizing the full problem. 
To perform this optimization, a factor graph, shown in Figure~\ref{fig:FactorGraph} is constructed with successive measurements and is iteratively optimized.
At each smoothing time step, an additional set of states X, V, and B are added to the graph.  
Additionally, measurement factors for the GPS and IMU sensors are added along with a bias smoothness factor. 
To keep the computational load low while maintaining high accuracy, the factor graph contains state nodes for measurements taken at 10~Hz. 
The factors in the graph correspond to GPS measurements and pre-integrated IMU measurements~\cite{Forster-RSS-15}. 
Online, the IMU measurements are integrated to interpolate the 10~Hz smoothed position to publish the state estimate at 200~Hz.  
Example trajectories are shown in Figure~\ref{fig:poses}.

In practice, the measurements from the GPS sensor can drift slightly from day to day primarily due to the RTK correction antenna position not being fixed from one test to the next. To counteract these changes, the robot is always positioned at the same place on the track when the state estimator is started. This track position is used as the origin of a local Euclidean coordinate system, oriented tangent to the GPS reference ellipsoid. This prevents GPS drift from effecting the vehicle state estimate relative to the fixed track boundaries.

\subsection{Semantic Segmentation Using Convolutional Neural Networks}

An important step in dynamic visual scene understanding for autonomous driving is to analyze the images captured from the on-board cameras and segment important objects such as cars, pedestrians, or signs (using bounding boxes) and label the key surfaces in the environment, such as streets, sidewalks, and building facades. The task of grouping pixels and labeling the surfaces they belong to is known as semantic segmentation. 
In the specific context of the racing domain, the primary goal is to be able to label pixels as track or not-track so that the vehicle knows where it can drive. In this section, we describe a method to create and train a deep convolutional neural network (CNN) \cite{krizhevsky2012imagenet,Goodfellow-et-al-2016} architecture that receives an image from the onboard cameras and outputs a pixel-wise labeling of track and not-track.

Recent works have demonstrated the success of deep neural network architectures consisting of multiple CNN layers in solving challenging semantic segmentation problems \cite{Kundu_2016_CVPR,badrinarayanan2015segnet,long2015fully,chen2016deeplab}. 
The structure of the CNN takes advantage of the spatial properties of an image, namely that nearby pixels are likely to have similar labels. In our case, we wish to perform semantic segmentation by learning a function that can map an input image $I(u,v)$ to an output image with binary pixel labels corresponding to track and non-track and given by:
\begin{equation}
Y(u,v) = f(I(u',v'),\Theta)\qquad u'=u-r\dots u+r\quad v'=v-r\dots v+r,
\end{equation}
where $u$ and $v$ are pixel coordinates, $r$ is the receptive field for an output pixel (which is dependent on the structure of $f$), and $\Theta$ is the set of parameters in $f(\dot)$ that we can change to create the desired function mapping.

The constructed deep neural network is composed of 10 convolutional layers and two, 2x2 max pooling layers after the 3rd and 6th layers to reduce the 640x480 sized input image to 160x128. All layers are 3x3 convolution kernels, except the last three layers which have 5x5 kernels. Layers before the first maxpool have 32 kernels, between the first and second have 64 kernels, and the final three have 96, 128, and 256 kernels, respectively.

The standard mini-batch gradient descent method was used to train a CNN function approximator to output the correct pixel labels given an input image. In order to use gradient descent, we must define a loss function $l$ to minimize. In our case, we use the cross-entropy function which rewards the network for producing an output closer to the desired value of not track (1) or track (0). The gradient of each parameter is computed with respect to the loss function ${\partial \Theta}/{\partial l}$ to give a direction and magnitude to move each parameter that pushes the generated output closer to the desired output. However, because computing this gradient over all images is computationally prohibitive, the derivative is computed for a random subset of 10 images, the parameters are moved a small amount, $\Theta$, in the direction of the gradient, and the process is repeated until convergence.
\begin{figure}[!htbp]
	\centering
	\includegraphics[width=\textwidth]{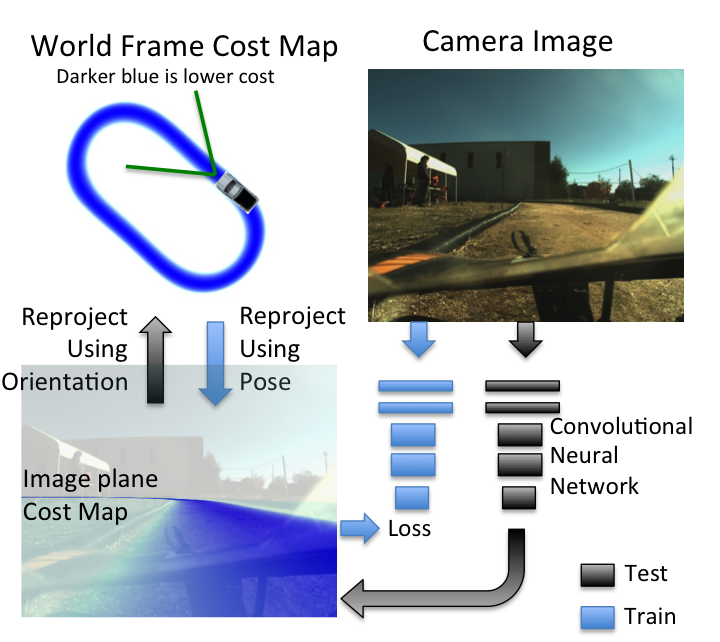}
	\caption{Convolutional neural network pipeline for training and testing pixel-wise image labeling of track and not track using images from onboard AutoRally. For training, a survey of the track boundaries, calibrated inertial measurement unit sensor to camera transform, and state estimate are used to automatically label image pixels. For testing, the network labels each pixel of an input image as track or not track.}
	\label{fig:cnn_pipeline}
\end{figure}
A large set of training data must be created to train a neural network from scratch. Over 100,000 ground truth labeled images were automatically generated using data collected from the AutoRally platform. The images were labeled using a precomputed GPS survey of the test track, the output from the state estimator, and the IMU to perform camera calibration. The training and testing pipeline is shown in Figure~\ref{fig:cnn_pipeline}. 
The state estimator frame is at the IMU, so a homography matrix must be computed that transforms the surveyed track map from world coordinates to image plane coordinates using the calibrated transformation between the IMU and the camera
\begin{equation}
H = k T^{\mathrm{car}}_{\mathrm{im}} T^{\mathrm{world}}_{\mathrm{car}},
\end{equation}
where $T^{\mathrm{world}}_{\mathrm{car}}$ is the position of the car in world coordinates, $T^{\mathrm{car}}_{\mathrm{im}}$ is the transformation between the IMU and camera reference frames, and $k$ is the matrix of camera intrinsics.  Given this mapping, points in the ground coordinate frame can be projected into the image
\begin{equation}
p_{\mathrm{im}} = \hat{H} p_{\mathrm{world}}; \hat{H} = 
\begin{bmatrix}
    	H_{11}       & H_{12} & H_{14} \\
    	H_{21}       & H_{22} & H_{24} \\
    	H_{31}       & H_{32} & H_{34}
\end{bmatrix},
\end{equation}
where $p_{\mathrm{im}}$ and $p_{\mathrm{world}}$ are homogeneous points.  Using this method, all points on the image plane for a given image can be projected to the ground plane and given a ground truth label.
\section{Experimental Results}
\label{sec:expResults}

In this section we describe the testing facility, show and validate the results from the parameter estimations using the bifilar pendulum method, standard UKF, the ALM-UKF, and the neural network cost map estimation.

Data for all of the estimation results were collected with a human manually driving the AutoRally robot around GT-ARF. The same data was also used to train the dynamics model used in the model predictive path integral (MPPI) controller described in ``\nameref{sidebar:mppi}". The data consists of approximately 30 minutes of human controlled driving at speeds varying between 4 and 10~m/s. The driving was broken into five distinct behaviors: (1) normal driving at low speeds (4-6~m/s), (2) zig-zag maneuvers performed at low speeds (4-6~m/s), (3) linear acceleration maneuvers which consist of accelerating the vehicle as much as possible in a straight line, and then braking before starting to turn, (4) sliding maneuvers, where the pilot attempts to slide the vehicle as much as possible, and (5) high speed driving up to 10~m/s where the pilot simply tries to drive the vehicle around the track as fast as possible. Each one of these maneuvers was performed for three minutes while moving around the track clockwise and for another three minutes moving counter-clockwise.

\subsection{Georgia Tech Autonomous Racing Facility}

\begin{figure}
  \centering
  \begin{subfigure}{0.49\textwidth}
    \centering
    \includegraphics[height=4.75cm]{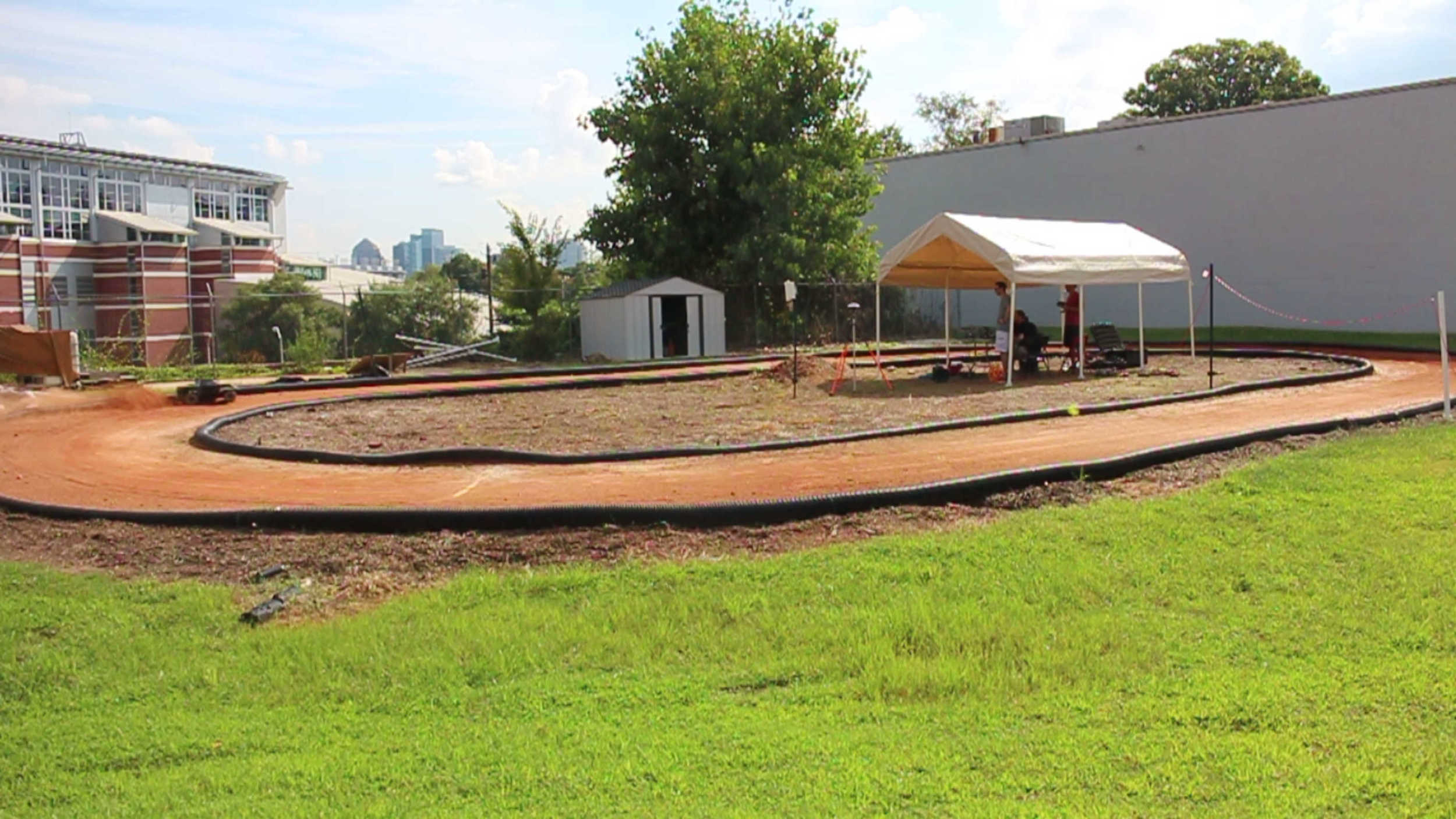}
    \caption{}
    \label{fig:track}
  \end{subfigure}
  \begin{subfigure}{0.49\textwidth}
  	\centering
    \includegraphics[height=4.75cm]{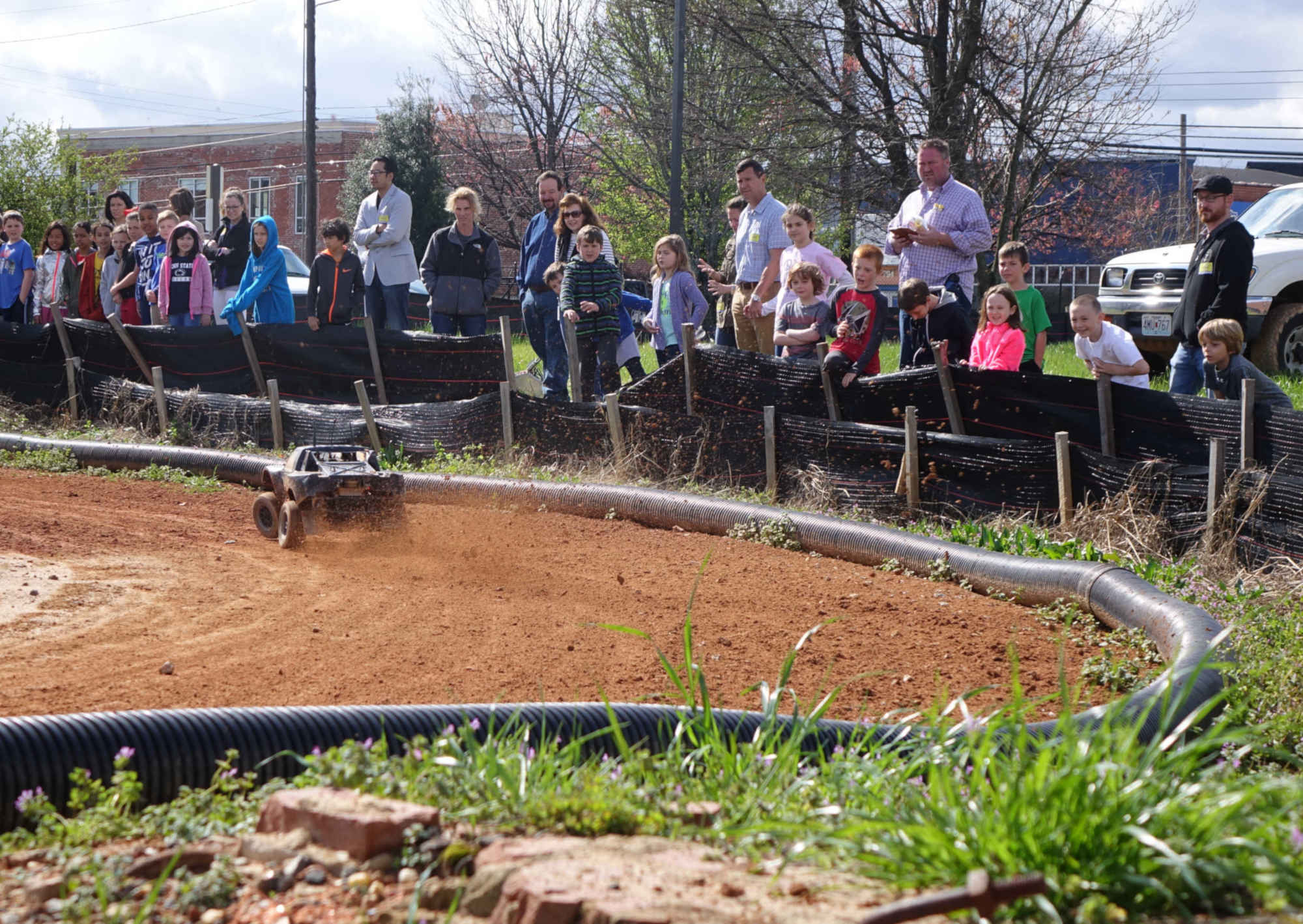}
    \caption{}
    \label{fig:demo}
  \end{subfigure}
  \caption{Georgia Tech Autonomous Racing Facility. (\subref{fig:track}) Aggressive autonomous driving experiments. (\subref{fig:demo}) Autonomous driving demo for third grade students and parents.}
  \label{fig:gtarf}
\end{figure}

All experiments were conducted at the GT-ARF, shown in Fig.~\ref{fig:gtarf}, which is a 68~m long dirt track with required site infrastructure to support autonomous vehicle testing. 

The track is a 3.3~m wide flat clay surface with two straights connected by 180 deg constant radius turns. The outer dimensions of the track are 27.5 m and 15.5 m, and the straights are 11.5~m long. 
The boundaries of the track are 0.15 m diameter corrugated drainage pipe secured in place with stakes to provide a semi-rigid crash barrier. 
The controlled track environment allows the robot to operate fully autonomously up to the limits of its mechanical, electrical, and software systems and beyond the friction limits of the track.

A ground station is set up in the center of the track which consists of an OCS laptop, wireless runstop box, and a base station GPS module that provides RTK corrections. The OCS laptop and associated is used to remotely communicate and monitor the status of the robot.

\subsection{Bifilar Pendulum}

Many vehicle parameters can be directly measured with a scale, ruler, or are available through manufacturer documentation. Those parameters for the AutoRally platform are listed in Table~\ref{tab:calibParams}.

\begin{table}
  \renewcommand{\arraystretch}{1.1}
  \centering
  \caption{AutoRally robot parameters. Values were measured with a digital scale, ruler, or provided by manufacturer documentation. Rear axle offset distances are with respect the to the center of gravity, CG.}
  \begin{tabular}{| l | c | c |}
    \hline
    Parameter & Value & Units \\ \hline
    Total mass & 21.88 & kg \\
    Front wheel mass, each & 0.82 & kg \\
    Rear wheel mass, each & 0.89 & kg \\
    Overall length & 0.90 & m \\
    Overall width & 0.46 & m \\
    Overall height & 0.32 & m \\
    Wheelbase & 0.57 & m \\
    Rear axle to CG, x offset & 0.23 & m \\
    Rear axle to CG, z offset & 0.12 & m \\
    Front Track & 0.395 & m \\
    Rear Track & 0.405 & m \\
    Wheel diameter & 0.195 & m \\
    \hline
  \end{tabular}
  \label{tab:calibParams}  
\end{table}

Data was collected using the setup shown in Figure~\ref{fig:moi} to estimate the MOI of the primary axis of the robot and the axis of rotation for the front and rear wheels. Table~\ref{tab:experimentalData} details the measured parameters for each configuration and the resulting MOI. In each case, $T$ is calculated as an average of 60 oscillation periods.
\begin{table}
  \renewcommand{\arraystretch}{1.2}
  \caption{Moments of inertia (MOI) about primary axes and related parameters. Values are experimentally calculated using the bilifilar pendulum method and~(\ref{eq:bifilar}), and the measured masses and lengths. The oscillation period $T$ is calculated by averaging 60 oscillation periods using the bifilar pendulum method. $g=9.81$ was used for all calculations. The front and rear wheel MOI include both wheels.}
  \label{tab:experimentalData}
  \centering
  \begin{tabular}{| c | c | c || c | c |}
    \hline
    $T$ [s] & b [m] & d [m] & Parameter & MOI [kg-m2]\\ \hline
    2.63 & 0.14 & 1.90 & $I_{\rm x}$ & 0.347\\ 
    2.12 & 0.295 & 1.88 & $I_{\rm y}$ & 1.131\\ 
    2.48 & 0.26 & 1.93 & $I_{\rm z}$ & 1.124\\ 
    1.81 & 0.182 & 0.935 & $I_{f}$ & 0.048\\ 
    1.67 & 0.182 & 0.915 & $I_{r}$ & 0.044\\
    \hline
  \end{tabular}
\end{table}

\subsection{Standard Unscented Kalman Filter}

We first implemented the standard JS-UKF using the three different vehicle models in the~\nameref{sec:estimation} Section.
The hyperparameters of the filter are critical for the
filter design, especially the process noise covariance $Q$ \cite{Saha2011}.
In this section, we tune the diagonal elements of these matrices recursively, until the parameterized vehicle model shows satisfactory simulation results. 

\begin{figure}[!htbp]
	\centering
	\includegraphics[width=\textwidth]{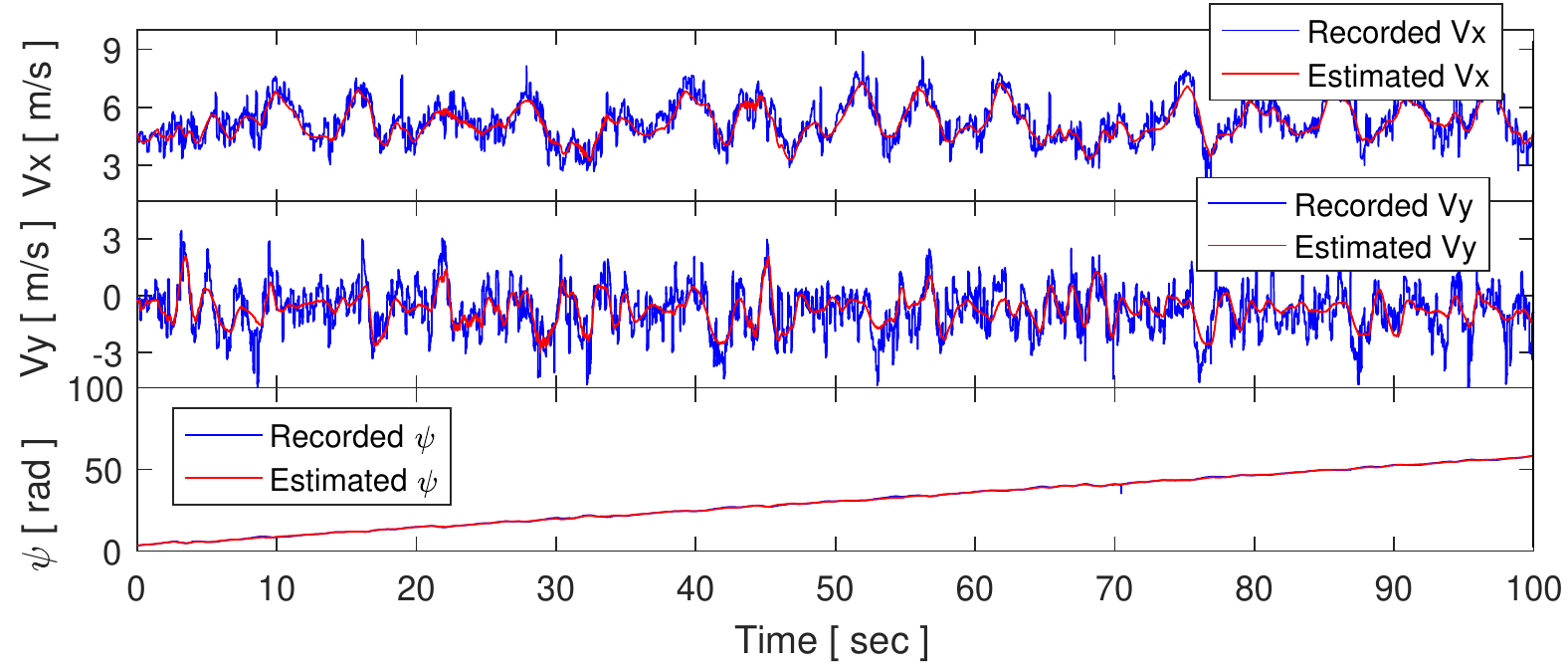}
	\caption{State estimation for the single-track model using joint-state unscented Kalman filter.}
	\label{06UKF_esti}
\end{figure}

We selected 113 seconds of experimental data generated by a human driver with the AutoRally vehicle. The first 100 seconds data were used to tune the hyperparameters and estimate the vehicle parameters, and the remaining 13 seconds (a complete cycle around the testing track) were used to validate the results. Figure~\ref{06UKF_esti} shows the estimates for several selected states of the system for the single-track model.
It can be seen that the estimates of the states agree well with the data. The results for the other vehicle models were similar and hence are omitted.

\begin{figure}[!htbp]
	\centering
	\includegraphics[width=\textwidth]{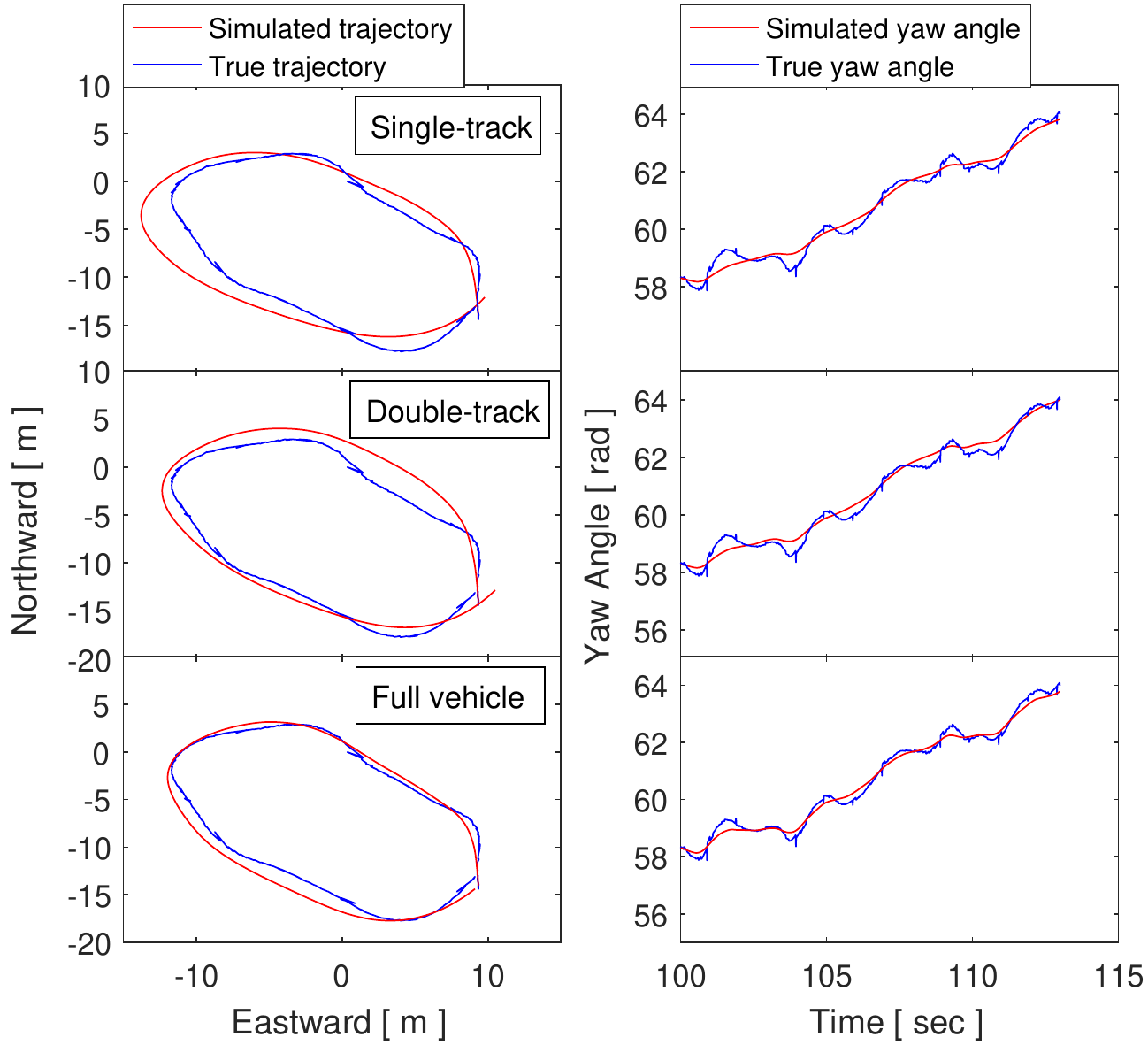}
	\caption{Simulation results of the estimated vehicle models using standard unscented Kalman filter.}
	\label{19UKF_valid}
\end{figure}

Next, we validated the estimated parameters in simulation.
This was done in order to ensure that the parameters we obtained were able to satisfactorily reproduce the data, hence accurately predicting the vehicle's motion in practical applications.
Figure~\ref{19UKF_valid} shows the simulated trajectories for different vehicle models configured with the estimated parameters.
The results in Figure~\ref{19UKF_valid} indicate that the larger the number of degrees of freedom (DoF) of the model, the more accurate the results and the better the agreement with data.

\subsection{Adaptive Limited Memory Unscented Kalman Filter}

Instead of tuning the noise, we implemented the ALM-UKF to find the suboptimal estimation of the noise statistics online, during which the augmented-state and the noise are estimated simultaneously. Experimental data collected with the AutoRally robot were used to validate the ALM-UKF, which is presented in algorithmic form in~\cite{you2017vehicle}. The noise samples at each time step $k$ are from the estimation based on the last 10 seconds of data. \par 
\begin{figure}[!htbp]
	\centering
	\includegraphics[width=\textwidth]{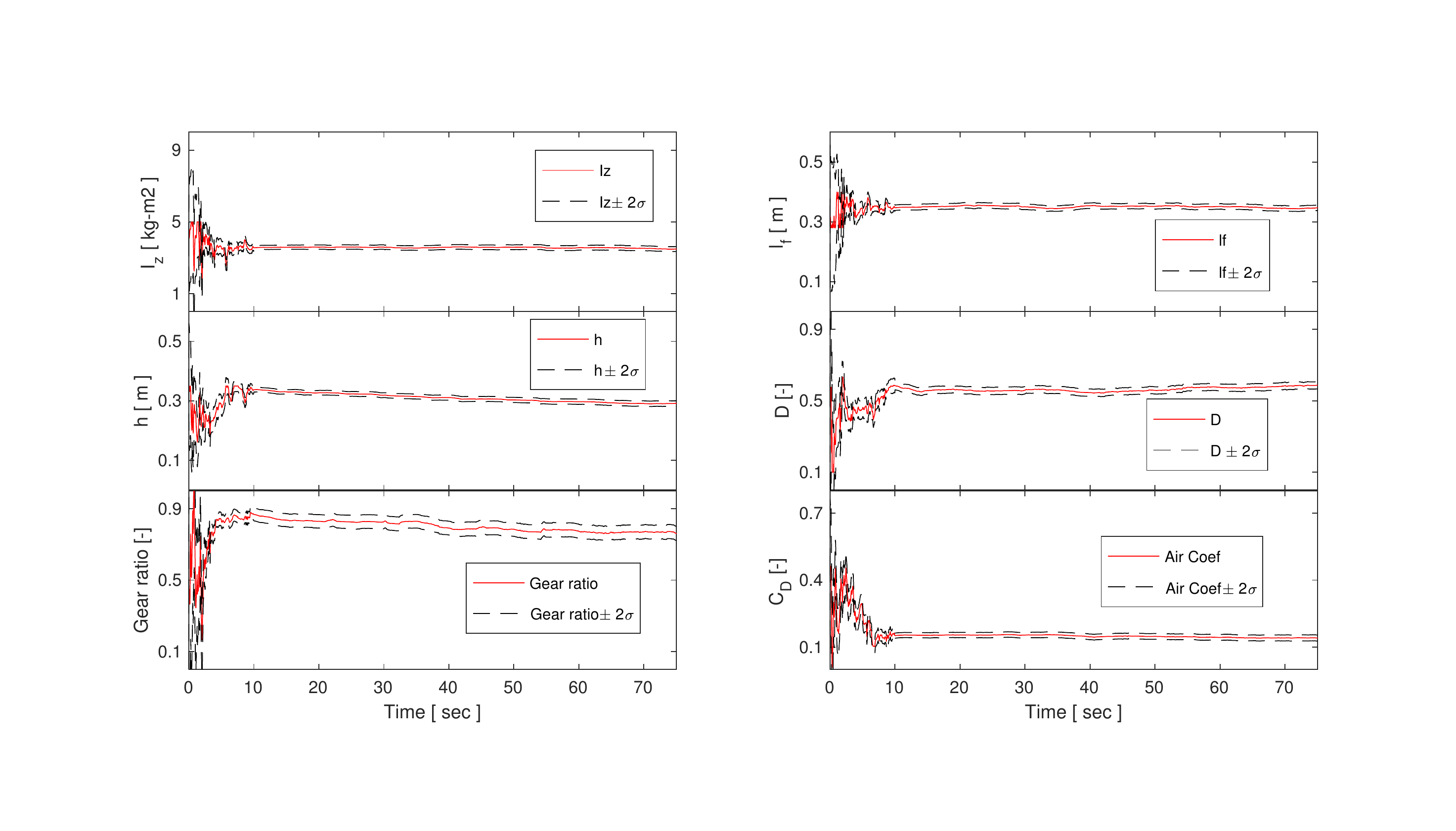}
	\caption{Convergence of the vehicle parameters along with the estimation process. All the parameters converge fast and stabilize after about 10 to 20 seconds.}
	\label{20AJUKF_paras}
\end{figure}
The estimation of the  the velocities, yaw angle, and positions states is not difficult. We thus only show the estimation results of the unknown vehicle parameters. We implemented the ALM-JUKF~\cite{you2017vehicle} to estimate the parameters of a full vehicle model from~(\ref{eqn:DoubleTrack1})-(\ref{eqn:FullVehicle6}) using the AutoRally experimental data, which includes the measurements for $V_x$, $V_y$, $V_z$, $r$, $\dot{\theta}$, $\dot{\phi}$, $\psi$, $\theta$, $\phi$ and the GPS coordinates.

In order to have stable state estimation, the joint-state system with the parameter dynamics equations in~(\ref{eqn:P}) is required to be observable.  To this end, we calculate the Jacobian of the joint-state system and inspect the rank of the observability matrix. The result indicates that the spring stiffness and the damping coefficients  of the suspension system  $K_i$ and $C_i$ ($i={\rm f, r}$) are unobservable (see Figure~\ref{11RidingModel}), which means, one cannot uniquely identify $K_i$ and $C_i$ ($i={\rm f, r}$) based on current measurements. We are able to address this issue to obtain the observability by fixing the values of any two of the four parameters, specifically we used $K_{\rm f}=K_{\rm r}=2000$~N/m. The values for $K_{\mathrm{f}}$ and $K_{\mathrm{r}}$ were chosen by assuming a 1-2~cm average deformation of the springs caused by the gravity of the sprung mass.

It is also worth mentioning that the artificial Gaussian process noise $w_k^{\rm p}$ in (\ref{eqn:P}) is used to change the parameter $p$ when the UKF is working. However, if the value of $w_k^{\rm p}$ is large, the parameter $p$ will be changed by a large amount at each time step. This condition may further cause the filter to diverge, since the parameterized vehicle models are sensitive to $p$ and may thus get unstable for unreasonable values of $p$. We addressed this problem by rescaling the diagonal entries of $Q^{\rm p}$ to be some small positive values at each time step. Other discussions on the numerical instability problems of the UKF can be found in \cite{Myers1976,Qi2015}.

Figure~\ref{20AJUKF_paras} shows the time trajectories of several parameters listed in Table~\ref{tab:sysidComp} during the estimation process, where all the parameters converge fast and stabilize after about 10 to 20 seconds.
\begin{figure}[!htbp]
	\centering
	\includegraphics[width=\textwidth]{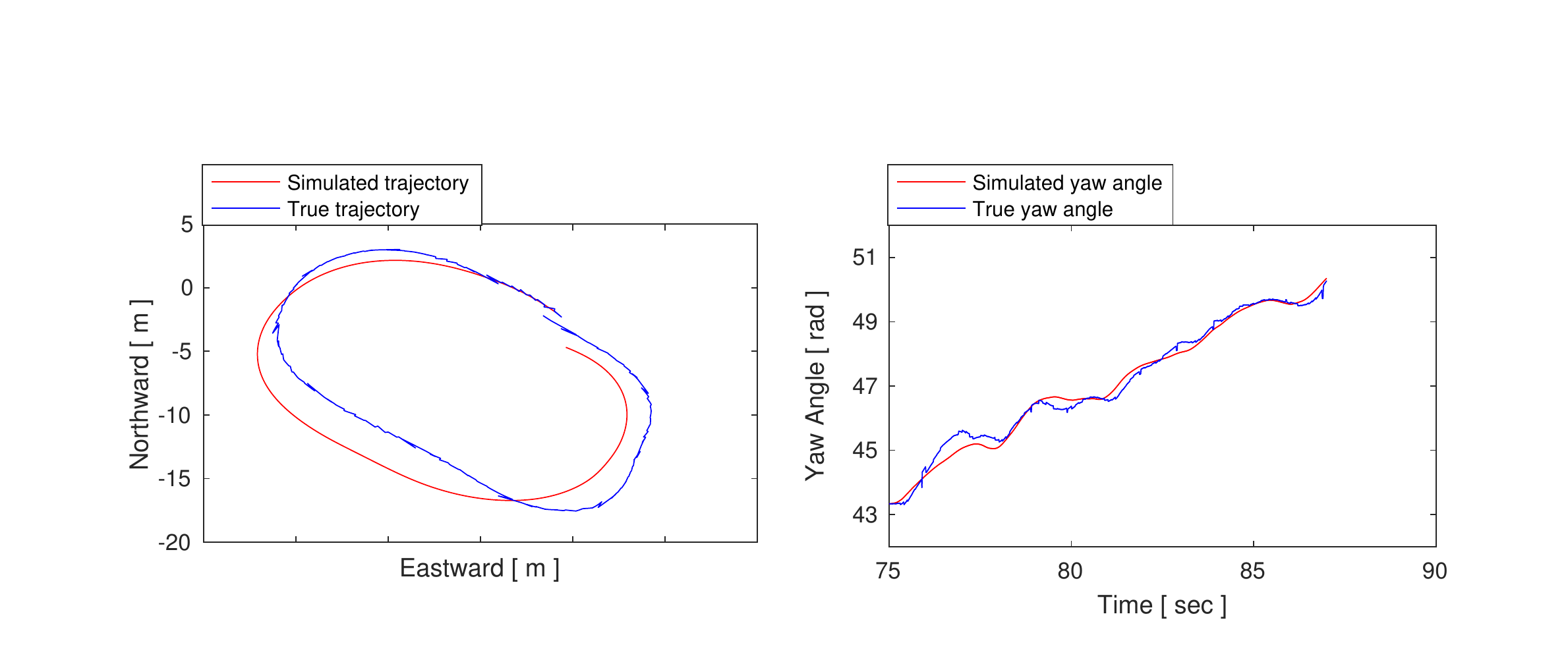}
	\caption{Simulation results of the estimated vehicle models using adaptive limited memory joint-state unscented Kalman filter.}
	\label{21AJUKF_vali}
\end{figure}
Figure~\ref{21AJUKF_vali} compares simulated outputs from the full vehicle model using the estimated parameters for the AutoRally platform and the real world data.
It can be seen that, as expected, the identified vehicle model can satisfactorily reproduce the data.
Data collected using the AutoRally vehicle show obvious non-Gaussian noise which may have some effect on the estimation process.

Compared with the results in Figure~\ref{19UKF_valid}, the simulated trajectories of the AutoRally vehicle in Figure~\ref{21AJUKF_vali} show larger deviation from the data.
The reason may be that we tuned the estimation of the noise statistics of the standard UKF to be optimal (in some degree), but the ALM-UKF algorithm was using a suboptimal estimator for the noise statistics. 
The advantages of the ALM-UKF are that it is more efficient and can work online.
We expect that the ALM-UKF algorithm is especially useful for time-varying parameters estimation problems such as estimation of a linear parameter varying (LPV) driver model~\cite{you2016optimal}.

Table~\ref{tab:almukf} shows the MF model parameters for the single track vehicle model estimated using the UKF for the AutoRally platform. All of the parameters stabilized after about 20 seconds of data.

\begin{table}
  \renewcommand{\arraystretch}{1.2}
  \caption{Tire model parameters estimated by the unscented Kalman filter for the single track model for the AutoRally platform. Human driving data was collected at the Georgia Tech Autonomous Racing Facility.}
  \setlength{\belowcaptionskip}{1pt}
  \label{tab:almukf}
  \centering
    \begin{tabular}{| c | c | c|}
      \hline
      Parameter & Value & Units \\
      \hline
      B & 1.1559 & none\\
      C & 1.1924 & none\\
      D & 0.9956 & none\\
      E & -0.8505 & none\\
      $S_h$ & -0.0540 & m\\
      $S_v$ & 0.1444 & m\\
      \hline
    \end{tabular}
    \label{tab:tiremodel}
  
\end{table}

Some parameters were estimated by both the bilfilar pendulum and the ALM-UKF methods. Table~\ref{tab:sysidComp} compares these values. While the mass and moment of inertia values estimated by both methods agree, the estimated dimensions are not as closely aligned with hand-measured values. This is because in the bilfilar method all parameters are measured when the vehicle is stationary, whereas the ALM-UKF method uses real world driving data where these dimensions are constantly changing due to suspension and steering articulation.

\begin{table}
  \renewcommand{\arraystretch}{1.2}
  \caption{Parameter comparison for the adaptive limited memory unscented Kalman filter (ALM-UKF) and bifilar pendulum methods. The parameters that do not closely match across methods are because the ALM-UKF method uses real world driving data where these dimensions are constantly changing due to suspension and steering articulation and the bilfilar method estimates all parameters when the vehicle is suspended and stationary.}
  \label{tab:sysidComp}
  \centering
  \begin{tabular}{| c | c | c | c | c |}
    \hline
    Paramter & ALM-UKF & Bifilar & Difference & units \\ \hline
    m           & 20.6093 & 21.88  & 5.8 \% & kg \\ 
    I$_{\rm z}$ & 1.024 & 1.124 & 8.9 \% & kg-m2 \\ 
    I$_{\rm r}$ & 0.0499 & 0.044 &  13.4 \% & kg-m2 \\ 
    h 			& 0.0961  & 0.12  & 19.9 \% & m \\
    ${\rm \ell}_{\rm f}$ & 0.4650 & 0.34 & 36.8 \% & m \\
    \hline
  \end{tabular}
\end{table}

\subsection{Track Surface Labeling}
\label{sec:Labeling}
\setlength{\fboxsep}{0pt}
\begin{figure}
  \centering
  \begin{subfigure}{0.32\textwidth}
  	\setlength{\fboxsep}{0pt}
    \fbox{\includegraphics[width=\textwidth]{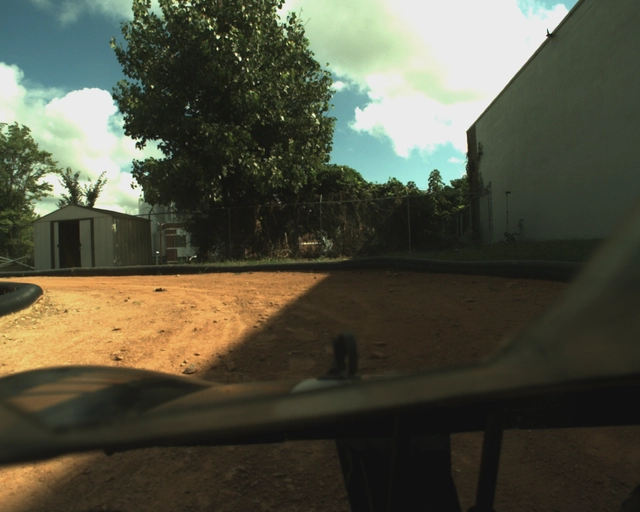}}
    \caption{}
    \label{fig:input_image}
  \end{subfigure}
  \begin{subfigure}{0.32\textwidth}
    \fbox{\includegraphics[width=\textwidth]{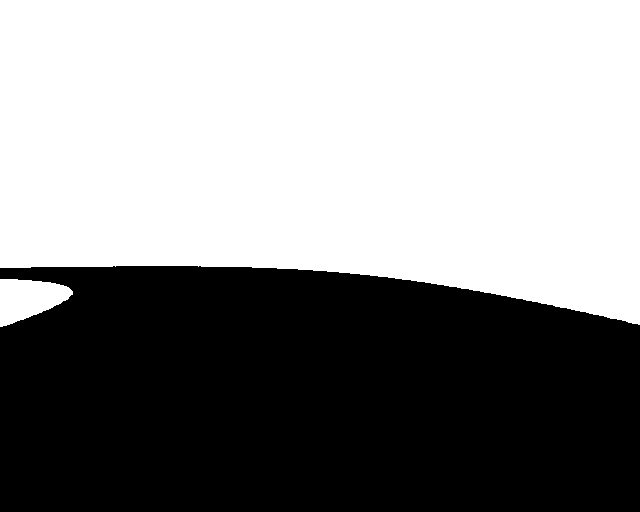}}
    \caption{}
    \label{fig:gt_image}
  \end{subfigure}
  \begin{subfigure}{0.32\textwidth}
    \fbox{\includegraphics[width=\textwidth]{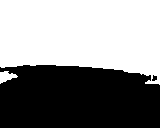}}
    \caption{}
    \label{fig:output_image}
  \end{subfigure}
  \caption{Neural network track surface labeling from camera images. (\subref{fig:input_image}) Training image captured by one of the cameras on AutoRally at the Georgia Tech Autonomous Racing Facility. (\subref{fig:gt_image}) Ground truth image generated by labeling pipeline. (\subref{fig:output_image}) Pixel labeling of track and not track produced by neural network.}
  \label{fig:nn_example}
\end{figure}
\begin{figure}
  \centering
  \begin{subfigure}{0.32\textwidth}
    \fbox{\includegraphics[width=\textwidth]{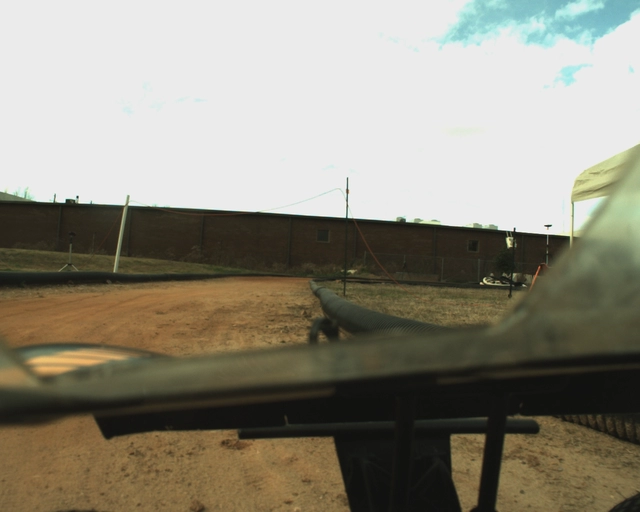}}
    \caption{}
    \label{fig:input_image_bad}
  \end{subfigure}
  \begin{subfigure}{0.32\textwidth}
    \fbox{\includegraphics[width=\textwidth]{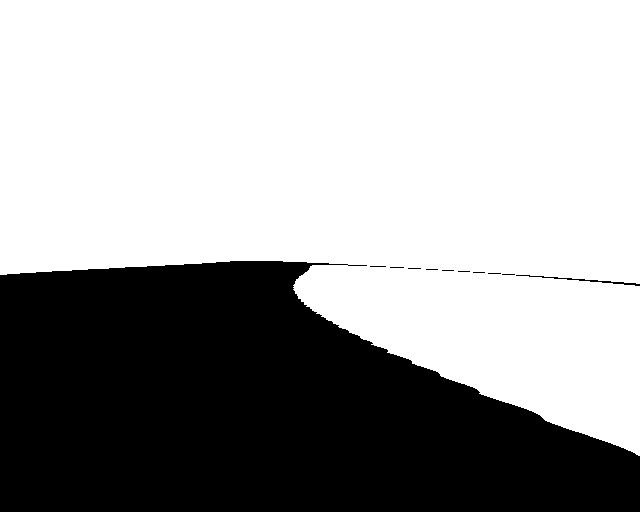}}
    \caption{}
    \label{fig:gt_image_bad}
  \end{subfigure}
  \begin{subfigure}{0.32\textwidth}
    \fbox{\includegraphics[width=\textwidth]{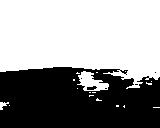}}
    \caption{}
    \label{fig:output_image_bad}
  \end{subfigure}
  \caption{Neural network image labeling failure case. (\subref{fig:input_image_bad}) Image input from camera onboard AutoRally. (\subref{fig:input_image_bad}) Ground truth image generated by labeling pipeline. (\subref{fig:output_image_bad}) Pixel labeling of track and not track produced by neural network. The failure is likely caused by the appearance similarity between the track surface and dry grass outside of the track.}
  \label{fig:nn_failure}
\end{figure}
A dataset of over 100,000 images, along with the corresponding position and orientation in a local coordinate frame, were recorded over the course of several days of testing with the MPPI algorithm at GT-ARF. The images were then post-processed using the automatic labeling pipeline to produce ground truth training images for the labeling task. The neural network was trained using the Tensorflow \cite{tensorflow2015} framework for 100,000 mini-batches, each containing 10 images. The results for an example training image can be seen in Figure \ref{fig:nn_example}, and a failure case of the network can be seen in Figure \ref{fig:nn_failure}. The failure was a result of the dry grass adjacent to the track having a similar color and texture to the track surface, and a mislabeling of the interior of the plastic protective body. Overall, the neural network was able to correctly label 97\% of pixels on the testing set, and 91\% of the pixels on a held out test set. Note that this approach may not generalize to variations track conditions such as illumination changes and dynamic obstacles.
\section{Conclusion}
Despite decades of prior research and a renewed interest from technology companies and the research community, many gaps still remain in the capabilities of autonomous vehicles. This article introduces \emph{AutoRally}, a high-performance robotics testbed, 1:5 the size of a passenger car, that enables researchers to conduct experiments and collect real-world data under driving conditions that were previously untestable due to safety and cost concerns. The robust design and small size of AutoRally ameliorate the risk of high-speed crashes, allowing researchers to evaluate algorithms under conditions that would be too dangerous or expensive with a full-sized vehicle, and too complex to model accurately in simulation.

A variety of online and offline estimation algorithms have been developed and applied to the platform to determine properties frequently required by control systems, including moments of inertia and other difficult to measure properties. In addition, we developed an approach based on convolutional neural networks to address the task of online image segmentation trained from automatically labeled ground truth images. To date, the fleet of six AutoRally platforms has logged over 300~km of fully autonomous driving using only onboard sensing and computing at the Georgia Tech Autonomous Racing Facility, resulting in the validation of multiple control and perception algorithms.

AutoRally is open source, so all of the documentation needed to build, configure, and run the platform is publicly available on the AutoRally GitHub repositories, including build instructions, a parts list, files required for custom fabrication, and operating procedures~\cite{autorallyInstructions}. Tutorials and example controllers written in C++ and Python with ROS and Gazebo~\cite{autorallySoftware} are also available, along with a dataset of human and autonomous driving. See ``\nameref{sidebar:buildautorally}" for more information and tips if you are thinking of building your own AutoRally platform. AutoRally opens a new frontier for the safe development and testing of autonomous vehicle technologies across a much more diverse set of operating regimes and to a broader audience of investigators than previous experimental platforms have supported.

\section{Acknowledgments}

The authors wish to thank all of the undergraduate students who have helped build and maintain the fleet of AutoRally platforms including Jason Gibson, Jeffrey McKendree, Alexandra Miner, Dominic Pattison, Sarah Selim, Cory Wacht, and Justin Zheng. This work was made possible, in part, by the ARO through MURI award W911NF-11-1-0046 and DURIP awards W911NF-12-1-0377 and N00014-17-1-2318.
%

%
\newpage
\bibliographystyle{IEEEtran}
\bibliography{autorally}


\newpage 

\section[AutoRally Summary]{Sidebar: AutoRally Summary}
\label{sidebar:autorallySummary}

AutoRally is an open-source 1:5 scale autonomous vehicle testbed for students, researchers, and engineers who are interested in autonomous vehicle technologies. It is designed with robustness and ease-of-use in mind. At 1 m in length, weighing 22 kg, and with a top speed of 90 kph, the platform is large enough to host powerful onboard computing and sensing and run state-of-the-art algorithms. At the same time, it is simple and small enough to be maintained and operated by two people, all while providing the capability to explore driving scenarios including drifting, jumping, high speed driving, and multi-vehicle interactions.

Build instructions are publicly available, along with a parts list, computer-aided design (CAD) models for fabricating custom components, and operating procedures. The platform uses the Robot Operating System (ROS) and can be programmed in Python or C++. Tutorials, reference algorithms, a Gazebo-based simulation environment, and a dataset structured as ROS bag files, are available from the AutoRally website. The fleet of six AutoRally platforms at Georgia Tech have been used to demonstrate control, perception, and estimation research in a high speed, off road driving domain. To date, the fleet has driven hundreds of kilometers autonomously at the Georgia Tech Autonomous Racing Facility.

\newpage

\section[Society of Automotive Engineers Levels of Driving Automation]{Sidebar: Society of Automotive Engineers Levels of Driving Automation}
\label{sidebar:driveAutomation}

Issued January 2014, the Society of Automotive Engineers (SAE) International\textquotesingle s J3016~\cite{sae2014taxonomy} provides a common taxonomy and definitions for automated driving in order to simplify communication and facilitate collaboration within technical and policy domains. It defines more than a dozen key terms, including those italicized below, and provides full descriptions and examples for each level.

The report\textquotesingle s six levels of driving automation span from no automation to full automation, and are described in detail in Figure~\ref{fig:saeJ3016}. A key distinction is between level 2, where the human driver performs part of the dynamic driving task, and level 3, where the automated driving system performs the entire dynamic driving task.

These levels are descriptive rather than normative, and technical rather than legal. They imply no particular order of market introduction. 
Elements indicate minimum rather than maximum system capabilities for each level. A particular vehicle may have multiple driving automation features such that it could operate at different levels depending upon the feature(s) that are engaged.

System refers to the driver assistance system, combination of driver assistance systems, or automated driving system. Excluded are warning and momentary intervention systems, which do not automate any part of the dynamic driving task on a sustained basis and therefore do not change the human driver\textquotesingle s role in performing the dynamic driving task.

\begin{figure}[ht]
  \setcounter{figure}{0}
  \renewcommand{\thefigure}{S\arabic{figure}}
  \centering
    \includegraphics[width=\textwidth]{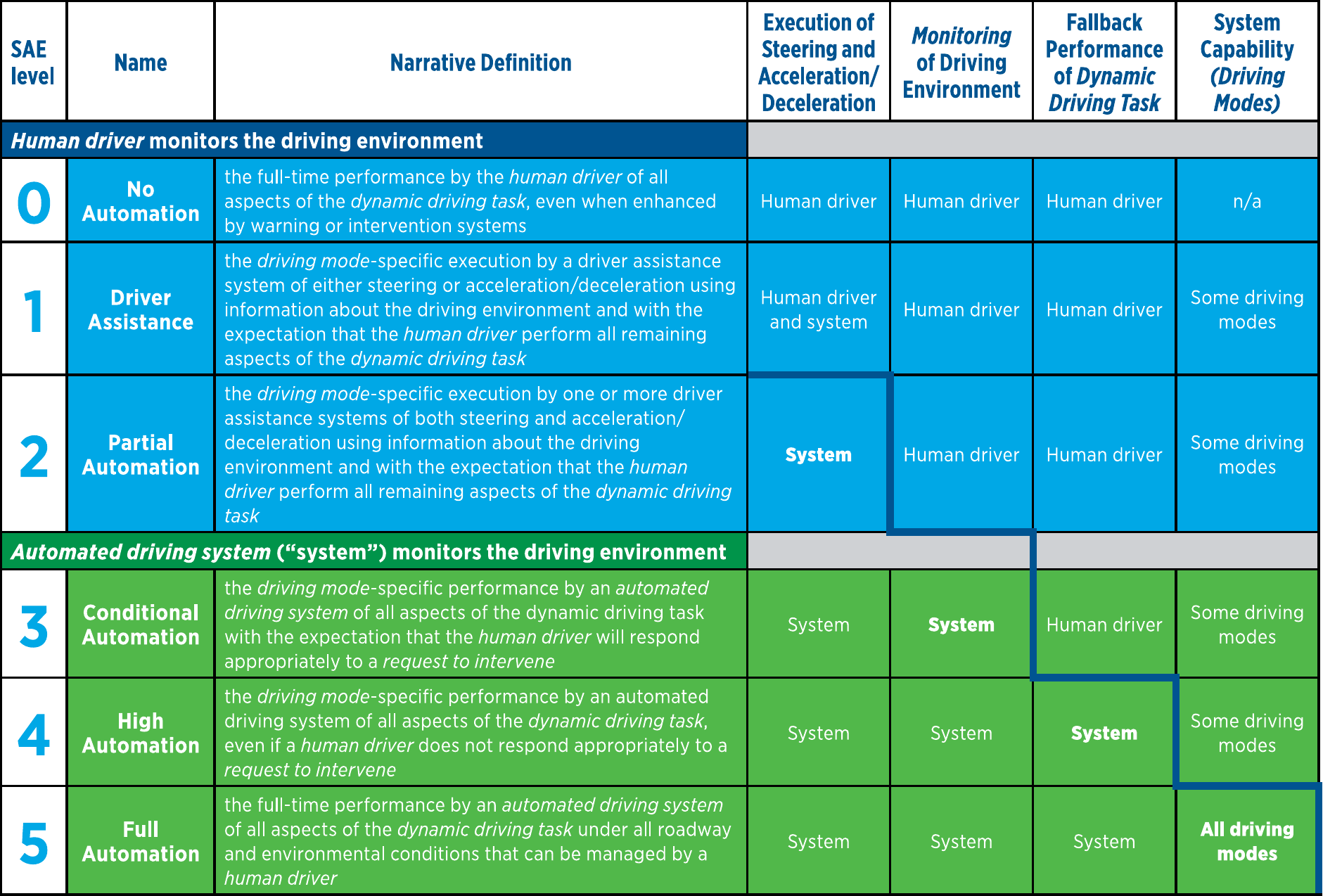}
  \caption{Society of Automotive Engineers (SAE) International J3016 levels of driving automation. (Courtesy of SAE International)}
  \label{fig:saeJ3016}
\end{figure}

\newpage


\section[Build Your Own AutoRally Platform]{Sidebar: Build Your Own AutoRally Platform}
\label{sidebar:buildautorally}

AutoRally is an open source platform, so complete documentation to build, configure, and run the platform is publicly available through the AutoRally platform build instructions GitHub repository~\cite{autorallyInstructions}. Documentation includes build instructions for the chassis and compute box, a parts list with suggested vendors, wiring diagrams, files required for custom part fabrication, and operating procedures. Tutorials and example controllers written in C++ and Python that use the Robot Operating System (ROS), a Gazebo-based simulation environment that resembles the Georgia Tech Autonomous Racing Facility (GT-ARF) oval track, and a dataset of human and autonomous driving captured at the GT-ARF tracks are available through the AutoRally software GitHub repository~\cite{autorallySoftware}. AutoRally is designed for robustness, ease of use, and reproducibility, so that a team of two people with limited knowledge of mechanical engineering, electrical engineering, and computer science can construct and then operate the testbed to collect real world autonomous driving data in whatever domain they wish to study. Links to tutorials for the background skills required to construct AutoRally are included in the instructions.

Construction time for an AutoRally chassis is 40 hours, and 60 hours for a compute box. The full platform construction time of 100 hours does not include custom part fabrication time that depends on the tools available. AutoRally can take significantly less than 100 hours to construct if you already have experience with radio controlled (RC) electronics, soldering, computer construction and wiring, or Ubuntu configuration. Conversely, the platform can take much more time to construct if individual assemblies are not thoroughly tested before integration, which can result in time consuming rebuilds during verification.

For fabrication of custom components, access to a 3D printer, laser cutter, water jet cutter, and aluminum welding are suggested. If you do not have access to a 3D printer, all custom parts provided as stereolithography (STL) files with the build documentation can be fabricated by a 3D printing service, many of which are available online. If you do not have access to a laser cutter, the custom foam and acrylic parts can be cut by hand with a blade by using the provided portable document format (PDF) files printed on US Letter paper as stencils. If you do not have access to a water jet for cutting aluminum parts or welding equipment for aluminum, most local metal fabrication shops should be able to fabricate the compute box enclosure and front brake bracket using the drawing interchange (DXF) files and bend patterns included in the instruction materials.

\newpage

\section[Georgia Tech Smoothing and Mapping]{Sidebar: Georgia Tech Smoothing and Mapping}
\label{sidebar:gtsam}

The Georgia Tech Smoothing and Mapping (GTSAM) toolbox~\cite{dellaert2012factor,dellaert2006square} is a C++ factor graph library released under the BSD license and developed by the Borg Lab at the Georgia Institute of Technology. 
It provides state-of-the-art solutions to the simultaneous localization and mapping (SLAM) and structure from motion (SFM) problems and can be used to model and solve many other simple and complex estimation problems. MATLAB and Python interfaces enable rapid prototyping, visualization, and user interaction.

GTSAM allows users to model a problem, such as state estimation, using a factor graph. A factor graph is a graphical model that contains variables related through factors. An example factor graph used for state estimation is shown in Figure~\ref{fig:FactorGraph}. The variables are vehicle states at specific points in time and the factors encode probabilistic information relating to sensor measurements of one or more of these variables.  The absolute or relative values of several variables and are the result of measurements by a sensor such as an inertial measurement unit (IMU) or global positioning system (GPS) receiver. This factor graph representation of the state estimation problem allows GTSAM to solve the problem by finding a Maximum A Posteriori (MAP) estimate of all of the variables in the graph.

In addition to simplifying SLAM and state estimation, it is easy to extend GTSAM to solve new problems which are naturally formulated as set of functions mapping hidden variables to measurements. Every factor in the graph can be expressed as a measurement function relating quantities to be estimated with sensor measurements. New types of factors are implemented by defining a measurement function and its derivative for each directly related variable

\sidebars

\begin{equation}
\label{eq_measurementfunction}
z_k = h_k(x_{k0},x_{k1} \dots x_{kn}),
\end{equation}
where $z_k$ is measurement $k$ and $x_{kn}$ is the state $x_n$ related to measurement $z_k$. As a concrete example, the GPS factors in Figure~\ref{fig:FactorGraph} are direct measurements of position, with additive Gaussian noise
\begin{equation}
\label{eq_measurementfunctiongps}
Z_i = X_i + \omega_i,
\end{equation}
where $Z_i=[x_{mi},y_{mi},z_{mi}]$ is the measured position in euclidean space, $X_i=[x_{i},y_{i},z_{i}]$ is the state variable we are estimating, and $\omega_i=[\omega_{xi},\omega_{yi},\omega_{zi}]$ is Gaussian noise on the measurement.

The noise in the variable $\omega$ has a straightforward interpretation as the measurement uncertainty.  
For Gaussian noise, one must assume that the measurement is unbiased, that is, the mean is zero.  
The variance of the noise then becomes the uncertainty on the measurement.  
For example, if the GPS receiver is specified to have a position standard deviation of 1~m in the x and y directions and 2~m in the z direction, these become the parameters of the Gaussian noise $\omega_{xi},\omega_{yi},\omega_{zi}$ in your factor graph.  

GTSAM exploits sparsity for computational efficiency. 
Typically, measurements only provide information on the relationship between a handful of variables, so the resulting factor graph will be sparsely connected. 
This is exploited by the algorithms implemented in GTSAM to reduce computational complexity. Even when graphs are too dense to be handled efficiently by direct methods, GTSAM provides iterative methods that are quite efficient regardless.

\newpage

\section[Model Predictive Path Integral Control]{Sidebar: Model Predictive Path Integral Control}
\label{sidebar:mppi}

All experimental data collected for training and testing track surface labeling was collected at the Georgia Tech Autonomous Racing Facility (GT-ARF) using the model predictive path integral (MPPI) controller. MPPI is a stochastic model predictive control (MPC) method that can drive AutoRally up to, and beyond, the friction limits of the track. It has been shown to work well in practice applied to AutoRally and a number of simulated systems~\cite{williams2016,williams2017,williams2017model}. A version of MPPI along with a dynamics model for AutoRally learned from human driving is available in the AutoRally GitHub repository.

MPC works by interleaving optimization and execution. First, an open loop control sequence over a finite time horizon is optimized, then the first control in that sequence is executed by the vehicle. Next, state feedback is received and the whole optimization process repeats. MPPI is a sampling based, derivative free approach to MPC which has been successfully applied to aggressive autonomous driving using learned non-linear dynamics \cite{williams2017}.

At each iteration, MPPI begins with the estimate of the optimal control sequence from the previous time step and uses importance sampling to generate thousands of new sequences of control inputs. These control sequences are then propagated forward in state space using the system dynamics, and each trajectory is evaluated according to a cost function. The estimate of the optimal control sequence is then updated with a cost-weighted average over the sampled trajectories. State feedback is then introduced to begin the next iteration. Real time execution of MPPI on AutoRally is enabled by the onboard Nvidia GPU.

Mathematically, let our current planned control sequence be $\left(u_0, u_1, \dots u_{T-1} \right) = U \in \mathbb{R}^{m \times T}$, and let $\left(\mathcal{E}_1, \mathcal{E}_2 \dots \mathcal{E}_K \right)$ be a set of random control sequences, with each $\mathcal{E}_k = \left(\epsilon_k^0, \dots \epsilon_k^{T-1}\right)$ and each $\epsilon_k^t \sim \mathcal{N}(u_t, \Sigma)$. Then, the control sequence update with
\begin{align}
\eta &= \sum_{k=1}^K \exp \left( -\frac{1}{\lambda} \left(S(\mathcal{E}_k) + \gamma \sum_{t=0}^{T-1} u_t^\rT \Sigma^{-1} \epsilon_k^t \right) \right), \\
U &= \frac{1}{\eta}\sum_{k=1}^K \left [ \exp \left( -\frac{1}{\lambda} \left(S(\mathcal{E}_k) + \gamma \sum_{t=0}^{T-1} u_t^\rT \Sigma^{-1} \epsilon_k^t \right) \right) \mathcal{E}_k \right].
\end{align}
The parameters $\lambda$ and $\gamma$ determine the selectiveness of the weighted average, also know as the temperature, and the importance of the control cost, respectively. The function $S(\mathcal{E})$ takes an input sequence and propagates it through the dynamics to find the resulting trajectory, and then computes the state-dependent cost of that trajectory sequence, which we denote as $C(x_0, x_1, \dots x_{T}) = \sum_{t=0}^T q(x_t)$. In the implementation of this algorithm on AutoRally, we only use an instantaneous running cost, so there is no terminal cost, and we sample trajectories on a GPU using several different dynamics models. The instantaneous running cost is
\begin{equation}
q(x) = w \cdot \left(C_M(p_x, p_y), (v_x - v_x^d)^2, 0.9^t I,  \left(\frac{v_y}{v_x}\right)^2 \right),
\label{equ:costFunction}
\end{equation}
where the first term, $C_M(p_x, p_y)$, is the positional cost of being at the body frame position $(p_x, p_y)$.  This positional cost is derived from a pre-surveyed GPS registered cost map in most of our data collection, and from the output of the neural network in the~\nameref{sec:Labeling} Section and~\cite{drews2017aggressive}. The second term is a cost for achieving a desired speed $v_x^d$, and the third term in the cost is an indicator variable which is turned on if the track-cost, roll angle, or heading velocity are too high. The final term in the cost is a penalty on the slip angle of the vehicle. The coefficient vector used in our experiments was $w = (100, 4.25, 10000, 1.75)$. Note that all but the first term are trivial to compute given the vehicle's state estimate, while first term requires analysis of the vehicle's environment. In practice, parameter weights of the cost function require some tuning, but the tuning is relatively intuitive as the weights correspond to easily understandable real world values.

This algorithm has driven the AutoRally platform several hundred kilometers around the GT-ARF.  Because it does not require explicit derivatives, it is can support advanced uses such as the noisy cost maps detailed in the~\nameref{sec:Labeling} Section.  It has shown good performance rejecting large disturbances from pot-holed and muddy track conditions while reliably running for a full battery charge at high speeds.

\newpage
\section{Author Biography}

Brian Goldfain received the B.S. in Electrical and Computer Engineering with a minor in robotics from Carnegie Mellon University in 2010 and the M.S. in Computer Science from Georgia Institute of Technology in 2013. He is currently a Robotics PhD student in the School of Interactive Computing at Georgia Institute of Technology. His research interest focus on autonomous vehicle testbed development and autonomous racing.

Paul Drews received the B.S. in Electrical Engineering from the Missouri University of Science and Technology in 2008. He is currently a Robotics PhD student in the School of Electrical and Computer Engineering at Georgia Institute of Technology.

Changxi You received the B.S. and M.S. degrees from the Department of Automotive Engineering, Tsinghua University of China, and M.S. degree from the Department of Automotive Engineering, RWTH-Aachen University of Germany. He is currently a Ph.D. student under the supervision of Prof. Panagiotis Tsiotras at the School of Aerospace Engineering, Georgia Institute of Technology. His current research interests focus on system identification, aggressive driving and control of (semi)autonomous vehicles.

Matthew Barulic received the B.S. in Computer Science from the Georgia Institute of Technology in 2016. He is currently a Software Engineer at Wheego Technologies, Inc.

Orlin Velev received the B.S. in Mechanical Engineering from the Georgia Institute of Technology in 2017. Currently, he is a Vehicle Structures Engineer at Space Exploration Technologies (SpaceX).

Panagiotis Tsiotras is a Professor in the School of Aerospace Engineering and the Institute for Robotics \& Intelligent Machines, Georgia Institute of Technology where he is the Director of the Dynamics and Control Systems Laboratory.

James M. Rehg is a Professor in the School of Interactive Computing at the Georgia Institute of Technology, where he is Director of the Center for Behavioral Imaging, co-Director of the Center for Computational Health, and co-Director of the Computational Perception Lab.

\end{document}